\documentclass[10pt]{article} 
\usepackage{enumitem}
\usepackage{xspace}
\usepackage{adjustbox}
\newcommand{\method}{\textsc{ReGUn}\xspace}

\usepackage[accepted]{tmlr}

\title{Reference-Guided Machine Unlearning}

\author{\name Jonas Mirlach\thanks{Equal contribution.} \email jmirlach@ethz.ch \\
\addr Department of Computer Science, ETH Zurich
\AND
\name Sonia Laguna\footnotemark[1]\email slaguna@inf.ethz.ch \\
\addr Department of Computer Science, ETH Zurich
\AND
\name Julia E. Vogt\email julia.vogt@inf.ethz.ch \\
\addr Department of Computer Science, ETH Zurich}
\def\month{10}  
\def\year{2026} 
\def\openreview{\url{https://openreview.net/forum?id=Vuo68BTxEc}} 

\begin{document}

\maketitle
\begin{abstract}

Machine unlearning aims to remove the influence of specific training data from a model while preserving its general utility. In vision, many approximate unlearning methods pursue this goal through degradation-based heuristics, such as loss maximization or random labeling. Yet making a model worse on forget samples is not the same as making it behave as if those examples had never been seen: these signals can be poorly conditioned, destabilize optimization, and harm generalization. We argue that approximate unlearning should instead prioritize distributional indistinguishability, aligning the model's predictive behavior on forget data with that on truly unseen data. Motivated by this principle, we propose Reference-Guided Unlearning (\method), a vision unlearning framework that uses disjoint held-out data to construct a principled, class-conditioned reference distribution for distillation. Rather than explicitly degrading predictions on forget examples, \method guides them toward non-member behavior through held-out supervision. Across multiple architectures, natural image datasets, and forget fractions,
\method achieves a competitive
forgetting--utility trade-off relative to standard approximate baselines while closely matching retrain-like membership inference behavior. As one instantiation of this principle, the results suggest that simple objectives designed around indistinguishability can provide an effective
alternative to complex degradation-based unlearning procedures.\looseness-1

\end{abstract}

\section{Introduction}
\label{introduction}

Machine unlearning is the principled process of updating a trained model to remove the influence of specified forget examples. It has become an operational requirement for deployed AI systems, driven by privacy regulations such as the European Union’s GDPR ``right to be forgotten''~\citep{mantelero2013eu} or the California Consumer Privacy Act~\citep{ccpa2018}, as well as the operational need to adapt models post-deployment~\citep{MULsurvey}. While retraining the model from scratch without the forget examples provides the most faithful solution, it is often computationally prohibitive at scale.
As a result, most unlearning methods rely on approximate updates (e.g., short fine-tuning or deletions) to emulate this retraining by reducing reliance on the forget data, but doing so without sacrificing retained utility remains challenging~\citep{MUprogress}.\looseness-1

A common strategy in approximate unlearning is to degrade model performance on forget examples, for instance, through loss maximization or random and pseudo-label supervision~\citep{MUBox}. However, such objectives can be poorly conditioned: they may induce large or misdirected gradients that change decision boundaries beyond the intended region and harm generalization~\citep{mavrothalassitis2026ascent}. 
To mitigate this damage, many methods introduce additional constraints, e.g., remaining close to the original model~\citep{SCRUB}, applying repair mechanisms~\citep{FastYetEffective}, or using constrained parameter editing~\citep{SALUN, SSD}.
This reveals a central mismatch: degradation objectives optimize for forgetting-like symptoms, but do not necessarily mimic unfamiliarity, i.e., the behavior expected from a model that had never seen the forget data.\looseness-1

\begin{figure}[t]
    \centering     
    \vspace{-0.35cm}
    \includegraphics[width=0.95\textwidth]{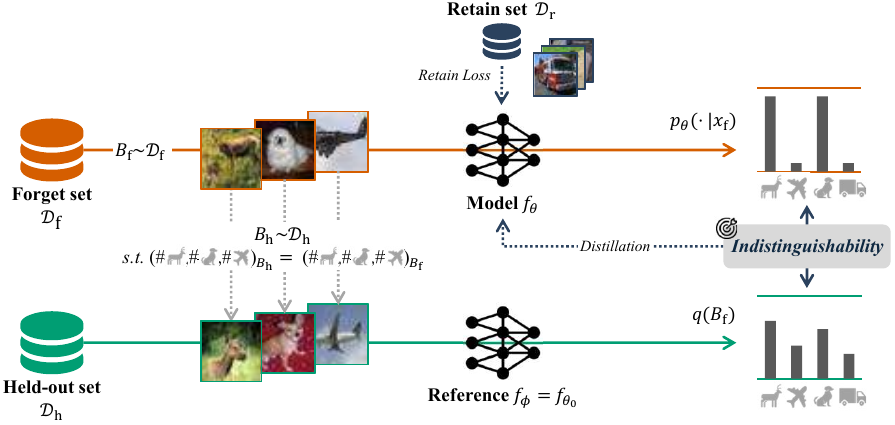}
    \vspace{-0.2cm}
     \caption{\textbf{Unlearning optimization in \method.} In \method, unlearning follows a reference-guided distillation objective based on \textit{indistinguishability}. Given a forget minibatch $B_{\mathrm{f}} \sim \mathcal{D}_{\mathrm{f}}$, \method samples a held-out minibatch $B_{\mathrm{h}} \sim \mathcal{D}_{\mathrm{h}}$ such that the class histogram of $B_{\mathrm{h}}$ matches that of $B_{\mathrm{f}}$. The samples in $B_{\mathrm{h}}$ are passed through a frozen reference model to construct a reference distribution $q$, which represents the desired ``unseen'' behavior for the forget examples. The current model $f_{\theta}$ is then updated by distilling its predictions on $B_{\mathrm{f}}$ toward this reference distribution, encouraging forget samples to become indistinguishable from non-member data. In parallel, a standard cross-entropy loss on retain samples is used to preserve model utility.\looseness-1}
    \label{fig:regun_visualization}
    \vspace*{-0.1cm}
\end{figure}

\vspace*{0.1cm}
\textbf{Forgetting should mean indistinguishability, not degradation.} Rather than merely making the model ``more wrong'', we argue that unlearning should align its behavior on forget data with that of truly unseen examples.
While this indistinguishability idea has been considered \citep{NecessAuditableAlgos}, a practical reference for ``unseen behavior'' remains missing.
We bridge this gap by proposing the paradigm of \textsc{Re}ference-\textsc{G}uided \textsc{Un}learning (\method)
\footnote{\label{fn:code} Code available at: \url{https://github.com/jmirlach/ReGUn}.} 
with held-out supervision. 
\method uses a disjoint held-out dataset as a stable proxy for ``unseen behavior'', aligning the model's outputs on forget examples with this reference distribution, 
as illustrated in~\Cref{fig:regun_visualization}. 
Crucially, this target is neither a uniform distribution, nor a random label, nor the original prediction of the model on the forget example itself. Instead, it is a model-induced distribution obtained by evaluating a frozen reference model on disjoint held-out samples with the same class composition as the forget batch. Thus, the forget examples are not pushed toward artificial uncertainty or corrupted supervision, but toward the average predictive behavior assigned to unseen samples from the relevant classes.

Prior reference-based methods supervise unlearning by replacing forget-sample outputs with pseudo-probabilities (e.g., uniform or global distributions)~\citep{PPU}, or by aligning them with generic third-party data (sometimes even unrelated noise), yielding a coarse, non-class-aware, and potentially noisy unlearning signal~\citep{FastYetVersatile}.
In contrast, \method uses a disjoint held-out subset as an explicit supervision source to construct the reference, enabling class-conditioned references rather than only marginal distribution matching. Finally, although a portion of the literature studies \textit{class}-level removal~\citep{zhang2025targeted, gandikota2023erasing}, we primarily focus on approximate \textit{instance}-level unlearning, which is more closely aligned with individual privacy and consent. This work targets the vision domain, building on the extensive body of literature~\citep{AMUN,hsuunseen,khalilcoun25,SALUN,leeruago} that focuses on unlearning for vision.

\paragraph{Contributions.} Our main contributions to the field of machine unlearning include:
\begin{itemize}[nosep, leftmargin=*]
\item We motivate and operationalize indistinguishability from unseen data as a direct optimization signal for approximate unlearning, rather than relying on objectives that mainly degrade forget-set performance.
\item We introduce \method, a reference-guided unlearning framework for vision that distills forget-set predictions toward class-conditioned distributions constructed from disjoint held-out data.
\item We empirically validate \method across multiple image-classification benchmarks, architectures, and forget fractions, showing improved forgetting--utility trade-offs over the studied baselines.\looseness-1
\end{itemize}

\section{Related Work}
\label{related_work}
\paragraph{Machine Unlearning.} Machine unlearning aims to remove the influence of specific training samples from a trained model while preserving utility on retained data. Early work focused primarily on exact unlearning methods with formal guarantees, including retraining-based approaches and data partitioning strategies such as SISA~\citep{cao2015towards,bourtoule2021machine,guo2020certified}. While theoretically appealing, these approaches remain computationally expensive and difficult to scale to modern deep learning settings.
As a result, recent research has increasingly focused on approximate unlearning methods~\citep{MULsurvey,MUNKEY}.
A dominant family of approaches performs post-hoc parameter updates designed to degrade model performance on forget samples. These include catastrophic-forgetting-based methods~\citep{goel2022towards}, gradient ascent approaches such as NegGrad and NegGrad+~\citep{SCRUB}, label perturbation strategies~\citep{graves2021amnesiac}, sparsity-based updates~\citep{l1sparse}, parameter dampening methods such as SSD~\citep{SSD}, saliency-guided editing~\citep{SALUN}, adversarial forgetting approaches~\citep{AMUN}, and contrastive formulations~\citep{khalilcoun25}.
Distillation-based approaches, including Bad Teacher~\citep{badteacher} and SCRUB~\citep{SCRUB}, attempt to repair utility during forgetting by transferring knowledge from auxiliary teacher models.\looseness-1

Despite their differences, these methods largely share a common paradigm: forgetting is typically encouraged through objectives that reduce, perturb, or disrupt model behavior on forget examples. However, degrading predictions does not necessarily imply that the resulting model behaves like one that never observed the forgotten data. In practice, these optimization signals can be poorly conditioned and may induce unnecessary changes to retained representations. Beyond standard instance-level unlearning, recent work has also studied related settings such as graph and recommendation unlearning~\citep{chen2022graph,chen2022recommendation}, incremental unlearning in pre-trained vision models~\citep{feng2025fg,kawamuraapproximate,wang2025machine}, and concept unlearning in generative or foundation models~\citep{gandikota2023erasing,gao2025meta,biswas2025cure,zhang2025targeted,dong2025machine}. While these directions broaden the scope of machine unlearning, our work primarily focuses on approximate instance-level unlearning for discriminative vision models, where the central challenge is balancing forgetting efficacy with retained utility.
\vspace*{-0.1cm}
\paragraph{Indistinguishability in Unlearning.}
A central objective of unlearning is to approximate the behavior of a model retrained without the forget set. Consequently, successful unlearning is often implicitly measured through indistinguishability: forget examples should become statistically indistinguishable from unseen data~\citep{thudi2022unrolling}, rather than from retained training samples. In practice, this principle is commonly evaluated through membership inference attacks (MIAs)~\citep{shokri2017membership,RMIA}, where successful forgetting corresponds to the inability to distinguish forget samples from non-members.
As an instantiation of this principle, statistical indistinguishability from a retrained model has been studied as a formal guarantee for machine unlearning~\citep{neel2021descent}. More recently, Datamodel Matching (DMM)~\citep{georgievattribute} estimates counterfactual retrained outputs through a data attribution framework. Other approximate methods pursue related goals through local smoothing~\citep{fosterinformation} or adversarial objectives~\citep{AMUN} that reduce sample-specific information. \method differs in the source and role of its supervision: rather than estimating retrained outputs or defining a local or adversarial target, it directly constructs a reference distribution from disjoint non-members drawn from the same data distribution.\looseness-1 

However, most approximate methods are not optimized for indistinguishability directly, creating a mismatch between the optimization objective and the privacy behavior ultimately being evaluated. \method aims to close this gap by directly optimizing toward unseen behavior using held-out references. A model may become less accurate on forget examples while still retaining detectable membership signals or residual information about the removed data. Recent work has increasingly highlighted this tension between forgetting strength and utility preservation~\citep{mavrothalassitis2026ascent,MUprogress}. Our work builds on this perspective and turns indistinguishability from an evaluation criterion into an optimization signal by distilling forget-set predictions toward held-out non-member behavior.\looseness-1

\paragraph{Reference Guidance.}
Several prior methods use auxiliary output distributions to guide unlearning. Pseudo-probability methods replace forget-set predictions with artificial targets, such as uniform probability distributions~\citep{PPU}. Others align forget-set outputs with generic third-party data, sometimes even unrelated noise, producing a marginal, non-class-aware forgetting signal~\citep{FastYetVersatile}.
Distillation-based methods use teacher-student objectives primarily to stabilize the forgetting process and preserve retain-set utility. In particular, Bad Teacher~\citep{badteacher} distills incorrect behavior on forget samples through an incompetent teacher, whereas SCRUB~\citep{SCRUB} combines forgetting objectives with distillation toward the original model on retained data. In contrast, \method uses same-dataset held-out samples to construct an explicit class-conditioned non-member target. This makes the target more specific: forget examples should behave like unseen samples from the same class composition, rather than being pushed toward uniform, generic third-party, noise-induced, or teacher-modified examples.\looseness-1

\section{\method: Reference-Guided Machine Unlearning}
\label{method}

\subsection{Problem Formulation}
We consider supervised $K$-class classification with input space $\mathcal{X}$ and labels $\mathcal{Y}=\{1,\dots,K\}$, and write $\mathcal{D}=\{(x_i,y_i)\}_{i=1}^{n}\subseteq\mathcal{X}\times\mathcal{Y}$ for a labeled dataset.
With $\Delta^K$ being the probability simplex over $K$ classes, let $f_\theta:\mathcal{X}\to\Delta^{K}$ be a probabilistic classifier with parameters $\theta$ and predictive distribution $p_\theta(\cdot\mid x)=f_\theta(x)$.
Machine unlearning starts with a model trained on $\mathcal{D}_{\mathrm{train}}$ and a request to remove the influence of a designated forget set $\mathcal{D}_{\mathrm{f}}\subset\mathcal{D}_{\mathrm{train}}$.
We write $\mathcal{D}_{\mathrm{train}}=\mathcal{D}_{\mathrm{r}}\cup\mathcal{D}_{\mathrm{f}}$ with $\mathcal{D}_{\mathrm{r}}\cap\mathcal{D}_{\mathrm{f}}=\emptyset$, where $\mathcal{D}_{\mathrm{r}}$ is the retain set.
Let $\theta_0$ be obtained by training on $\mathcal{D}_{\mathrm{train}}$, before any unlearning request is applied.
The goal of machine unlearning is to produce parameters $\theta_{\mathrm{u}}$ such that $f_{\theta_{\mathrm{u}}}$ behaves like the retraining baseline that would result if $\mathcal{D}_{\mathrm{f}}$ had never been seen.
Concretely, letting $\theta_{\mathrm{r}}$ denote parameters obtained by training on $\mathcal{D}_{\mathrm{r}}$ only from scratch, we aim for $f_{\theta_{\mathrm{u}}}\approx f_{\theta_{\mathrm{r}}}$, while avoiding full retraining. This approximation is empirical rather than certified and does not provide a formal guarantee of equivalence to retraining. Therefore, \method is framed as an approximate unlearning method.
Finally, we assume access to a disjoint held-out labeled set
$\mathcal{D}_{\mathrm{h}}=\{(x_j,y_j)\}_{j=1}^{n_{\mathrm{h}}}$ with $\mathcal{D}_{\mathrm{h}}\cap\mathcal{D}_{\mathrm{train}}=\emptyset$, which is used only to construct reference distributions during unlearning.\looseness-1

\subsection{Unlearning from Reference Distributions}

Rather than corrupting forget-set predictions, we aim to align them with the behavior corresponding to inputs never seen by the model. Ideally, this corresponds to matching the unavailable retrained model $f_{\theta_{\mathrm{r}}}$ on the forget set, i.e., minimizing $\mathbb{E}_{(x,y)\in\mathcal{D}_{\mathrm{f}}}\!\left[\mathrm{KL}\!\left(p_{\theta_{\mathrm{r}}}(\cdot\mid x)\,\|\,p_\theta(\cdot\mid x)\right)\right]$. Since $\theta_{\mathrm{r}}$ is precisely the retrain-from-scratch solution that approximate unlearning seeks to avoid, we instead use the held-out set $\mathcal{D}_{\mathrm{h}}$ to operationalize this ``unseen behavior'' and construct a reference from disjoint held-out inputs. We thus treat unlearning as distilling forget-set predictions toward this reference distribution, as detailed in~\Cref{alg:unlearning_main} and illustrated in~\Cref{fig:regun_visualization}. \looseness-1

\vspace*{-0.1cm}
\begin{algorithm}[h!]
\caption{\method: Reference-Guided Unlearning}
\label{alg:unlearning_main}
\begin{algorithmic}[1]
\State \textbf{Input:} initial model $f_{\theta_0}$, retain $\mathcal{D}_{\mathrm{r}}$, forget $\mathcal{D}_{\mathrm{f}}$, held-out $\mathcal{D}_{\mathrm{h}}$, weights $\lambda_{\mathrm{f}},\lambda_{\mathrm{r}}$, steps $T$, operator \textsc{RefDist}$_\mathrm{cls}$
\State Initialize $\theta \leftarrow \theta_0$
\State Set $f_\phi \leftarrow f_{\theta_0}$ \Comment{default; alternative $f_\phi$s ablated in Sec.~\ref{sec:results}.}
\For{$t=1,\dots,T$}
  \State Sample minibatches $B_{\mathrm{f}} \sim \mathcal{D}_{\mathrm{f}}$, $B_{\mathrm{r}} \sim \mathcal{D}_{\mathrm{r}}$
  \State $q(B_{\mathrm{f}}) \leftarrow \textsc{RefDist}_{\mathrm{cls}}(B_{\mathrm{f}};\mathcal{D}_{\mathrm{h}},f_{\phi})$\Comment{default; alternative \textsc{RefDist}s ablated in Sec.~\ref{sec:results}.}
  \State $\mathcal{L}_{\mathrm{f}} \leftarrow \frac{1}{|B_{\mathrm{f}}|}\sum_{(x,\cdot)\in B_{\mathrm{f}}}\mathrm{KL}\!\left(q(B_{\mathrm{f}})\middle\|\,p_\theta(\cdot\mid x)\right)$
  \State $\mathcal{L}_{\mathrm{r}} \leftarrow \frac{1}{|B_{\mathrm{r}}|}\sum_{(x,y)\in B_{\mathrm{r}}}\mathrm{CE}\!\left(p_\theta(\cdot\mid x),y\right)$
    \State $\mathcal{L} \leftarrow \lambda_{\mathrm{f}} \mathcal{L}_{\mathrm{f}} + \lambda_{\mathrm{r}} \mathcal{L}_{\mathrm{r}}$
  \State Update $\theta$ with one optimizer step on $\mathcal{L}$
\EndFor
\State $\theta_{\mathrm{u}} \leftarrow \theta$
\State \textbf{Output:} unlearned model $f_{\theta_{\mathrm{u}}}$
\end{algorithmic}
\end{algorithm}
\paragraph{Reference Distribution.} 
At each iteration in the unlearning phase, we randomly sample a forget minibatch $B_{\mathrm{f}} \sim \mathcal{D}_{\mathrm{f}}$ and compute a batch-level soft target as detailed in~\Cref{alg:refdist} via \textsc{RefDist}$_{\mathrm{cls}}$:
\begin{equation}
q(B_{\mathrm{f}})\;:=\;\textsc{RefDist}_{\mathrm{cls}}(B_{\mathrm{f}};\mathcal{D}_{\mathrm{h}},f_\phi)\in\Delta^K,
\end{equation}
where $f_\phi$ is a reference model. 
A natural choice is $f_\phi=f_{\theta_{\mathrm{r}}}$, because it is trained only on the retain set.
However, $f_{\theta_{\mathrm{r}}}$ is unavailable in approximate unlearning.
We therefore set $f_\phi=f_{\theta_0}$, the initial model state, as a proxy to avoid extra training and prevent reference drift. Importantly, although the reference model $f_{\theta_0}$ still retains influence from $\mathcal{D}_{\mathrm{f}}$, it is queried only on disjoint held-out inputs when constructing the reference.
This choice is motivated by the hypothesis that aggregation over these unseen inputs attenuates differences between the frozen-base and $f_{\theta_{\mathrm{r}}}$ teachers, yielding similar reference targets. \Cref{tab:teacher_target_similarity} tests and supports this hypothesis under random instance forgetting.
To construct the reference target $q(B_{\mathrm{f}})$, $\textsc{RefDist}_{\mathrm{cls}}$ samples a held-out batch $B_{\mathrm{h}} \sim \mathcal{D}_{\mathrm{h}}$ with the same class histogram as $B_{\mathrm{f}}$, and aggregates its predictions under $f_\phi$ into a single distribution.
 This controls for label-prior differences and yields a batch-level estimate of unseen behavior without matching forget examples to individual held-out counterparts. These defaults are not strict requirements for \method. In~\Cref{sec:results}, we ablate both components of the reference construction: the reference model $f_\phi$ and the selector \textsc{RefDist} used to sample $B_{\mathrm{h}}$. Specifically, we compare the frozen teacher $f_{\theta_0}$ against online $f_\theta$, its Exponential Moving Average (EMA) $f_{\mathrm{EMA}}$, and retrained $f_{\theta_{\mathrm{r}}}$ teachers, and compare class-matched sampling against uniform and feature-nearest-neighbor held-out selectors.\looseness-1

To aggregate the predictions from $B_{\mathrm{h}}$, we evaluate the reference model
$f_\phi$ on each held-out input and average the resulting predictive
distributions: \looseness-1
\vspace*{-0.1cm}
\begin{equation}
q(B_{\mathrm{f}})\;=\;
\frac{1}{|B_{\mathrm{h}}|}\sum_{(x,\cdot)\in B_{\mathrm{h}}} p_{\phi}(\cdot\mid x),
\end{equation}
which corresponds to an empirical mixture of the reference model's outputs on the selected held-out inputs. 
The same $q(B_{\mathrm{f}})$ is used for all $x\in B_{\mathrm{f}}$ and, notably, it depends on $B_{\mathrm{f}}$ through batch statistics, i.e., the label histogram, rather than through instance-wise matching. This batch-level construction avoids introducing noise from potentially brittle pairwise matching between forget and held-out examples. \Cref{alg:refdist} provides the step-by-step formulation of the reference distribution. 
Although \method can be viewed as non-member distillation, its target is not a hand-designed entropy prior such as a uniform distribution, random label, or label-smoothed target~\citep{muller2019does,badteacher}. Instead, it is estimated from reference-model predictions on disjoint examples, yielding a class-conditioned non-member target that preserves model-specific uncertainty while avoiding distillation from the original forget-sample predictions.\looseness-1

\begin{algorithm}[h!]
\caption{\textsc{RefDist}$_{\mathrm{cls}}$: Class-Conditioned Held-out Reference Distribution}
\label{alg:refdist}
\begin{algorithmic}[1]
\State \textbf{Input:} forget batch $B_{\mathrm{f}}$, held-out set $\mathcal{D}_{\mathrm{h}}$, reference model $f_{\phi}$
\State Let $c_k=\sum_{(\cdot, y)\in B_{\mathrm{f}}}\mathbf{1}[y=k]$ for $k\in\{1,\dots,K\}$
\State Let $\mathcal{D}_{\mathrm{h}}^{(k)}=\{(x,y)\in\mathcal{D}_{\mathrm{h}}:y=k\}$ for $k\in\{1,\dots,K\}$
\State Sample $B_{\mathrm{h}}$: draw $c_k$ examples uniformly from $\mathcal{D}_{\mathrm{h}}^{(k)}$ for each class $k$ (with replacement if needed)
\State Aggregate reference predictions: $q \leftarrow \frac{1}{|B_{\mathrm{h}}|}\sum_{(x,\cdot)\in B_{\mathrm{h}}} p_{\phi}(\cdot\mid x)$
\State \textbf{Output:} $q \in \Delta^K$
\end{algorithmic}
\end{algorithm}

\paragraph{Unlearning Objective.}
Given the reference distribution $q(B_{\mathrm{f}})$, unlearning is performed by optimizing two complementary terms. The forget term aligns the model predictions on forget inputs with the held-out reference distribution through a KL loss, encouraging forget examples to behave like non-member data. In parallel, a retain term anchors unlearning to preserve utility on retained data by applying standard cross-entropy on retain samples. For a retain minibatch $B_{\mathrm{r}} \sim \mathcal{D}_{\mathrm{r}}$, \method minimizes
\vspace*{-0.15cm}
\begin{equation}
\mathcal{L}(\theta;B_{\mathrm{f}},B_{\mathrm{r}})
\;=\;
\lambda_{\mathrm{f}}\,\frac{1}{|B_{\mathrm{f}}|}\sum_{(x,\cdot)\in B_{\mathrm{f}}}\mathrm{KL}\!\left(q(B_{\mathrm{f}})\,\middle\|\,p_\theta(\cdot\mid x)\right)
\;+\;
\lambda_{\mathrm{r}}\,\frac{1}{|B_{\mathrm{r}}|}\sum_{(x,y)\in B_{\mathrm{r}}}\mathrm{CE}\!\left(p_\theta(\cdot\mid x),y\right),
\end{equation}
where $\lambda_{\mathrm{f}},\lambda_{\mathrm{r}}>0$ trade off forgetting strength and retain utility, and $\mathrm{CE}(p,y)=-\log p(y)$ is the standard cross-entropy for a hard label $y$. Since $q(B_{\mathrm{f}})$ is fixed with respect to $\theta$ within each update, minimizing the forward $\mathrm{KL}(q(B_{\mathrm{f}})\|p_\theta(\cdot\mid x))$ is equivalent to minimizing the cross-entropy with the held-out soft target $q(B_{\mathrm{f}})$ up to an additive constant. Thus, the forget term can be interpreted as standard distillation toward a held-out teacher distribution.\looseness-1
\vspace*{0.2cm}
\vspace*{-0.1cm}
\section{Experimental Setup}
\label{experimental_setup}
\vspace*{-0.1cm}
\paragraph{Datasets.}
We evaluate \method on three standard image classification benchmarks~\citep{MUBox} to analyze the forgetting--utility trade-off: CIFAR-10, CIFAR-100~\citep{CIFAR}, and Tiny-ImageNet~\citep{TinyImagenet}. CIFAR-10 and CIFAR-100 each contain 60,000 $32\times32$ color images split into 50,000 training and 10,000 test images, with 10 and 100 classes, respectively. Tiny-ImageNet contains 100,000 training and 10,000 validation images, used for testing, at $64\times64$ resolution across 200 classes. For Transformer-based architectures, the images are upsampled to $224\times224$ resolution, whereas CNN baselines utilize the native dimensions. We outline additional implementation details and augmentation pipelines in \Cref{appdx:exp_details}.

Following our unlearning protocol, we split the original training set $\mathcal{D}_{\mathrm{orig}}$ into disjoint subsets using a seeded random split. We first reserve a validation split $\mathcal{D}_{\mathrm{val}}$ of size $0.1|\mathcal{D}_{\mathrm{orig}}|$ for hyperparameter selection. From the remaining pool $\mathcal{D}_{\mathrm{orig}}\setminus\mathcal{D}_{\mathrm{val}}$, we sample the forget set $\mathcal{D}_{\mathrm{f}}$ uniformly at random with \mbox{$|\mathcal{D}_{\mathrm{f}}| / |\mathcal{D}_{\mathrm{orig}}\setminus\mathcal{D}_{\mathrm{val}}| \in \{1\%, 10\%, 50\%\}$}, and reserve a held-out set $\mathcal{D}_{\mathrm{h}}$ of size $0.1|\mathcal{D}_{\mathrm{orig}}\setminus\mathcal{D}_{\mathrm{val}}|$. The remaining samples form the retain set $\mathcal{D}_{\mathrm{r}}$. The base model is trained from scratch on $\mathcal{D}_{\mathrm{train}}=\mathcal{D}_{\mathrm{r}}\cup\mathcal{D}_{\mathrm{f}}$ and then unlearned. We assume access to labels for the retained, forgotten, and held-out splits during unlearning. \looseness=-1
\vspace*{-0.1cm}
\paragraph{Baselines.}
We compare against a diverse set of approximate unlearning methods spanning gradient-ascent, fine-tuning, sparsification, learning-to-unlearn, parameter-dampening, saliency-based, adversarial, and distillation-based approaches. We include \textsc{NegGrad} and \textsc{NegGrad+}~\citep{SCRUB}, which apply gradient ascent on forget samples, with \textsc{NegGrad+} additionally using retain-set supervision. We evaluate \textsc{Finetune}, which continues training only on the retain set, and $\ell_1$-\textsc{sparse}~\citep{l1sparse}, which augments retain-set fine-tuning with an $\ell_1$ sparsity penalty on the model parameters. We further include \textsc{LUR}~\citep{LUR}, which uses a learning-to-unlearn objective to mitigate retain--forget gradient conflicts. Notably, for the Transformer-based setting, we use an adapted implementation described in~\Cref{appdx:exp_details}. We also compare against \textsc{SSD}~\citep{SSD}, an optimization-free Fisher-dampening method that suppresses parameters associated with forget samples. To cover localization-based unlearning, we include \textsc{SalUn}~\citep{SALUN}, which constructs gradient-based weight-saliency masks and restricts random-label unlearning updates to the salient parameters while preserving retain-set utility. We also include \textsc{Amun}~\citep{AMUN} as an adversarial unlearning baseline that uses adversarial examples during unlearning to enforce forgetting while maintaining utility on retained data. Finally, we include Bad Teacher (\textsc{BadTeach})~\citep{badteacher}, a teacher-guided distillation method for unlearning. Together, these baselines cover a broad range of representative approximate-unlearning strategies. For reference, we also report results for $f_{\theta_{\mathrm{r}}}$ (\textsc{Retrain}) and $f_{\theta_0}$ (\textsc{Base}). \looseness=-1
\vspace*{-0.1cm}
\paragraph{Models and Experimental Settings.}
In the main text, we report two model--dataset settings, covering both standard CNN-based image classification and a higher-resolution Transformer-based benchmark: ResNet-18~\citep{ResNet} on CIFAR-10, and Swin-T~\citep{swin} on Tiny-ImageNet~\citep{TinyImagenet}. ResNet-18 is a compact residual CNN with approximately 11M parameters, providing a standard convolutional backbone for low-resolution image classification. By contrast, Swin-T is a hierarchical vision Transformer with approximately 28M parameters based on shifted-window self-attention. Results on two additional settings with ResNet-18 on CIFAR-100 and ResNet-18 on Tiny-ImageNet are provided in \Cref{appdx:add_experiments_extra}. 
All training details, optimizer settings, and hyperparameter searches for \method and the studied baselines are reported in \Cref{appdx:exp_details}. \looseness=-1

\paragraph{Metrics.}
We evaluate unlearning along its two core objectives: utility preservation and forgetting efficacy. 
For utility, we report retain accuracy (RA) on $\mathcal{D}_{\mathrm{r}}$ and test accuracy (TA) on the test set. For forgetting, we report forget accuracy (FA) on $\mathcal{D}_{\mathrm{f}}$ and membership inference performance. For membership inference, we use an offline likelihood-ratio score based on robust MIA (RMIA)~\citep{RMIA}: For each candidate example $x$ with label $y$, we compute the log-ratio between the target model's true-label probability $p_\theta(y \mid x)$ and the average true-label probability assigned by four attack reference models trained on $\mathcal{D}_{\mathrm{r}}$ and therefore without the forget examples. We report the receiver operating characteristic area under the curve (ROC-AUC) of this attack score, denoted $\mathrm{RMIA}_{\mathrm{AUC}}$. Values close to 50\% indicate chance-level discrimination between forget samples and non-members, matching the indistinguishability criterion underlying unlearning.
To summarize performance across metrics, we follow existing work~\citep{LUR,SALUN} and report $\mathrm{Gap}_{\mathrm{Avg}}$. This metric averages the deviation from the retrain-from-scratch baseline (\textsc{Retrain}) across RA, TA, FA, and $\mathrm{RMIA}_{\mathrm{AUC}}$. Lower values indicate closer agreement with the desired \textsc{Retrain} version.
Further, to capture the utility-forgetting trade-off more directly, we treat TA and $\mathrm{RMIA}_{\mathrm{AUC}}$ as the primary representatives of utility and forgetting, respectively. We use these two metrics in the trade-off analysis and also report $\mathrm{Gap}^{\mathrm{F\text{-}U}}_{\mathrm{Avg}}$, defined as the average gap to \textsc{Retrain} over TA and $\mathrm{RMIA}_{\mathrm{AUC}}$.
Finally, we include a standard loss-based MIA ($\mathrm{SMIA}_{\mathrm{AUC}}$) and its corresponding $\mathrm{Gap}^{\mathrm{SMIA}}_{\mathrm{Avg}}$, defined as the same aggregate gap but replacing RMIA with SMIA. This complements the RMIA-based evaluation with a simpler loss-threshold diagnostic, closer in spirit to the MIA-style scores commonly used in unlearning evaluations~\citep{badteacher,SSD,SCRUB} and grounded in standard loss-based membership inference~\citep{MIA}. Note that, as with any suite of empirical attacks, these evaluations cannot certify the absence of residual membership leakage.
All results reported in this manuscript are averaged over three seeds and reported as mean~$\pm$~std, with all metrics expressed in \%. \looseness=-1

\section{Results}
\label{sec:results}
To evaluate whether indistinguishability-based supervision in \method provides an effective signal for approximate machine unlearning, we address five questions: 
\textit{(i) Unseen behavior as supervision:} Does a held-out reference distribution improve unlearning over degradation-based objectives?
\textit{(ii) Forgetting--utility trade-off:} Does \method better balance forgetting efficacy and retained utility?
\textit{(iii) Computational cost:} What overhead does reference-guided unlearning introduce?
\textit{(iv) Reference construction:} How do the teacher model and held-out selector affect performance?
\textit{(v) Held-out data dependence:} How sensitive is \method to the size of the held-out supervision set?

\paragraph{Unseen Behavior is a Strong Unlearning Signal.}

\Cref{tab:main-res,tab:main-res-swin} present our main findings across the two main-text settings and three forget fractions. Overall, \method achieves the strongest average agreement with the retrain-from-scratch baseline in most settings, indicating a favorable forgetting--utility trade-off. In the ResNet-18/CIFAR-10 setting, \method obtains the best $\mathrm{Gap}_{\mathrm{Avg}}$ in the 1\% and 50\% forgetting regimes and remains competitive at 10\%, where it is the runner-up. This improvement is mainly driven by reduced membership inference risk relative to the \textsc{Base} model, while maintaining competitive test and forget accuracy. Although \method is not always the best on each individual submetric, this is expected in a trade-off setting: different baselines favor different aspects of retrain matching, whereas \method is consistently among the closest methods in aggregate. The advantage becomes clearer in the higher-resolution Transformer setting with Swin-T on Tiny-ImageNet. Here, \method achieves the best $\mathrm{Gap}_{\mathrm{Avg}}$ in the 10\% and 50\% forgetting regimes and remains the runner-up at 1\%, where the forget signal is weaker and several methods perform similarly. More broadly, all approximate methods exhibit larger gaps to \textsc{Retrain} in the Swin-T setting than in the ResNet-18 setting, suggesting that Transformer-based unlearning remains a more challenging and comparatively underexplored regime. Among the baselines, performance varies substantially: specialized methods such as \textsc{Amun} and \textsc{SalUn} are often competitive, while simpler methods such as \textsc{Finetune} and \textsc{NegGrad+} can remain strong in individual metrics in some Transformer settings. Results on the two additional settings in~\Cref{appdx:add_experiments_extra} follow the same general trend: \method tends to perform best in aggregate, while the strongest method on individual metrics varies across datasets and forget fractions. Taken together, these results suggest that reference-guided distillation provides a strong and stable unlearning signal without relying on aggressive loss-ascent updates to approach \textsc{Retrain} behavior.
\paragraph{Understanding the Forgetting--Utility Trade-off.}
\Cref{fig:util_vs_forget_resc10_swin} visualizes the trade-off between retained utility and forgetting. The $y$-axis reports test accuracy, while the $x$-axis reports the gap between each method's $\mathrm{RMIA}_{\mathrm{AUC}}$ and the retrain-from-scratch baseline; therefore, points closer to the dashed vertical line and higher on the plot are preferable. For methods that admit an explicit retain--forget trade-off, we reparameterize the objective as \mbox{$(1-w)\cdot \mathcal{L}_{\mathrm{forget}} + w\cdot \mathcal{L}_{\mathrm{retain}}$} and sweep $w \in [0.1,0.9]$ to examine how each method behaves as the retain--forget balance is varied.
In the ResNet-18/CIFAR-10 setting, several methods occupy a similar plot region, consistent with the prior observation that this benchmark is close to saturated. The additional ResNet-18 results in~\Cref{fig:util_vs_forget_resc100_restiny} show a similar pattern on CIFAR-100, where several baselines reach comparable regions of the trade-off space, while the ResNet-18/Tiny-ImageNet setting shows a somewhat clearer advantage for \method. More prominently, the Swin-T/Tiny-ImageNet setting reveals the clearest separation: many baselines improve forgetting only at the cost of lower test accuracy, whereas \method maintains relatively stable utility while remaining close to the retrain $\mathrm{RMIA}_{\mathrm{AUC}}$ level. This suggests that reference-guided distillation provides a less destructive forgetting signal, improving the forgetting--utility trade-off without relying on aggressive loss-ascent behavior. In milder regimes, however, simpler methods remain competitive.\looseness-1

\input{inputs/plot_utility_forgetting_tradeoff_resnetcifar10_swinimagenet}

\paragraph{Computational Cost of \method{}.}
We briefly analyze \method{}'s runtime in terms of leading computational operations. A full method-wise comparison with empirical runtime results is given in Appendix~\ref{app:system_costs}. Let $C_G(b)$ denote the cost of one gradient update on a batch of size $b$, and let $C_F(b)$ denote the cost of a no-gradient forward pass. Each \method{} step performs one standard first-order update on the retain and forget minibatches of sizes $b_{\mathrm{r}}$ and $b_{\mathrm{f}}$, and additionally constructs the reference target from a held-out minibatch of size $b_{\mathrm{h}}$. Its cost over $T$ unlearning steps is
$\mathcal{T}_{\method} = T\left[C_G(b_{\mathrm{f}}+b_{\mathrm{r}}) + C_F(b_{\mathrm{h}}) + \mathcal{O}(b_{\mathrm{h}}K)\right]$,
with $b_{\mathrm{h}}=b_{\mathrm{f}}$ by default and $K$ denoting the number of classes. The first term is the standard gradient update through the current model, while the second term is the no-gradient reference-model forward pass used to construct the held-out target. The final term accounts for averaging the held-out probability vectors into the batch-level reference distribution.
Thus, relative to a standard retain--forget first-order update such as \textsc{NegGrad+}, the only additional leading operation is one no-gradient forward pass through the frozen reference model. If the reference predictions $p_\phi(\cdot\mid x)$ for $x\in\mathcal{D}_{\mathrm{h}}$ are precomputed, this overhead reduces to sampling and averaging cached $K$-dimensional probability vectors, with one-time preprocessing cost $C_F(n_{\mathrm{h}})$ and storage $\mathcal{O}(n_{\mathrm{h}}K)$. Overall, \method{} remains close to standard first-order unlearning methods in leading-order complexity: its main extra cost is reference construction, rather than importance-estimation passes, saliency-mask construction, adversarial-example generation, or inner-loop computation.

\begin{table}[h!]
\begin{center}
\small
\renewcommand{\arraystretch}{0.88}
\setlength{\tabcolsep}{1pt}
\caption{\textbf{Forgetting and utility results for ResNet-18 on CIFAR-10.}
Studied under 1\%, 10\%, and 50\% random forgetting. RA/FA/TA denote retain, forget, and test accuracy; $\mathrm{RMIA}_{\mathrm{AUC}}$ and $\mathrm{SMIA}_{\mathrm{AUC}}$ report membership inference performance, with 50\% corresponding to chance-level discrimination. Gap metrics measure deviation from \textsc{Retrain}, so lower is better. Results are mean~$\pm$~std over three seeds, with all values in percent. \textbf{Bold} and \second{underlined} denote best and second best, where ``best'' denotes smallest gap to \textsc{Retrain}.\looseness-1}
\label{tab:main-res}
\vspace*{0.2cm}
\begin{tabular}{@{} l *{4}{M} G *{3}{M} @{}}
\toprule
\addlinespace[0.3em]
& \multicolumn{8}{c}{\textbf{Forget 1\%}} \\
\addlinespace[0.3em]
\multicolumn{1}{c}{\small } & \multicolumn{1}{M}{\small RA} & \multicolumn{1}{M}{\small FA} & \multicolumn{1}{M}{\small TA} & \multicolumn{1}{M}{\small $\mathrm{RMIA}_{\mathrm{AUC}}$} & \multicolumn{1}{G}{\small $\mathrm{Gap}_{\mathrm{Avg}}$} & \multicolumn{1}{M}{\small $\mathrm{Gap}^{\mathrm{F\text{-}U}}_{\mathrm{Avg}}$} & \multicolumn{1}{M}{\small $\mathrm{SMIA}_{\mathrm{AUC}}$} & \multicolumn{1}{M}{\small $\mathrm{Gap}^{\mathrm{SMIA}}_{\mathrm{Avg}}$}\\
\addlinespace[0.2em]
\cmidrule(r){2-6}\cmidrule(l){7-9}

\textsc{Retrain} & \makebox[4.9em][c]{\makebox[2.75em][c]{$100.00$}\,\pmSmall{0.00}} & \makebox[4.9em][c]{\makebox[2.75em][c]{$94.22$}\,\pmSmall{1.24}} & \makebox[4.9em][c]{\makebox[2.75em][c]{$94.34$}\,\pmSmall{0.02}} & \makebox[4.9em][c]{\makebox[2.75em][c]{$49.98$}\,\pmSmall{1.26}} & \makebox[4.9em][c]{\makebox[2.75em][c]{$0.00$}\,\pmSmall{0.00}} & \makebox[4.9em][c]{\makebox[2.75em][c]{$0.00$}\,\pmSmall{0.00}} & \makebox[4.9em][c]{\makebox[2.75em][c]{$50.78$}\,\pmSmall{0.93}} & \makebox[4.9em][c]{\makebox[2.75em][c]{$0.00$}\,\pmSmall{0.00}} \\
\addlinespace[0.3em]
\textsc{Base} & \makebox[4.9em][c]{\makebox[2.75em][c]{$100.00$}\,\pmSmall{0.00}} & \makebox[4.9em][c]{\makebox[2.75em][c]{$100.00$}\,\pmSmall{0.00}} & \makebox[4.9em][c]{\makebox[2.75em][c]{$94.20$}\,\pmSmall{0.11}} & \makebox[4.9em][c]{\makebox[2.75em][c]{$60.11$}\,\pmSmall{1.30}} & \makebox[4.9em][c]{\makebox[2.75em][c]{$3.88$}\,\pmSmall{0.90}} & \makebox[4.9em][c]{\makebox[2.75em][c]{$5.14$}\,\pmSmall{0.93}} & \makebox[4.9em][c]{\makebox[2.75em][c]{$59.99$}\,\pmSmall{0.38}} & \makebox[4.9em][c]{\makebox[2.75em][c]{$3.78$}\,\pmSmall{0.40}} \\
\bottomrule
\end{tabular}
\end{center}
\end{table}

\begin{table}[h!]
\begin{center}
\small
\renewcommand{\arraystretch}{0.88}
\setlength{\tabcolsep}{1pt}
\caption{\textbf{Forgetting and utility results for Swin-T on Tiny-ImageNet.}
Studied under 1\%, 10\%, and 50\% random forgetting. RA/FA/TA denote retain, forget, and test accuracy; $\mathrm{RMIA}_{\mathrm{AUC}}$ and $\mathrm{SMIA}_{\mathrm{AUC}}$ report membership inference performance, with 50\% corresponding to chance-level discrimination. Gap metrics measure deviation from \textsc{Retrain}, so lower is better. Results are mean~$\pm$~std over three seeds, with all values in percent. \textbf{Bold} and \second{underlined} denote best and second best, where ``best'' denotes smallest gap to \textsc{Retrain}.\looseness-1}
\label{tab:main-res-swin}
\vspace*{0.2cm}
\begin{tabular}{@{} l *{4}{M} G *{3}{M} @{}}

\toprule
\addlinespace[0.3em]
& \multicolumn{8}{c}{\textbf{Forget 1\%}} \\
\addlinespace[0.3em]
\multicolumn{1}{c}{\small } & \multicolumn{1}{M}{\small RA} & \multicolumn{1}{M}{\small FA} & \multicolumn{1}{M}{\small TA} & \multicolumn{1}{M}{\small $\mathrm{RMIA}_{\mathrm{AUC}}$} & \multicolumn{1}{G}{\small $\mathrm{Gap}_{\mathrm{Avg}}$} & \multicolumn{1}{M}{\small $\mathrm{Gap}^{\mathrm{F\text{-}U}}_{\mathrm{Avg}}$} & \multicolumn{1}{M}{\small $\mathrm{SMIA}_{\mathrm{AUC}}$} & \multicolumn{1}{M}{\small $\mathrm{Gap}^{\mathrm{SMIA}}_{\mathrm{Avg}}$}\\
\addlinespace[0.2em]
\cmidrule(r){2-6}\cmidrule(l){7-9}

\textsc{Retrain} & \makebox[4.9em][c]{\makebox[2.75em][c]{$99.99$}\,\pmSmall{0.00}} & \makebox[4.9em][c]{\makebox[2.75em][c]{$61.48$}\,\pmSmall{1.23}} & \makebox[4.9em][c]{\makebox[2.75em][c]{$60.89$}\,\pmSmall{0.15}} & \makebox[4.9em][c]{\makebox[2.75em][c]{$49.79$}\,\pmSmall{1.38}} & \makebox[4.9em][c]{\makebox[2.75em][c]{$0.00$}\,\pmSmall{0.00}} & \makebox[4.9em][c]{\makebox[2.75em][c]{$0.00$}\,\pmSmall{0.00}} & \makebox[4.9em][c]{\makebox[2.75em][c]{$50.50$}\,\pmSmall{0.69}} & \makebox[4.9em][c]{\makebox[2.75em][c]{$0.00$}\,\pmSmall{0.00}} \\
\addlinespace[0.3em]
\textsc{Base} & \makebox[4.9em][c]{\makebox[2.75em][c]{$99.99$}\,\pmSmall{0.00}} & \makebox[4.9em][c]{\makebox[2.75em][c]{$99.96$}\,\pmSmall{0.06}} & \makebox[4.9em][c]{\makebox[2.75em][c]{$61.24$}\,\pmSmall{0.07}} & \makebox[4.9em][c]{\makebox[2.75em][c]{$87.77$}\,\pmSmall{0.18}} & \makebox[4.9em][c]{\makebox[2.75em][c]{$19.21$}\,\pmSmall{0.41}} & \makebox[4.9em][c]{\makebox[2.75em][c]{$19.15$}\,\pmSmall{0.69}} & \makebox[4.9em][c]{\makebox[2.75em][c]{$94.68$}\,\pmSmall{0.10}} & \makebox[4.9em][c]{\makebox[2.75em][c]{$20.75$}\,\pmSmall{0.36}} \\
\bottomrule
\end{tabular}
\end{center}
\vspace{0.5cm}
\end{table}

\begin{table}[h!]
\vspace{-0.3cm}
\begin{center}
\scriptsize
\renewcommand{\arraystretch}{1.15}
\setlength{\tabcolsep}{2pt}
\caption{\textbf{Teacher ablation results.} \textbf{Bold} and \second{underlined} denote best and second best, where ``best'' denotes smallest gap to \textsc{Retrain}.}
\label{tab:ablation_teacher}
\begin{adjustbox}{width=\linewidth}
\begin{tabular}{@{} l  c c >{\columncolor{gapcol}[\tabcolsep][\tabcolsep]}c @{\hspace{8pt}} c c >{\columncolor{gapcol}[\tabcolsep][\tabcolsep]}c @{\hspace{8pt}} c c >{\columncolor{gapcol}[\tabcolsep][0pt]}c @{}}
\toprule
& \multicolumn{3}{c}{\textbf{Forget 1\%}} & \multicolumn{3}{c}{\textbf{Forget 10\%}} & \multicolumn{3}{c}{\textbf{Forget 50\%}} \\
\addlinespace[0.5em]
\multicolumn{1}{c}{} & TA & $\mathrm{RMIA}_{\mathrm{AUC}}$ & $\mathrm{Gap}_{\mathrm{Avg}}$ & TA & $\mathrm{RMIA}_{\mathrm{AUC}}$ & $\mathrm{Gap}_{\mathrm{Avg}}$ & TA & $\mathrm{RMIA}_{\mathrm{AUC}}$ & $\mathrm{Gap}_{\mathrm{Avg}}$ \\
\midrule
\multicolumn{10}{l}{\textbf{ResNet-18 on CIFAR-10}} \\
\midrule
\textsc{Retrain}-Teacher & $91.95$ \pmSmall{0.03} & $51.08$ \pmSmall{1.20} & $1.70$ \pmSmall{0.65} & $\second{90.45}$ \pmSmall{0.63} & $\best{50.45}$ \pmSmall{0.91} & $\second{2.02}$ \pmSmall{0.15} & $\best{89.45}$ \pmSmall{0.61} & $52.19$ \pmSmall{0.12} & $\best{1.54}$ \pmSmall{0.26} \\
Online Student & $91.83$ \pmSmall{0.15} & $\best{50.53}$ \pmSmall{1.15} & $\best{1.47}$ \pmSmall{0.67} & $89.83$ \pmSmall{1.72} & $50.95$ \pmSmall{0.73} & $2.28$ \pmSmall{0.28} & $88.62$ \pmSmall{0.60} & $\second{52.13}$ \pmSmall{0.33} & $1.74$ \pmSmall{0.19} \\
EMA 0.99 & $\second{91.98}$ \pmSmall{0.09} & $50.78$ \pmSmall{1.07} & $1.56$ \pmSmall{0.62} & $89.43$ \pmSmall{1.62} & $\second{50.70}$ \pmSmall{0.85} & $2.41$ \pmSmall{0.41} & $\second{89.04}$ \pmSmall{0.77} & $52.18$ \pmSmall{0.05} & $\second{1.62}$ \pmSmall{0.07} \\
EMA 0.999 & $\second{91.99}$ \pmSmall{0.07} & $50.77$ \pmSmall{1.08} & $1.58$ \pmSmall{0.64} & $88.77$ \pmSmall{2.32} & $50.78$ \pmSmall{0.85} & $2.85$ \pmSmall{1.06} & $88.39$ \pmSmall{0.65} & $52.17$ \pmSmall{0.55} & $1.72$ \pmSmall{0.10} \\
\bottomrule
\end{tabular}
\end{adjustbox}
\end{center}
\vspace{-0.3cm}
\end{table}

\paragraph{The Effect of the Teacher: A Frozen Base Model is a Strong Reference Teacher.}
\method uses the original trained model $f_{\theta_0}$ as a frozen reference teacher to construct the held-out target distribution. A natural question is whether this simple choice is sufficient, or whether the reference should instead be produced by a teacher that evolves during unlearning or more closely approximates \textsc{Retrain}. We therefore compare the frozen-base teacher with four alternatives: the online student $f_\theta$, whose reference distribution changes throughout unlearning; two EMA teachers, $f_{\mathrm{EMA}}$ with decay factors $0.99$ and $0.999$; and the \textsc{Retrain}-Teacher $f_{\theta_{\mathrm{r}}}$, trained from scratch on $\mathcal{D}_{\mathrm{r}}$. The latter represents the ideal post-unlearning model, which is unavailable in realistic unlearning settings, but when queried on $\mathcal{D}_{\mathrm{h}}$ it still provides a batch-level held-out reference rather than the ideal per-example target $p_{\theta_{\mathrm r}}(\cdot\mid x_{\mathrm f})$.
\Cref{tab:ablation_teacher} shows that the frozen base model is already a strong reference teacher. Across both main settings, it is consistently competitive with online, EMA, and \textsc{Retrain}-Teacher. 
Given the observed seed-to-seed variability and the variation across configurations, the teacher variants exhibit broadly comparable overall performance. This is somewhat surprising, since one might expect the \textsc{Retrain}-Teacher to provide the most suitable reference distribution. To understand why this expected advantage does not consistently translate into better unlearning performance, we compare the aggregated reference targets generated by the frozen teacher and the \textsc{Retrain}-Teacher.\looseness-1 

We separate two questions: how strongly the frozen teacher and the \textsc{Retrain}-Teacher disagree on individual inputs, and how much of this disagreement remains in the aggregated target used by \method.
Define the pointwise disagreement as $\delta(x)=\mathrm{JS}\!\left(p_{\theta_0}(\cdot\mid x),p_{\theta_{\mathrm r}}(\cdot\mid x)\right)$, where $\mathrm{JS}$ is the Jensen–Shannon divergence. We define $d_{\mathrm f}=|\mathcal D_{\mathrm f}|^{-1}\sum_{x\in\mathcal D_{\mathrm f}}\delta(x)$ on the full forget set and $d_{\mathrm h}=\mathbb E\!\left[|B_{\mathrm h}|^{-1}\sum_{x\in B_{\mathrm h}}\delta(x)\right]$ over held-out batches $B_{\mathrm h}$ matched to a forget-batch class histogram. The ratio $d_{\mathrm f}/d_{\mathrm h}$ indicates whether the disagreement is concentrated on the forget set: values close to one indicate comparable disagreement on forget and held-out inputs, whereas values above one indicate stronger disagreement on forget inputs. 
For a held-out batch $B_{\mathrm h}$, let $\bar{p}_\phi(B_{\mathrm h})=|B_{\mathrm h}|^{-1}\sum_{x\in B_{\mathrm h}}p_\phi(\cdot\mid x)$. Using identical held-out inputs for both teachers, we define $d_{\mathrm{tgt}}=\mathbb E\!\left[\mathrm{JS}\!\left(\bar{p}_{\theta_0}(B_{\mathrm h}),\bar{p}_{\theta_{\mathrm r}}(B_{\mathrm h})\right)\right]$. This measures the target shift associated with changing between the frozen teacher and \textsc{Retrain}-Teacher while controlling for the sampled held-out inputs. The ratio $d_{\mathrm{tgt}}/d_{\mathrm h}$ quantifies how much of the pointwise held-out disagreement remains after aggregation, with smaller values indicating stronger attenuation.
As a reference scale, we independently sample a second held-out batch $B'_{\mathrm h}$ with the same class histogram as $B_{\mathrm h}$ and define $d_{\mathrm{samp}}=\mathbb E\!\left[\mathrm{JS}\!\left(\bar{p}_{\theta_{\mathrm r}}(B_{\mathrm h}),\bar{p}_{\theta_{\mathrm r}}(B'_{\mathrm h})\right)\right]$. The ratio $d_{\mathrm{tgt}}/d_{\mathrm{samp}}$ compares the teacher-induced target discrepancy with batch-to-batch variation under a fixed \textsc{Retrain}-Teacher. Values below one indicate that changing the teacher affects the target less than resampling the held-out batch. \looseness-1

The results in \Cref{tab:teacher_target_similarity} are consistent across all six configurations. Teacher disagreement is systematically larger on forget than on held-out inputs, particularly for Swin-T/Tiny-ImageNet and at smaller forget fractions. Aggregation strongly attenuates this discrepancy: only $2.6$--$11.5\%$ of the pointwise held-out disagreement remains, and the resulting shift is smaller than ordinary held-out resampling in every setting. Thus, although the frozen and retrained checkpoints remain distinct at the example level, they induce substantially more similar aggregated reference targets. This helps explain \Cref{tab:ablation_teacher} and supports the narrower conclusion that the frozen model is an adequate reference teacher for \method in the random-forgetting settings.
This does not imply general equivalence to the \textsc{Retrain}-Teacher. Under full-class forgetting, systematic class-level differences persist after aggregation, and the \textsc{Retrain}-Teacher provides a stronger reference, as seen in~\Cref{appdx:class_forgetting}.\looseness-1

\begin{table}[H]
\begin{center}
\small
\caption{\textbf{$\bm{f_{\theta_0}}$--$\bm{f_{\theta_{\mathrm r}}}$ target similarity under random forgetting.}
Target-space quantities use $1{,}000$ class-matched batches of size $128$ per seed. Ratios are formed per seed, and entries report means $\pm$ sample standard deviations across three seeds.}
\label{tab:teacher_target_similarity}
\renewcommand{\arraystretch}{1.15}
\setlength{\tabcolsep}{6pt}
\vspace*{0.2cm}
\resizebox{\linewidth}{!}{%
\begin{tabular}{@{} l l c c c c @{}}
\toprule
 & & \textbf{Pointwise contrast} & \multicolumn{3}{c}{\textbf{Aggregated target}} \\
\cmidrule(lr){3-3}\cmidrule(l){4-6}
& & $d_{\mathrm{f}}/d_{\mathrm{h}}$ & $d_{\mathrm{tgt}}$ (nats) & $d_{\mathrm{tgt}}/d_{\mathrm{h}}$ (\%) & $d_{\mathrm{tgt}}/d_{\mathrm{samp}}$ \\
\midrule
& Forget $1\%$  & $1.70$ \pmSmall{0.26} & $0.00056$ \pmSmall{0.00003} & $2.66$ \pmSmall{0.15} & $0.333$ \pmSmall{0.024} \\
ResNet-18 on CIFAR-10
& Forget $10\%$ & $1.46$ \pmSmall{0.11} & $0.00064$ \pmSmall{0.00006} & $2.63$ \pmSmall{0.13} & $0.354$ \pmSmall{0.023} \\
& Forget $50\%$ & $1.34$ \pmSmall{0.06} & $0.00113$ \pmSmall{0.00003} & $2.67$ \pmSmall{0.09} & $0.435$ \pmSmall{0.040} \\
\midrule
& Forget $1\%$  & $4.04$ \pmSmall{0.04} & $0.00689$ \pmSmall{0.00023} & $9.52$ \pmSmall{0.19} & $0.166$ \pmSmall{0.004} \\
Swin-T on Tiny-ImageNet
& Forget $10\%$ & $3.90$ \pmSmall{0.03} & $0.00755$ \pmSmall{0.00021} & $9.72$ \pmSmall{0.10} & $0.187$ \pmSmall{0.006} \\
& Forget $50\%$ & $2.66$ \pmSmall{0.01} & $0.01578$ \pmSmall{0.00011} & $11.49$ \pmSmall{0.07} & $0.410$ \pmSmall{0.012} \\
\bottomrule
\end{tabular}
}
\end{center}
\vspace{-0.25cm}
\end{table}

\begin{table}[h!]
\begin{center}
\scriptsize
\renewcommand{\arraystretch}{1.15}
\caption{\textbf{Selector ablation results.} \textbf{Bold} and \second{underlined} denote best and second best, where ``best'' denotes smallest gap to \textsc{Retrain}.}
\label{tab:class_cond}
\setlength{\tabcolsep}{2pt}
\begin{adjustbox}{width=\linewidth}
\begin{tabular}{@{} l  c c >{\columncolor{gapcol}[\tabcolsep][\tabcolsep]}c @{\hspace{8pt}} c c >{\columncolor{gapcol}[\tabcolsep][\tabcolsep]}c @{\hspace{8pt}} c c >{\columncolor{gapcol}[\tabcolsep][0pt]}c @{}}
\toprule
& \multicolumn{3}{c}{\textbf{Forget 1\%}} & \multicolumn{3}{c}{\textbf{Forget 10\%}} & \multicolumn{3}{c}{\textbf{Forget 50\%}} \\
\addlinespace[0.5em]
\multicolumn{1}{c}{} & TA & $\mathrm{RMIA}_{\mathrm{AUC}}$ & $\mathrm{Gap}_{\mathrm{Avg}}$ & TA & $\mathrm{RMIA}_{\mathrm{AUC}}$ & $\mathrm{Gap}_{\mathrm{Avg}}$ & TA & $\mathrm{RMIA}_{\mathrm{AUC}}$ & $\mathrm{Gap}_{\mathrm{Avg}}$ \\
\midrule
\multicolumn{10}{l}{\textbf{ResNet-18 on CIFAR-10}} \\
\midrule
Uniform & $\second{91.97}$ \pmSmall{0.06} & $\second{50.74}$ \pmSmall{0.93} & $\second{1.52}$ \pmSmall{0.31} & $88.93$ \pmSmall{3.30} & $\best{50.44}$ \pmSmall{0.69} & $2.90$ \pmSmall{1.57} & $88.62$ \pmSmall{1.01} & $\best{51.91}$ \pmSmall{0.38} & $\second{1.64}$ \pmSmall{0.22} \\
Feature-1NN & $91.95$ \pmSmall{0.15} & $50.80$ \pmSmall{0.92} & $1.55$ \pmSmall{0.50} & $\second{89.99}$ \pmSmall{1.47} & $\best{50.44}$ \pmSmall{0.68} & $\second{2.13}$ \pmSmall{0.27} & $\second{88.63}$ \pmSmall{0.14} & $\second{51.92}$ \pmSmall{0.16} & $\best{1.61}$ \pmSmall{0.12} \\
Class-Conditioned & $\best{92.15}$ \pmSmall{0.17} & $\best{50.71}$ \pmSmall{0.95} & $\best{1.49}$ \pmSmall{0.81} & $\best{90.63}$ \pmSmall{0.96} & $\second{50.46}$ \pmSmall{1.05} & $\best{1.99}$ \pmSmall{0.26} & $\best{88.81}$ \pmSmall{0.82} & $52.07$ \pmSmall{0.25} & $\second{1.64}$ \pmSmall{0.15} \\
\addlinespace[0.6em]
\midrule
\multicolumn{10}{l}{\textbf{Swin-T on Tiny-ImageNet}} \\
\midrule
Uniform & $50.83$ \pmSmall{0.48} & $\best{55.77}$ \pmSmall{0.61} & $\second{4.88}$ \pmSmall{0.15} & $52.36$ \pmSmall{0.54} & $\best{50.50}$ \pmSmall{0.73} & $\second{3.06}$ \pmSmall{0.35} & $44.88$ \pmSmall{0.77} & $47.42$ \pmSmall{0.18} & $1.82$ \pmSmall{0.41} \\
Feature-1NN & $\second{51.07}$ \pmSmall{0.81} & $56.56$ \pmSmall{2.06} & $5.35$ \pmSmall{0.29} & $\best{52.72}$ \pmSmall{0.83} & $50.95$ \pmSmall{0.83} & $3.33$ \pmSmall{0.35} & $\best{45.72}$ \pmSmall{0.50} & $\best{48.27}$ \pmSmall{0.36} & $\second{1.36}$ \pmSmall{0.18} \\
Class-Conditioned & $\best{51.23}$ \pmSmall{0.23} & $\second{56.07}$ \pmSmall{0.84} & $\best{4.68}$ \pmSmall{0.47} & $\second{52.66}$ \pmSmall{0.20} & $\second{49.88}$ \pmSmall{0.81} & $\best{3.00}$ \pmSmall{0.21} & $\second{45.64}$ \pmSmall{0.40} & $\second{47.86}$ \pmSmall{0.18} & $\best{1.34}$ \pmSmall{0.13} \\
\bottomrule
\end{tabular}
\end{adjustbox}
\end{center}
\vspace{-0.3cm}
\end{table}

\paragraph{Class-Conditioned References Improve Unlearning.}
\method constructs its reference distribution by selecting held-out samples from $\mathcal{D}_{\mathrm{h}}$. We compare the default class-conditioned selector, which matches the class histogram of the forget batch, against uniform sampling (\Cref{alg:refdist_uniform}) and feature-nearest-neighbor retrieval (\Cref{alg:refdist_feature_1nn}). As shown in~\Cref{tab:class_cond}, selector choice has a relatively small effect at lower forget fractions, where all strategies yield similar performance. However, class-conditioned sampling is consistently competitive and becomes more beneficial as the forget fraction increases. On ResNet-18/CIFAR-10, it achieves the best $\mathrm{Gap}_{\mathrm{Avg}}$ at 1\% and 10\% forgetting and remains comparable at 50\%. On Swin-T/Tiny-ImageNet, it obtains the best $\mathrm{Gap}_{\mathrm{Avg}}$ at 1\% and 50\% forgetting, and matches the best result at 10\%.
Overall, these results support the use of class-conditioned references as the default selector. Matching the held-out batch to the forget-batch class histogram preserves label-prior structure in the target distribution, yielding a more stable non-member reference than class-agnostic sampling, especially when stronger forgetting is required.\looseness-1

\paragraph{\method is Stable Across Held-Out Set Sizes.}
\begin{figure}[h!]
\vspace{-0.2cm}
\centering
\definecolor{col1}{RGB}{0,114,178}     
\definecolor{col10}{RGB}{213,94,0}     
\definecolor{col50}{RGB}{0,158,115}    

\pgfplotsset{
    forgetOne/.style={col1, thick, mark=o, mark size=2pt},
    forgetTen/.style={col10, thick, mark=o, mark size=2pt},
    forgetFifty/.style={col50, thick, mark=o, mark size=2pt},
    forgetShadeOne/.style={col1, fill=col1, fill opacity=0.14, draw=none},
    forgetShadeTen/.style={col10, fill=col10, fill opacity=0.14, draw=none},
    forgetShadeFifty/.style={col50, fill=col50, fill opacity=0.14, draw=none},
}

\noindent\hspace*{-24pt}
\begin{tikzpicture}

\begin{axis}[
    name=resnet,
    width=7.6cm,
    height=6cm,
    at={(0,0)},
    xmode=log,
    log basis x=10,
    y filter/.code={\pgfmathparse{#1*100}},
    ylabel={\small $\mathrm{Gap}_{\mathrm{Avg}}$},
    ylabel style={yshift=-3pt},
    title={\small \textbf{\textsc{ResNet-18 on CIFAR-10}}},
    label style={font=\small},
    tick label style={font=\scriptsize},
    title style={font=\small, yshift=-3pt},
    xmin=0.0008, xmax=1.25,
    ymin=0.9, ymax=3.2,
    xtick={0.001,0.01,0.1,1},
    xticklabels={$10^{-3}$,$10^{-2}$,$10^{-1}$,$1$},
    ytick={1.0,1.5,2.0,2.5, 3.0},
    scaled y ticks=false,
    grid=both,
    minor grid style={dashed, opacity=0.15},
    grid style={dashed, opacity=0.3},
    clip=true,
]

\addplot[name path=resnetFOneUpper, draw=none, forget plot] coordinates {(0.001,0.021575) (0.00316,0.025352) (0.01,0.025743) (0.0316,0.025654) (0.1,0.026273) (0.316,0.026471) (1,0.025682)};
\addplot[name path=resnetFOneLower, draw=none, forget plot] coordinates {(0.001,0.013105) (0.00316,0.010174) (0.01,0.011606) (0.0316,0.011176) (0.1,0.011423) (0.316,0.011288) (1,0.011507)};
\addplot[forgetShadeOne, forget plot] fill between[of=resnetFOneUpper and resnetFOneLower];
\addplot[forgetOne] coordinates {(0.001,0.017340) (0.00316,0.017763) (0.01,0.018674) (0.0316,0.018415) (0.1,0.018848) (0.316,0.018880) (1,0.018594)};

\addplot[name path=resnetFTenUpper, draw=none, forget plot] coordinates {(0.001,0.030562) (0.00316,0.024268) (0.01,0.023190) (0.0316,0.023315) (0.1,0.022553) (0.316,0.022430) (1,0.023321)};
\addplot[name path=resnetFTenLower, draw=none, forget plot] coordinates {(0.001,0.018746) (0.00316,0.017491) (0.01,0.019447) (0.0316,0.019344) (0.1,0.019054) (0.316,0.019492) (1,0.019068)};
\addplot[forgetShadeTen, forget plot] fill between[of=resnetFTenUpper and resnetFTenLower];
\addplot[forgetTen] coordinates {(0.001,0.024654) (0.00316,0.020880) (0.01,0.021318) (0.0316,0.021329) (0.1,0.020804) (0.316,0.020961) (1,0.021195)};

\addplot[name path=resnetFFiftyUpper, draw=none, forget plot] coordinates {(0.001,0.019181) (0.00316,0.016636) (0.01,0.016311) (0.0316,0.016933) (0.1,0.017823) (0.316,0.018256) (1,0.020641)};
\addplot[name path=resnetFFiftyLower, draw=none, forget plot] coordinates {(0.001,0.014938) (0.00316,0.015496) (0.01,0.013593) (0.0316,0.014492) (0.1,0.013376) (0.316,0.015195) (1,0.014730)};
\addplot[forgetShadeFifty, forget plot] fill between[of=resnetFFiftyUpper and resnetFFiftyLower];
\addplot[forgetFifty] coordinates {(0.001,0.017059) (0.00316,0.016066) (0.01,0.014952) (0.0316,0.015713) (0.1,0.015599) (0.316,0.016725) (1,0.017685)};

\end{axis}

\begin{axis}[
    name=swin,
    width=7.6cm,
    height=6cm,
    at={(resnet.north east)},
    anchor=north west,
    xshift=1.0cm,
    xmode=log,
    log basis x=10,
    y filter/.code={\pgfmathparse{#1*100}},
    title={\small \textbf{\textsc{Swin-T on Tiny-ImageNet}}},
    label style={font=\small},
    tick label style={font=\scriptsize},
    title style={font=\small, yshift=-3pt},
    xmin=0.0008, xmax=1.25,
    ymin=0, ymax=8,
    xtick={0.001,0.01,0.1,1},
    xticklabels={$10^{-3}$,$10^{-2}$,$10^{-1}$,$1$},
    ytick={0, 2.5, 5, 7.5},
    scaled y ticks=false,
    grid=both,
    minor grid style={dashed, opacity=0.15},
    grid style={dashed, opacity=0.3},
    clip=true,
    legend columns=1,
    legend style={
        draw=none,
        fill=none,
        at={(1.05,0.5)},
        anchor=west,
        font=\scriptsize,
        inner xsep=4pt,
        row sep=4pt,
        legend cell align={left},
    },
]

\addplot[name path=swinFOneUpper, draw=none, forget plot] coordinates {(0.001,0.070957) (0.00316,0.059262) (0.01,0.057203) (0.0316,0.054442) (0.1,0.054183) (0.316,0.053413) (1,0.052239)};
\addplot[name path=swinFOneLower, draw=none, forget plot] coordinates {(0.001,0.052717) (0.00316,0.046726) (0.01,0.042266) (0.0316,0.041343) (0.1,0.043048) (0.316,0.044011) (1,0.048461)};
\addplot[forgetShadeOne, forget plot] fill between[of=swinFOneUpper and swinFOneLower];
\addplot[forgetOne] coordinates {(0.001,0.061837) (0.00316,0.052994) (0.01,0.049734) (0.0316,0.047893) (0.1,0.048615) (0.316,0.048712) (1,0.050350)};
\addlegendentry{Forget $1\%$}

\addplot[name path=swinFTenUpper, draw=none, forget plot] coordinates {(0.001,0.065017) (0.00316,0.032883) (0.01,0.024646) (0.0316,0.030402) (0.1,0.030517) (0.316,0.031529) (1,0.032918)};
\addplot[name path=swinFTenLower, draw=none, forget plot] coordinates {(0.001,0.027228) (0.00316,0.025631) (0.01,0.022652) (0.0316,0.027325) (0.1,0.028756) (0.316,0.029071) (1,0.027530)};
\addplot[forgetShadeTen, forget plot] fill between[of=swinFTenUpper and swinFTenLower];
\addplot[forgetTen] coordinates {(0.001,0.046123) (0.00316,0.029257) (0.01,0.023649) (0.0316,0.028864) (0.1,0.029636) (0.316,0.030300) (1,0.030224)};
\addlegendentry{Forget $10\%$}

\addplot[name path=swinFFiftyUpper, draw=none, forget plot] coordinates {(0.001,0.067291) (0.00316,0.038256) (0.01,0.014845) (0.0316,0.016826) (0.1,0.016342) (0.316,0.015621) (1,0.015879)};
\addplot[name path=swinFFiftyLower, draw=none, forget plot] coordinates {(0.001,0.060172) (0.00316,0.020589) (0.01,0.012548) (0.0316,0.013539) (0.1,0.014818) (0.316,0.013778) (1,0.012599)};
\addplot[forgetShadeFifty, forget plot] fill between[of=swinFFiftyUpper and swinFFiftyLower];
\addplot[forgetFifty] coordinates {(0.001,0.063731) (0.00316,0.029422) (0.01,0.013697) (0.0316,0.015182) (0.1,0.015580) (0.316,0.014699) (1,0.014239)};
\addlegendentry{Forget $50\%$}

\end{axis}

\node[anchor=north, font=\small, yshift=-4mm, xshift=0.5cm] at (resnet.south east)
    {Held-out Subsample Fraction};

\end{tikzpicture}
\caption{\textbf{Held-out subsample ablation.}
$\mathrm{Gap}_{\mathrm{Avg}}$ is reported as a function of the fraction of the held-out split used to construct \method reference distributions. A fraction of $1$ uses the full held-out split, corresponding to $\approx 9\%$ of the original training set $\mathcal{D}_{\mathrm{orig}}$, or about 4,500 images for CIFAR-10 and 9,000 for Tiny-ImageNet. Lines show the mean across seeds; shaded regions indicate $\pm 1$ sample standard deviation.}
\label{fig:heldout_frac_average_gap_auc}
\end{figure}
We next analyze the sensitivity of \method to the amount of held-out supervision $\mathcal{D}_{\mathrm{h}}$ used to construct the reference distribution. Figure~\ref{fig:heldout_frac_average_gap_auc} reports $\mathrm{Gap}_{\mathrm{Avg}}$ when progressively subsampling $\mathcal{D}_{\mathrm{h}}$ on varying forgetting regimes. On ResNet-18/CIFAR-10, performance is highly stable across held-out fractions; the average gap varies only mildly, even when the held-out split is reduced by several orders of magnitude. This suggests that, in simpler low-class-count settings, a very small number of held-out non-member examples can already provide an effective reference signal.
 In contrast, the Swin-T/Tiny-ImageNet setting shows a clearer dependence on held-out diversity, especially at the smallest subsample fraction. However, even these highly reduced held-out subsets still yield meaningful unlearning behavior. Performance improves once the held-out fraction reaches $10^{-2}$ and then remains relatively stable as more held-out data is added, suggesting that only a modest amount of class coverage is needed to recover most of the benefit of the full held-out split. This behavior is consistent with the known data dependence of Transformer-based architectures, which typically require stronger supervision signals to maintain stable representations.
Overall, this ablation shows that \method does not rely on a large auxiliary dataset. Held-out data is used primarily to estimate class-conditioned non-member behavior, not to retrain the model. Small held-out subsets are already sufficient in simpler settings, while more diverse held-out coverage provides additional stability as the dataset, number of classes, and backbone complexity increase.\looseness-1
\section{Conclusion}
\label{conclusion}

We introduced \method, a machine unlearning framework for vision that reframes approximate unlearning from performance degradation to distributional matching. Instead of pushing the model to be wrong on forget examples via often poorly conditioned objectives, \method aligns the model's behavior on forget samples with its predictive distribution on disjoint held-out, non-member data. This design is motivated by the principle of indistinguishability: a truly unlearned model should treat the forget set as if it were a future unseen test set, precisely the property that many membership inference attacks aim to detect. Across diverse architectures, model scales, datasets, and forgetting regimes, our results show that held-out reference supervision is a simple yet effective unlearning signal. In \method, the frozen base model is used as a practical reference rather than as generally equivalent to the \textsc{Retrain}-Teacher, especially under full-class forgetting. In particular, \method achieves a favorable forgetting--utility trade-off, performs strongly under membership inference evaluation, and remains computationally on par with standard fine-tuning-based unlearning. Taken together, these findings suggest that approximate unlearning can benefit substantially from aligning the
optimization objective with the evaluation criterion: making forgotten data behave like non-member data. We show that simple procedures can outperform substantially more complex methods when optimized toward the right objective.

\paragraph{Limitations and Future Work.}

A practical consideration for \method is that it requires access to a disjoint held-out set, which may not always be available in real-world deployments. Nevertheless, our results suggest that \method can remain effective with very small held-out subsets and that labels are helpful but not strictly required, since uniform and feature-based selectors remain competitive. Such data could be reserved during data collection or obtained later from verified non-members, particularly in settings where new data continue to arrive. However, it still assumes access to a distribution sufficiently aligned with the training data. Relaxing this requirement is an interesting direction for future work, for example, by exploring references from external datasets, out-of-distribution data, or synthetic samples. Beyond this limitation, the flexibility of this framework is what opens several avenues for research.
While we focused primarily on class-conditioned references, future variants of \method could explore richer semantic conditioning strategies, such as adaptive instance-conditioned references or learned reference selection mechanisms. Beyond discriminative tasks, extending these principles to generative modeling remains a critical frontier, particularly for understanding how distributional matching objectives scale to high-dimensional foundation models. \method is not intended to provide a formal unlearning guarantee. Extending \method toward certified unlearning would require coupling the reference-guided objective with mechanisms that provide formal stability, privacy, or influence-removal guarantees. Rather, it is an approximate unlearning procedure designed to align forget-set behavior with held-out, non-member behavior. Its effectiveness is therefore assessed empirically, serving as a practical proxy for indistinguishability rather than formal certificates. Overall, our results show that even simple unlearning objectives can be highly effective when aligned with the right principle: making forgotten data behave like unseen data. We hope these findings encourage future research to prioritize indistinguishability as a core objective, even when developing more complex unlearning solutions.\looseness-1

\subsubsection*{Acknowledgements}
SL is supported by the Swiss State Secretariat for Education, Research and Innovation (SERI) under contract number MB22.00047. Computational data analysis was performed at Leonhard Med (\url{https://sis.id.ethz.ch/services/sensitiveresearchdata/}) secure trusted research environment at ETH Zurich.

\newpage
\bibliographystyle{tmlr}
\bibliography{references}

\newpage
\appendix
\label{appdx}
\section{Experimental Protocol and Computational Cost}
\label[appsec]{appdx:exp_setup_details}
\subsection{Implementation and Hyperparameter Details for \method{} and Baselines}
\label[appsec]{appdx:exp_details}

We summarize implementation details and hyperparameter search spaces and choices below. 

\textsc{Base \& Retrain}. For base training (from scratch) and retraining, we follow standard recipes.
For ResNet-18, we train for 100 epochs using SGD with momentum 0.9, learning rate 0.1, batch size 128, and apply standard data augmentation consisting of random crop with padding (32$\times$32 on CIFAR and 64$\times$64 on Tiny-ImageNet, pad=4), random horizontal flipping (p=0.5), and mild color jitter (brightness/contrast/saturation=0.1, hue=0.02).
For Swin-T, we train for 200 epochs using AdamW with learning rate $3\cdot 10^{-4}$ and a cosine schedule, batch size 128, and apply the standard data augmentation consisting of random horizontal flipping (p=0.5), and mild color jitter (brightness/contrast/saturation=0.1, hue=0.02) plus additional stronger augmentation in the form of RandAugment (N=2, M=9) and Random Erasing (p=0.25, area 2–20\%).
For ResNet-18, we use the native data resolutions of 32$\times$32 on CIFAR, and 64$\times$64 on Tiny-ImageNet.
For Swin-T on Tiny-ImageNet, we use 224$\times$224 resolution to align with the architecture’s default patch-based configuration.
\vspace{-0.06cm}

For unlearning (except \textsc{NegGrad}), we use a fixed budget of 10 epochs for ResNet-18 and 20 epochs for Swin-T. All other settings are kept the same as base training except that we omit the strong data augmentation in the Swin-T setup. The detailed configurations per method are the following:

\textsc{NegGrad}.
We run gradient ascent for 2 epochs and tune the learning rate.
ResNet-18: $\mathrm{lr}\in\{5\mathrm{e}{-2},\,1\mathrm{e}{-2},\,5\mathrm{e}{-3},\,1\mathrm{e}{-3}\}$.
Swin-T: $\mathrm{lr}\in\{1\mathrm{e}{-5},\,5\mathrm{e}{-6},\,1\mathrm{e}{-6},\,5\mathrm{e}{-7},\,1\mathrm{e}{-7},\,5\mathrm{e}{-8},\,1\mathrm{e}{-8}\}$.
\vspace{-0.06cm}

\textsc{NegGrad+}.
We combine a negative-gradient forget objective with a retain objective weighted by $w$. We tune the learning rate and the retain weight $w$.
ResNet-18: $\mathrm{lr}\in\{1\mathrm{e}{-1},\,5\mathrm{e}{-2},\,1\mathrm{e}{-2},\,5\mathrm{e}{-3}\}$, $w\in[0.8,\,0.99]$.
Swin-T: $\mathrm{lr}\in\{1\mathrm{e}{-3},\,5\mathrm{e}{-4},\,1\mathrm{e}{-4}\}$, $w\in[0.8,\,0.99]$.
\vspace{-0.06cm}

\textsc{Finetune}.
We fine-tune on the full retain set $\mathcal{D}_{\mathrm{r}}$ and tune the learning rate.
ResNet-18: $\mathrm{lr}\in\{1\mathrm{e}{-1},\,5\mathrm{e}{-2},\,1\mathrm{e}{-2},\,5\mathrm{e}{-3}\}$.
Swin-T: $\mathrm{lr}\in\{1\mathrm{e}{-3},\,5\mathrm{e}{-4},\,1\mathrm{e}{-4}\}$.
\vspace{-0.06cm}

$\ell_1$-\textsc{sparse}.
We implement sparsity-aware unlearning as fine-tuning on $\mathcal{D}_{\mathrm{r}}$ with an $\ell_1$ penalty weight $\gamma$. We tune the learning rate and $\gamma$.
ResNet-18: $\mathrm{lr}\in\{1\mathrm{e}{-1},\,5\mathrm{e}{-2},\,1\mathrm{e}{-2},\,5\mathrm{e}{-3}\}$, $\gamma\in[5\mathrm{e}{-6},\,5\mathrm{e}{-3}]$.
Swin-T: $\mathrm{lr}\in\{1\mathrm{e}{-3},\,5\mathrm{e}{-4},\,1\mathrm{e}{-4}\}$, $\gamma\in[5\mathrm{e}{-7},\,5\mathrm{e}{-5}]$.
\vspace{-0.06cm}

\textsc{LUR}.
We follow the original \textsc{LUR} setup and tune the learning rate and retain weight $w$. We use the LUR pruning/reinitialization step before unlearning, with batch-normalization parameters excluded from pruning, and fix the inner-step size to $\alpha=10^{-2}$.
ResNet-18: $\mathrm{lr}\in\{1\mathrm{e}{-1},\,5\mathrm{e}{-2},\,1\mathrm{e}{-2},\,5\mathrm{e}{-3}\}$, $w\in[0.1,\,0.99]$.
Swin-T: $\mathrm{lr}\in\{1\mathrm{e}{-3},\,5\mathrm{e}{-4},\,1\mathrm{e}{-4}\}$, $w\in[0.1,\,0.99]$.
Since full-model \textsc{LUR} is not computationally practical for larger models such as Swin-T at $224\times224$ resolution in our setup, for Swin-T experiments, we restrict unlearning and pruning to the last stage, final normalization layer, and classification head.
\vspace{-0.06cm}

\textsc{SSD}.
For ResNet-18, we use the recommended settings from the paper ($\alpha=10.0$, $\lambda=1.0$).
For Swin-T, we tune $\alpha$ and $\lambda$.
Swin-T: $\alpha\in[1,\,10]$, $\lambda\in[0.7,\,1.0]$.
Note that \textsc{SSD} is not designed for large random-forgetting regimes, which should be considered when interpreting the results.
\vspace{-0.06cm}

\textsc{SalUn}.
We fix the sparsity threshold to $50\%$ and tune the learning rate and a retain weight $w$ that balances the retain objective with the forgetting loss.
ResNet-18: $\mathrm{lr}\in\{1\mathrm{e}{-1},\,5\mathrm{e}{-2},\,1\mathrm{e}{-2},\,5\mathrm{e}{-3}\}$, $w\in[0.1,\,0.9]$.
Swin-T: $\mathrm{lr}\in\{1\mathrm{e}{-3},\,5\mathrm{e}{-4},\,1\mathrm{e}{-4}\}$, $w\in[0.1,\,0.9]$.
\vspace{-0.06cm}

\textsc{Amun}.
Let $\mathcal{D}_{\mathrm{A}} $ denote the adversarial examples generated from the forget data. We evaluate all four configurations described in the paper, corresponding to fine-tuning on $\mathcal{D}_{\mathrm{A}} \cup \mathcal{D}_{\mathrm{f}} \cup\mathcal{D}_{\mathrm{r}}$, $\mathcal{D}_{\mathrm{A}} \cup \mathcal{D}_{\mathrm{f}}$, $\mathcal{D}_{\mathrm{A}} \cup \mathcal{D}_{\mathrm{r}}$, and $\mathcal{D}_{\mathrm{A}}$ and tune the learning rate.
ResNet-18: $\mathrm{lr}\in\{1\mathrm{e}{-1},\,5\mathrm{e}{-2},\,1\mathrm{e}{-2},\,5\mathrm{e}{-3}\}$.
Swin-T: $\mathrm{lr}\in\{1\mathrm{e}{-3},\,5\mathrm{e}{-4},\,1\mathrm{e}{-4}\}$.
\vspace{-0.06cm}

\textsc{BadTeach}.
We follow the original \textsc{BadTeach} setup and tune the learning rate and retain-subset fraction $\rho_{\mathrm{BT}}$, where we unlearn on $\mathcal{D}_{\mathrm f}\cup\widetilde{\mathcal{D}}_{\mathrm r}$ with $|\widetilde{\mathcal{D}}_{\mathrm r}|=\lfloor\rho_{\mathrm{BT}}|\mathcal{D}_{\mathrm r}|\rfloor$.
ResNet-18: $\mathrm{lr}\in\{1\mathrm{e}{-1},\,5\mathrm{e}{-2}, \,1\mathrm{e}{-2},\,5\mathrm{e}{-3},\allowbreak\,1\mathrm{e}{-3},\,5\mathrm{e}{-4}\}$, $\rho_{\mathrm{BT}}\in[0.05,\,1]$.
Swin-T: $\mathrm{lr}\in\{1\mathrm{e}{-3},\,5\mathrm{e}{-4},\,1\mathrm{e}{-4},\,5\mathrm{e}{-5},\,1\mathrm{e}{-5}\}$, $\rho_{\mathrm{BT}}\in[0.05,\,1]$.
\vspace{-0.06cm}

\method.
We tune the learning rate and the retain/forget trade-off weight $w$ (corresponding to $\lambda_{\mathrm{r}} = w$ and $\lambda_{\mathrm{f}} = 1-w$ in the main objective).
ResNet-18: $\mathrm{lr}\in\{1\mathrm{e}{-1},\,5\mathrm{e}{-2},\,1\mathrm{e}{-2},\,5\mathrm{e}{-3}\}$, $w\in[0.1,\,0.9]$.
Swin-T: $\mathrm{lr}\in\{1\mathrm{e}{-3},\,5\mathrm{e}{-4},\,1\mathrm{e}{-4}\}$, $w\in[0.1,\,0.9]$.

To produce the results for \cref{fig:util_vs_forget_resc10_swin}, for each method, we fixed the best hyperparameter setting (excluding $w$) and then swept $w$. \textsc{NegGrad+}, \textsc{SalUn}, \textsc{LUR}, and \method inherently support this sweep.
For \textsc{Amun} and \textsc{BadTeach}, we rewrote the objective as
$(1-w)\mathcal{L}_{\mathrm{forget}} + w\mathcal{L}_{\mathrm{retain}}$:
for \textsc{Amun}, $\mathcal{L}_{\mathrm{forget}}$ is the loss on $\mathcal{D}_{\mathrm A}$ and $\mathcal{L}_{\mathrm{retain}}$ the loss on $\mathcal{D}_{\mathrm r}$;
for \textsc{BadTeach}, $\mathcal{L}_{\mathrm{forget}}$ is the distillation loss toward the incompetent teacher on $\mathcal{D}_{\mathrm f}$ and $\mathcal{L}_{\mathrm{retain}}$ the distillation loss toward the competent teacher on $\widetilde{\mathcal{D}}_{\mathrm r}$.

\subsection{Computational Cost and Runtime of \method and Baselines}
\label{app:system_costs}
In this section, we compare the compute structure and empirical runtime of the studied unlearning methods. The analysis considers one selected unlearning run after training, not an iterative stream of deletion requests where cached quantities could be amortized across requests.

\paragraph{Computational Cost Analysis.}
Let $T$ denote the number of unlearning steps, $n_{\mathrm{r}}=|\mathcal{D}_{\mathrm{r}}|$, $n_{\mathrm{f}}=|\mathcal{D}_{\mathrm{f}}|$, and $n_{\mathrm{h}}=|\mathcal{D}_{\mathrm{h}}|$. We write $b_{\mathrm{r}}=|B_{\mathrm{r}}|$, $b_{\mathrm{f}}=|B_{\mathrm{f}}|$, and $b_{\mathrm{h}}=|B_{\mathrm{h}}|$ for retain, forget, and held-out reference minibatches, with $b_{\mathrm{h}}=b_{\mathrm{f}}$ in the default \method{} implementation. Let $K$ be the number of classes and $P$ the number of model parameters. Finally, let $C_G(b)$ denote the cost of one gradient update on a batch of size $b$, including the forward pass, backward pass, loss computation, and optimizer update, and let $C_F(b)$ denote the cost of a no-gradient forward pass. Loss computations such as cross-entropy or KL terms are included in $C_G$ and are not counted separately.
Note that the number of unlearning steps $T$ should be interpreted together with the dataset that drives the update schedule. If a method is run for $E_{\mathrm{u}}$ unlearning epochs over a set of size $n_{\mathrm{s}}$ with batch size $b_{\mathrm{s}}$, then $T \approx E_{\mathrm{u}} \lceil n_{\mathrm{s}}/b_{\mathrm{s}}\rceil$. This matters because some methods naturally iterate over $\mathcal{D}_{\mathrm{f}}$, whereas retain-only fine-tuning may iterate over the much larger $\mathcal{D}_{\mathrm{r}}$. For comparability, in our evaluations all methods with explicit iterative updates use a forget-set-sized update budget, with retain minibatches sampled as needed. \looseness-1

\textsc{Finetune}, \textsc{NegGrad}, and \textsc{NegGrad+} mainly differ in which data enter an otherwise standard ``first-order'' update: retain batches, forget batches, or both retain and forget batches, respectively. Methods such as $\ell_1$-\textsc{sparse} follow the same first-order optimization path but add parameter-wise regularization, with $b_{\mathrm{upd}}$ denoting the batch size used by the corresponding update objective, typically $b_{\mathrm{upd}}=b_{\mathrm{r}}$. Its runtime is therefore also mainly driven by the dense forward/backward computation, with the additional parameter-wise regularization remaining comparatively small. Their costs can be summarized as
\[
\mathcal{T}_{\textsc{Finetune}} \approx T C_G(b_{\mathrm{r}}), \hspace{0.4em}
\mathcal{T}_{\textsc{NegGrad}} \approx T C_G(b_{\mathrm{f}}), \hspace{0.4em}
\mathcal{T}_{\textsc{NegGrad+}} \approx T C_G(b_{\mathrm{f}}+b_{\mathrm{r}}), \hspace{0.4em}
\mathcal{T}_{\ell_1\textsc{-sparse}} \approx T C_G(b_{\mathrm{upd}}) + \mathcal{O}(TP).
\]
The remaining baselines introduce method-specific stages. \textsc{SSD} estimates parameter-importance statistics, typically Fisher-based over $\mathcal{D}_{\mathrm{r}} \cup \mathcal{D}_{\mathrm{f}}$ and $\mathcal{D}_{\mathrm{f}}$ in the default single-request setting, and then applies a comparatively cheap parameter-wise dampening step, of cost $\mathcal{O}(P)$. 
The dominant cost comes from the importance-estimation stage, which can require expensive per-sample gradients. Let $S \subseteq \mathcal{D}_{\mathrm{train}}$ denote the dataset for importance scores; if the gradients are computed in minibatches of size $b$, we write
$\mathcal{T}_{\mathrm{Fisher}}(S)
\approx
\left\lceil \frac{|S|}{b} \right\rceil C_{\nabla}(b)$,
where $C_{\nabla}(b)$ is the cost of computing these gradients, without optimizer updates. The final SSD cost is
\[
\mathcal{T}_{\textsc{SSD}}
\approx
\mathcal{T}_{\mathrm{Fisher}}(\mathcal{D}_{\mathrm{r}} \cup \mathcal{D}_{\mathrm{f}})
+
\mathcal{T}_{\mathrm{Fisher}}(\mathcal{D}_{\mathrm{f}})
+
\mathcal{O}(P).
\]
\textsc{SalUn} similarly adds a preprocessing stage to construct a saliency mask before unlearning. The first term captures the one-time gradient-based mask-construction pass over forget data, while the second term is the subsequent retain--forget unlearning phase. Unless sparsity is exploited structurally, the mask changes which parameters are updated but does not substantially reduce the dense forward/backward cost of each update:
\[
\mathcal{T}_{\textsc{SalUn}} \approx \left\lceil \frac{n_{\mathrm{f}}}{b_{\mathrm{f}}} \right\rceil C_{\nabla}(b_{\mathrm{f}}) + T C_G(b_{\mathrm{f}}+b_{\mathrm{r}}).
\]
\textsc{Amun} adds a different source of overhead: adversarial-example generation for forget samples. Compared with standard first-order methods, the additional cost therefore comes primarily from constructing adversarial forget samples, not from the final unlearning update itself. Its adversarial cost $\mathcal{T}_{\mathrm{adv}}$ scales with the number of attack steps and the number of forget examples:
\[
\mathcal{T}_{\textsc{Amun}} \approx \mathcal{T}_{\mathrm{adv}} + T C_G(b_{\mathrm{upd}}).
\]
\textsc{LUR}, in contrast, introduces overhead through an inner-update or bilevel structure, which can make each unlearning step more expensive than a single standard first-order update. If $S_{\mathrm{in}}$ inner updates are used per outer step, its cost is at least
\[
\mathcal{T}_{\textsc{LUR}} \gtrsim T (S_{\mathrm{in}}+1) C_G(b_{\mathrm{upd}}).
\]
\textsc{BadTeach} runs first-order teacher--student distillation on shuffled batches from $\mathcal{D}_{\mathrm{f}}\cup\widetilde{\mathcal{D}}_{\mathrm{r}}$, where $\widetilde{\mathcal{D}}_{\mathrm{r}}\subseteq\mathcal{D}_{\mathrm{r}}$ is the selected retain subset and $b_{\mathrm{upd}}$ is the mixed minibatch size. Ignoring one-time teacher copy and initialization costs, each step uses two no-gradient teacher forward passes and one KL-based student update:
\[
\mathcal{T}_{\textsc{BadTeach}} \approx T\left[C_G(b_{\mathrm{upd}}) + 2C_F(b_{\mathrm{upd}})\right].
\]
Finally, \method{} follows the same first-order retain--forget update structure as \textsc{NegGrad+}, but replaces the loss-ascent forget target with a held-out reference distribution. At each iteration, \method{} samples $B_{\mathrm{f}}$ and $B_{\mathrm{r}}$, constructs $q(B_{\mathrm{f}})$ from a held-out batch $B_{\mathrm{h}}$ using the reference model $f_{\phi}$, and performs one gradient update on the combined objective. Its cost over $T$ unlearning steps is
\[
\mathcal{T}_{\method} = T \left[C_G(b_{\mathrm{f}}+b_{\mathrm{r}}) + C_F(b_{\mathrm{h}}) + \mathcal{O}(b_{\mathrm{h}}K) \right], 
\]
with $b_{\mathrm{h}}=b_{\mathrm{f}}$ by default. The three terms correspond to the standard current-model gradient update, a no-gradient reference-model forward pass, and the practically negligible averaging of held-out probability vectors into the batch-level reference distribution. Thus, relative to \textsc{NegGrad+}, \method{} essentially adds one no-gradient reference-model forward pass per unlearning step. With precomputed reference predictions $p_{\phi}(\cdot \mid x)$ for $x\in\mathcal{D}{\mathrm{h}}$, this overhead reduces to sampling and averaging cached $K$-dimensional vectors:
\[ \mathcal{T}_{\method}^{\mathrm{cached}} = T \left[C_G(b_{\mathrm{f}}+b_{\mathrm{r}}) +  \mathcal{O}(b_{\mathrm{h}}K) \right], \]
with one-time preprocessing cost $C_F(n_{\mathrm{h}})$ and storage $\mathcal{O}(n_{\mathrm{h}}K)$. Overall, \method{} remains close to standard first-order retain--forget unlearning methods in leading-order complexity. Its main extra cost is reference construction from held-out data, whereas other advanced baselines introduce importance-estimation passes, saliency-mask construction, adversarial-example generation, or inner-loop computation.

\paragraph{Empirical Runtime Results.}
Figure~\ref{fig:empirical_runtime} complements the theoretical cost analysis with empirical wall-clock runtimes for all studied methods. All measurements here, and in the full manuscript, were taken on a single NVIDIA GeForce RTX 4090 GPU. For iterative methods, one unlearning epoch corresponds to a forget-set-sized update budget; we use 10 such epochs for ResNet-18 and 20 for Swin-T, sampling retain minibatches as needed. We exclude all validation steps to isolate the unlearning phase cost. The results are indicative runtime comparisons rather than fully optimized method benchmarks.\looseness-1
\begin{figure}[h!]
\centering
\vspace{-0.05cm}
\definecolor{col1}{RGB}{0,114,178}     
\definecolor{col10}{RGB}{213,94,0}     
\definecolor{col50}{RGB}{0,158,115}    

\pgfplotsset{
    runtimeOne/.style={
        col1, thick, only marks, mark=|, mark size=4.2pt,
        mark options={col1, very thick},
        error bars/x dir=both,
        error bars/x explicit,
        error bars/error bar style={col1, very thick},
        error bars/error mark=|,
        error bars/error mark options={col1, thick, mark size=2.7pt},
        legend image code/.code={
            \draw[col1, very thick] (-0.24cm,0cm) -- (0.24cm,0cm);
            \draw[col1, thick] (-0.24cm,-0.055cm) -- (-0.24cm,0.055cm);
            \draw[col1, thick] (0.24cm,-0.055cm) -- (0.24cm,0.055cm);
            \draw[col1, very thick] (0cm,-0.095cm) -- (0cm,0.095cm);
        }
    },
    runtimeTen/.style={
        col10, thick, only marks, mark=|, mark size=4.2pt,
        mark options={col10, very thick},
        error bars/x dir=both,
        error bars/x explicit,
        error bars/error bar style={col10, very thick},
        error bars/error mark=|,
        error bars/error mark options={col10, thick, mark size=2.7pt},
        legend image code/.code={
            \draw[col10, very thick] (-0.24cm,0cm) -- (0.24cm,0cm);
            \draw[col10, thick] (-0.24cm,-0.055cm) -- (-0.24cm,0.055cm);
            \draw[col10, thick] (0.24cm,-0.055cm) -- (0.24cm,0.055cm);
            \draw[col10, very thick] (0cm,-0.095cm) -- (0cm,0.095cm);
        }
    },
    runtimeFifty/.style={
        col50, thick, only marks, mark=|, mark size=4.2pt,
        mark options={col50, very thick},
        error bars/x dir=both,
        error bars/x explicit,
        error bars/error bar style={col50, very thick},
        error bars/error mark=|,
        error bars/error mark options={col50, thick, mark size=2.7pt},
        legend image code/.code={
            \draw[col50, very thick] (-0.24cm,0cm) -- (0.24cm,0cm);
            \draw[col50, thick] (-0.24cm,-0.055cm) -- (-0.24cm,0.055cm);
            \draw[col50, thick] (0.24cm,-0.055cm) -- (0.24cm,0.055cm);
            \draw[col50, very thick] (0cm,-0.095cm) -- (0cm,0.095cm);
        }
    },
}

\noindent\hspace*{-4pt}
\begin{tikzpicture}

\begin{axis}[
    name=resnet,
    width=7.6cm,
    height=6cm,
    at={(0,0)},
    xmode=log,
    log basis x=10,
    xmin=0.35, xmax=12,
    ymin=0.4, ymax=10.6,
    y dir=reverse,
    title={\small \textbf{\textsc{ResNet-18 on CIFAR-10}}},
    label style={font=\small},
    tick label style={font=\small},
    title style={font=\small, yshift=-3pt},
    ytick={1,2,3,4,5,6,7,8,9,10},
    yticklabels={
    \textsc{NegGrad},
    \textsc{NegGrad+},
    \textsc{Finetune},
    $\ell_1$-\textsc{sparse},
    \textsc{LUR},
    \textsc{SSD},
    \textsc{SalUn},
    \textsc{Amun},
    \textsc{BadTeach},
    \textbf{\textsc{\method}}
    },
    xtick={0.5,1,2,5,10},
    xticklabels={$0.5$,$1$,$2$,$5$,$10$},
    grid=both,
    minor grid style={dashed, opacity=0.15},
    grid style={dashed, opacity=0.3},
    clip=false,
]

\addplot[runtimeOne]
table[x=runtime, y=method, x error minus=errminus, x error plus=errplus] {
runtime method errminus errplus
0.4500 1 0.0000 0.0583
0.5000 2 0.0250 0.0750
0.4667 3 0.0167 0.0167
0.4667 4 0.0167 0.0167
1.2333 5 0.0250 0.0583
1.7833 6 0.0833 0.0333
0.5167 7 0.0333 0.0500
1.3833 8 0.0583 0.0250
0.8167 9 0.0500 0.2250
1.2333 10 0.0750 0.0000
};

\addplot[runtimeTen]
table[x=runtime, y=method, x error minus=errminus, x error plus=errplus] {
runtime method errminus errplus
0.5333 1 0.0167 0.0333
0.6333 2 0.0167 0.0250
0.5167 3 0.0000 0.0417
0.5667 4 0.0167 0.0167
1.6333 5 0.0250 0.0333
1.9333 6 0.1500 0.0917
0.6833 7 0.0333 0.0500
2.9667 8 0.0250 0.0833
0.8667 9 0.0417 0.1750
1.3167 10 0.0250 0.0583
};

\addplot[runtimeFifty]
table[x=runtime, y=method, x error minus=errminus, x error plus=errplus] {
runtime method errminus errplus
0.8333 1 0.0250 0.0667
1.1333 2 0.0500 0.0500
0.7667 3 0.0167 0.0083
0.9167 4 0.0250 0.0417
3.0333 5 0.0333 0.0417
2.6500 6 0.1167 0.0667
1.2333 7 0.0250 0.0250
9.9500 8 0.2000 0.0583
1.1667 9 0.0333 0.1583
1.9667 10 0.0500 0.0667
};

\end{axis}

\begin{axis}[
    name=swin,
    width=7.6cm,
    height=6cm,
    at={(resnet.north east)},
    anchor=north west,
    xshift=0.1cm,
    xmode=log,
    log basis x=10,
    xmin=2.5, xmax=220,
    ymin=0.4, ymax=10.6,
    y dir=reverse,
    title={\small \textbf{\textsc{Swin-T on Tiny-ImageNet}}},
    label style={font=\small},
    tick label style={font=\scriptsize},
    title style={font=\small, yshift=-3pt},
    ytick={1,2,3,4,5,6,7,8,9,10},
    yticklabels=\empty,
    xtick={5,10,20,50,100,200},
    xticklabels={$5$,$10$,$20$,$50$,$100$,$200$},
    grid=both,
    minor grid style={dashed, opacity=0.15},
    grid style={dashed, opacity=0.3},
    clip=false,
    legend columns=1,
    legend style={
        draw=none,
        fill=none,
        at={(1.04,0.5)},
        anchor=west,
        font=\scriptsize,
        inner xsep=4pt,
        row sep=4pt,
        legend cell align={left},
    },
]

\addplot[runtimeOne]
table[x=runtime, y=method, x error minus=errminus, x error plus=errplus] {
runtime method errminus errplus
3.1333 1 0.0500 0.1167
5.4500 2 0.3833 0.5417
3.1833 3 0.0667 0.1000
3.1167 4 0.0167 0.1083
5.9500 5 0.5250 1.1167
6.6333 6 0.3250 0.2333
5.6500 7 0.6333 0.4167
39.8000 8 0.5917 0.3583
18.8500 9 0.1083 0.6583
5.6167 10 0.1667 0.7417
};
\addlegendentry{Forget $1\%$}

\addplot[runtimeTen]
table[x=runtime, y=method, x error minus=errminus, x error plus=errplus] {
runtime method errminus errplus
6.8833 1 0.1583 0.0750
14.1500 2 0.4833 0.4833
7.7333 3 0.9333 0.4583
8.0167 4 0.6250 0.0750
10.4667 5 0.4750 1.2333
7.4500 6 0.8917 0.2667
12.9333 7 0.5250 0.2917
65.4833 8 0.5917 0.2583
22.6500 9 0.2167 0.5750
13.8833 10 0.4417 0.3833
};
\addlegendentry{Forget $10\%$}

\addplot[runtimeFifty]
table[x=runtime, y=method, x error minus=errminus, x error plus=errplus] {
runtime method errminus errplus
22.2833 1 0.2500 0.0583
41.1000 2 0.1667 0.4083
18.4000 3 0.2417 0.2333
19.5167 4 0.2500 0.1833
25.0167 5 3.9000 0.4083
9.8667 6 0.3750 0.3000
36.1833 7 0.3833 0.1250
179.5833 8 1.6250 0.8500
38.8500 9 0.2667 0.1917
45.4500 10 0.4583 0.3333
};
\addlegendentry{Forget $50\%$}

\end{axis}

\node[anchor=north, font=\small, yshift=-4mm, xshift=0.4cm] at (resnet.south east)
    {Unlearning Runtime (minutes)};

\end{tikzpicture}
\vspace{-0.7cm}
\caption{\textbf{Empirical Runtime Results.} Wall-clock runtime of the studied unlearning methods for the unlearning phase. Central vertical bars show the median over 15 runs and horizontal whiskers indicate the interquartile range (25th--75th percentile).}
\label{fig:empirical_runtime}
\end{figure}

\newpage
\section{Supplementary Experimental Results}
\label[appsec]{appdx:add_experiments}

\subsection{Ablation Details and Results}
\label[appsec]{appdx:add_experiments_main}

We provide full details and extended results for the ablation studies from \Cref{sec:results}. 
To make the ablations directly comparable, we use a fixed hyperparameter setting for all variants within each model--dataset and forget-fraction configuration, rather than conducting an independent hyperparameter search for every ablated variant. This protocol reduces confounding from variant-specific tuning and highlights the effect of the design choice under study. Consequently, the reported ablation numbers may differ slightly from the main benchmark results, where hyperparameters are selected for the full method comparison.

The fixed settings are as follows. For ResNet-18 on CIFAR-10, we use retain weight $w=0.8$ and learning rate $0.005$ for both 1\% and 10\% forgetting, and $w=0.9$ with learning rate $0.05$ for 50\% forgetting. For Swin-T on Tiny-ImageNet, we use $w=0.9$ and learning rate $0.0005$ for 1\% forgetting, and $w=0.7$ with learning rate $0.0005$ for both 10\% and 50\% forgetting.

\Cref{tab:ablation_teacher_resnet_full,tab:ablation_teacher_swint_full} report the complete teacher ablation results, with the same metric suite as the main results. We define the two non-default selector variants used in the selector ablation, \textsc{RefDist}$_{\mathrm{uni}}$ in \Cref{alg:refdist_uniform} and \textsc{RefDist}$_{\mathrm{1NN}}$  in \Cref{alg:refdist_feature_1nn}, and report their full metric breakdown in \Cref{tab:ablation_selector_resnet,tab:ablation_selector_swint}. Finally, \Cref{tab:ablation_heldoutsize_resnet,tab:ablation_heldoutsize_swint} report the held-out-size ablation results, complementing the aggregate trends shown in \Cref{fig:heldout_frac_average_gap_auc}.

\vspace{0.5cm}
\begin{table}[h!]
\begin{center}
\small
\caption{\textbf{Teacher ablation full metrics for ResNet-18 on CIFAR-10.} \textbf{Bold} and \second{underlined} denote best and second best, where ``best'' denotes smallest gap to \textsc{Retrain}.}
\label{tab:ablation_teacher_resnet_full}
\vspace*{0.2cm}
\renewcommand{\arraystretch}{0.88}
\setlength{\tabcolsep}{1pt}
\begin{tabular}{@{} l *{4}{>{\centering\arraybackslash}m{4.95em}} >{\columncolor{gapcol}[\tabcolsep][\tabcolsep]\centering\arraybackslash}m{4.95em} *{3}{>{\centering\arraybackslash}m{4.95em}} @{}}

\toprule
\addlinespace[0.2em]
& \multicolumn{8}{c}{\textbf{Forget 1\%}} \\
\addlinespace[0.2em]
\multicolumn{1}{c}{\small } & \multicolumn{1}{>{\centering\arraybackslash}m{4.95em}}{\small RA} & \multicolumn{1}{>{\centering\arraybackslash}m{4.95em}}{\small FA} & \multicolumn{1}{>{\centering\arraybackslash}m{4.95em}}{\small TA} & \multicolumn{1}{>{\centering\arraybackslash}m{4.95em}}{\small $\mathrm{RMIA}_{\mathrm{AUC}}$} & \multicolumn{1}{>{\columncolor{gapcol}[\tabcolsep][0pt]\centering\arraybackslash}m{4.95em}}{\small $\mathrm{Gap}_{\mathrm{Avg}}$} & \multicolumn{1}{>{\centering\arraybackslash}m{4.95em}}{\small $\mathrm{Gap}^{\mathrm{F\text{-}U}}_{\mathrm{Avg}}$} & \multicolumn{1}{>{\centering\arraybackslash}m{4.95em}}{\small $\mathrm{SMIA}_{\mathrm{AUC}}$} & \multicolumn{1}{>{\centering\arraybackslash}m{4.95em}}{\small $\mathrm{Gap}^{\mathrm{SMIA}}_{\mathrm{Avg}}$}\\
\addlinespace[0.1em]
\cmidrule(r){2-6}\cmidrule(l){7-9}

\textsc{Retrain}-Teacher & \makebox[4.95em][c]{\makebox[2.85em][c]{$\second{99.77}$}\,\pmSmall{0.05}} & \makebox[4.95em][c]{\makebox[2.85em][c]{$97.93$}\,\pmSmall{0.51}} & \makebox[4.95em][c]{\makebox[2.85em][c]{$91.95$}\,\pmSmall{0.03}} & \makebox[4.95em][c]{\makebox[2.85em][c]{$51.08$}\,\pmSmall{1.20}} & \makebox[4.95em][c]{\makebox[2.85em][c]{$1.70$}\,\pmSmall{0.65}} & \makebox[4.95em][c]{\makebox[2.85em][c]{$1.88$}\,\pmSmall{0.69}} & \makebox[4.95em][c]{\makebox[2.85em][c]{$\best{48.92}$}\,\pmSmall{1.34}} & \makebox[4.95em][c]{\makebox[2.85em][c]{$2.05$}\,\pmSmall{0.53}} \\
\bottomrule
\end{tabular}
\end{center}
\end{table}

\clearpage
\begin{table}[h!]
\begin{center}
\small
\caption{\textbf{Teacher ablation full metrics for Swin-T on Tiny-ImageNet.} \textbf{Bold} and \second{underlined} denote best and second best, where ``best'' denotes smallest gap to \textsc{Retrain}.}
\label{tab:ablation_teacher_swint_full}
\vspace*{0.2cm}
\renewcommand{\arraystretch}{0.88}
\setlength{\tabcolsep}{1pt}
\begin{tabular}{@{} l *{4}{>{\centering\arraybackslash}m{4.95em}} >{\columncolor{gapcol}[\tabcolsep][\tabcolsep]\centering\arraybackslash}m{4.95em} *{3}{>{\centering\arraybackslash}m{4.95em}} @{}}

\toprule
\addlinespace[0.2em]
& \multicolumn{8}{c}{\textbf{Forget 1\%}} \\
\addlinespace[0.2em]
\multicolumn{1}{c}{\small } & \multicolumn{1}{>{\centering\arraybackslash}m{4.95em}}{\small RA} & \multicolumn{1}{>{\centering\arraybackslash}m{4.95em}}{\small FA} & \multicolumn{1}{>{\centering\arraybackslash}m{4.95em}}{\small TA} & \multicolumn{1}{>{\centering\arraybackslash}m{4.95em}}{\small $\mathrm{RMIA}_{\mathrm{AUC}}$} & \multicolumn{1}{>{\columncolor{gapcol}[\tabcolsep][0pt]\centering\arraybackslash}m{4.95em}}{\small $\mathrm{Gap}_{\mathrm{Avg}}$} & \multicolumn{1}{>{\centering\arraybackslash}m{4.95em}}{\small $\mathrm{Gap}^{\mathrm{F\text{-}U}}_{\mathrm{Avg}}$} & \multicolumn{1}{>{\centering\arraybackslash}m{4.95em}}{\small $\mathrm{SMIA}_{\mathrm{AUC}}$} & \multicolumn{1}{>{\centering\arraybackslash}m{4.95em}}{\small $\mathrm{Gap}^{\mathrm{SMIA}}_{\mathrm{Avg}}$}\\
\addlinespace[0.1em]
\cmidrule(r){2-6}\cmidrule(l){7-9}

\textsc{Retrain}-Teacher & \makebox[4.95em][c]{\makebox[2.85em][c]{$\best{98.05}$}\,\pmSmall{0.21}} & \makebox[4.95em][c]{\makebox[2.85em][c]{$63.59$}\,\pmSmall{4.28}} & \makebox[4.95em][c]{\makebox[2.85em][c]{$51.24$}\,\pmSmall{1.03}} & \makebox[4.95em][c]{\makebox[2.85em][c]{$57.24$}\,\pmSmall{1.67}} & \makebox[4.95em][c]{\makebox[2.85em][c]{$5.76$}\,\pmSmall{0.42}} & \makebox[4.95em][c]{\makebox[2.85em][c]{$8.54$}\,\pmSmall{0.47}} & \makebox[4.95em][c]{\makebox[2.85em][c]{$\best{48.38}$}\,\pmSmall{0.52}} & \makebox[4.95em][c]{\makebox[2.85em][c]{$3.96$}\,\pmSmall{1.16}} \\
\bottomrule
\end{tabular}
\end{center}
\end{table}

\vspace*{0.5cm}
\begin{table}[h!]
\begin{center}
\small
\caption{\textbf{Selector ablation full metrics for ResNet-18 on CIFAR-10.} \textbf{Bold} and \second{underlined} denote best and second best, where ``best'' denotes smallest gap to \textsc{Retrain}.}
\label{tab:ablation_selector_resnet}
\vspace*{0.2cm}
\renewcommand{\arraystretch}{0.88}
\setlength{\tabcolsep}{1pt}
\begin{tabular}{@{} l *{4}{>{\centering\arraybackslash}m{4.95em}} >{\columncolor{gapcol}[\tabcolsep][\tabcolsep]\centering\arraybackslash}m{4.95em} *{3}{>{\centering\arraybackslash}m{4.95em}} @{}}

\toprule
\addlinespace[0.2em]
& \multicolumn{8}{c}{\textbf{Forget 1\%}} \\
\addlinespace[0.2em]
\multicolumn{1}{c}{\small } & \multicolumn{1}{>{\centering\arraybackslash}m{4.95em}}{\small RA} & \multicolumn{1}{>{\centering\arraybackslash}m{4.95em}}{\small FA} & \multicolumn{1}{>{\centering\arraybackslash}m{4.95em}}{\small TA} & \multicolumn{1}{>{\centering\arraybackslash}m{4.95em}}{\small $\mathrm{RMIA}_{\mathrm{AUC}}$} & \multicolumn{1}{>{\columncolor{gapcol}[\tabcolsep][0pt]\centering\arraybackslash}m{4.95em}}{\small $\mathrm{Gap}_{\mathrm{Avg}}$} & \multicolumn{1}{>{\centering\arraybackslash}m{4.95em}}{\small $\mathrm{Gap}^{\mathrm{F\text{-}U}}_{\mathrm{Avg}}$} & \multicolumn{1}{>{\centering\arraybackslash}m{4.95em}}{\small $\mathrm{SMIA}_{\mathrm{AUC}}$} & \multicolumn{1}{>{\centering\arraybackslash}m{4.95em}}{\small $\mathrm{Gap}^{\mathrm{SMIA}}_{\mathrm{Avg}}$}\\
\addlinespace[0.1em]
\cmidrule(r){2-6}\cmidrule(l){7-9}

Uniform & \makebox[4.95em][c]{\makebox[2.85em][c]{$99.70$}\,\pmSmall{0.08}} & \makebox[4.95em][c]{\makebox[2.85em][c]{$\best{97.48}$}\,\pmSmall{0.90}} & \makebox[4.95em][c]{\makebox[2.85em][c]{$\second{91.97}$}\,\pmSmall{0.06}} & \makebox[4.95em][c]{\makebox[2.85em][c]{$\second{50.74}$}\,\pmSmall{0.93}} & \makebox[4.95em][c]{\makebox[2.85em][c]{$\second{1.52}$}\,\pmSmall{0.31}} & \makebox[4.95em][c]{\makebox[2.85em][c]{$\second{1.70}$}\,\pmSmall{0.25}} & \makebox[4.95em][c]{\makebox[2.85em][c]{$\second{48.60}$}\,\pmSmall{1.06}} & \makebox[4.95em][c]{\makebox[2.85em][c]{$\best{2.03}$}\,\pmSmall{0.52}} \\
\bottomrule
\end{tabular}
\end{center}
\end{table}

\begin{table}[h!]
\begin{center}
\small
\caption{\textbf{Selector ablation full metrics for Swin-T on Tiny-ImageNet.} \textbf{Bold} and \second{underlined} denote best and second best, where ``best'' denotes smallest gap to \textsc{Retrain}.}
\label{tab:ablation_selector_swint}
\vspace*{0.2cm}
\renewcommand{\arraystretch}{0.88}
\setlength{\tabcolsep}{1pt}
\begin{tabular}{@{} l *{4}{>{\centering\arraybackslash}m{4.95em}} >{\columncolor{gapcol}[\tabcolsep][\tabcolsep]\centering\arraybackslash}m{4.95em} *{3}{>{\centering\arraybackslash}m{4.95em}} @{}}

\toprule
\addlinespace[0.2em]
& \multicolumn{8}{c}{\textbf{Forget 1\%}} \\
\addlinespace[0.2em]
\multicolumn{1}{c}{\small } & \multicolumn{1}{>{\centering\arraybackslash}m{4.95em}}{\small RA} & \multicolumn{1}{>{\centering\arraybackslash}m{4.95em}}{\small FA} & \multicolumn{1}{>{\centering\arraybackslash}m{4.95em}}{\small TA} & \multicolumn{1}{>{\centering\arraybackslash}m{4.95em}}{\small $\mathrm{RMIA}_{\mathrm{AUC}}$} & \multicolumn{1}{>{\columncolor{gapcol}[\tabcolsep][0pt]\centering\arraybackslash}m{4.95em}}{\small $\mathrm{Gap}_{\mathrm{Avg}}$} & \multicolumn{1}{>{\centering\arraybackslash}m{4.95em}}{\small $\mathrm{Gap}^{\mathrm{F\text{-}U}}_{\mathrm{Avg}}$} & \multicolumn{1}{>{\centering\arraybackslash}m{4.95em}}{\small $\mathrm{SMIA}_{\mathrm{AUC}}$} & \multicolumn{1}{>{\centering\arraybackslash}m{4.95em}}{\small $\mathrm{Gap}^{\mathrm{SMIA}}_{\mathrm{Avg}}$}\\
\addlinespace[0.1em]
\cmidrule(r){2-6}\cmidrule(l){7-9}

Uniform & \makebox[4.95em][c]{\makebox[2.85em][c]{$\second{97.90}$}\,\pmSmall{0.30}} & \makebox[4.95em][c]{\makebox[2.85em][c]{$60.63$}\,\pmSmall{2.78}} & \makebox[4.95em][c]{\makebox[2.85em][c]{$50.83$}\,\pmSmall{0.48}} & \makebox[4.95em][c]{\makebox[2.85em][c]{$\best{55.77}$}\,\pmSmall{0.61}} & \makebox[4.95em][c]{\makebox[2.85em][c]{$\second{4.88}$}\,\pmSmall{0.15}} & \makebox[4.95em][c]{\makebox[2.85em][c]{$\second{8.01}$}\,\pmSmall{0.70}} & \makebox[4.95em][c]{\makebox[2.85em][c]{$47.51$}\,\pmSmall{0.46}} & \makebox[4.95em][c]{\makebox[2.85em][c]{$4.00$}\,\pmSmall{0.80}} \\
\bottomrule
\end{tabular}
\end{center}
\end{table}

\begin{table}[h!]
\vspace*{0.5cm}
\begin{center}
\small
\caption{\textbf{Held-out subsample ablation full metrics for ResNet-18 on CIFAR-10.} \textbf{Bold} and \second{underlined} denote best and second best, where ``best'' denotes smallest gap to \textsc{Retrain}.}
\label{tab:ablation_heldoutsize_resnet}
\vspace*{0.2cm}
\renewcommand{\arraystretch}{0.88}
\setlength{\tabcolsep}{1pt}
\begin{tabular}{@{} l *{4}{>{\centering\arraybackslash}m{4.95em}} >{\columncolor{gapcol}[\tabcolsep][\tabcolsep]\centering\arraybackslash}m{4.95em} *{3}{>{\centering\arraybackslash}m{4.95em}} @{}}

\toprule
\addlinespace[0.2em]
& \multicolumn{8}{c}{\textbf{Forget 1\%}} \\
\addlinespace[0.2em]
\multicolumn{1}{c}{\small } & \multicolumn{1}{>{\centering\arraybackslash}m{4.95em}}{\small RA} & \multicolumn{1}{>{\centering\arraybackslash}m{4.95em}}{\small FA} & \multicolumn{1}{>{\centering\arraybackslash}m{4.95em}}{\small TA} & \multicolumn{1}{>{\centering\arraybackslash}m{4.95em}}{\small $\mathrm{RMIA}_{\mathrm{AUC}}$} & \multicolumn{1}{>{\columncolor{gapcol}[\tabcolsep][0pt]\centering\arraybackslash}m{4.95em}}{\small $\mathrm{Gap}_{\mathrm{Avg}}$} & \multicolumn{1}{>{\centering\arraybackslash}m{4.95em}}{\small $\mathrm{Gap}^{\mathrm{F\text{-}U}}_{\mathrm{Avg}}$} & \multicolumn{1}{>{\centering\arraybackslash}m{4.95em}}{\small $\mathrm{SMIA}_{\mathrm{AUC}}$} & \multicolumn{1}{>{\centering\arraybackslash}m{4.95em}}{\small $\mathrm{Gap}^{\mathrm{SMIA}}_{\mathrm{Avg}}$}\\
\addlinespace[0.1em]
\cmidrule(r){2-6}\cmidrule(l){7-9}

Held-out $0.1\%$ & \makebox[4.95em][c]{\makebox[2.85em][c]{$99.54$}\,\pmSmall{0.16}} & \makebox[4.95em][c]{\makebox[2.85em][c]{$95.78$}\,\pmSmall{1.39}} & \makebox[4.95em][c]{\makebox[2.85em][c]{$91.73$}\,\pmSmall{0.14}} & \makebox[4.95em][c]{\makebox[2.85em][c]{$\best{47.60}$}\,\pmSmall{1.40}} & \makebox[4.95em][c]{\makebox[2.85em][c]{$\best{1.73}$}\,\pmSmall{0.42}} & \makebox[4.95em][c]{\makebox[2.85em][c]{$\best{2.24}$}\,\pmSmall{0.77}} & \makebox[4.95em][c]{\makebox[2.85em][c]{$\best{46.30}$}\,\pmSmall{0.81}} & \makebox[4.95em][c]{\makebox[2.85em][c]{$\second{2.28}$}\,\pmSmall{0.56}} \\
\bottomrule
\end{tabular}
\end{center}
\end{table}

\begin{table}[h!]
\begin{center}
\small
\caption{\textbf{Held-out subsample ablation full metrics for Swin-T on Tiny-ImageNet.} \textbf{Bold} and \second{underlined} denote best and second best, where ``best'' denotes smallest gap to \textsc{Retrain}.}
\label{tab:ablation_heldoutsize_swint}
\vspace*{0.2cm}
\renewcommand{\arraystretch}{0.88}
\setlength{\tabcolsep}{1pt}
\begin{tabular}{@{} l *{4}{>{\centering\arraybackslash}m{4.95em}} >{\columncolor{gapcol}[\tabcolsep][\tabcolsep]\centering\arraybackslash}m{4.95em} *{3}{>{\centering\arraybackslash}m{4.95em}} @{}}

\toprule
\addlinespace[0.2em]
& \multicolumn{8}{c}{\textbf{Forget 1\%}} \\
\addlinespace[0.2em]
\multicolumn{1}{c}{\small } & \multicolumn{1}{>{\centering\arraybackslash}m{4.95em}}{\small RA} & \multicolumn{1}{>{\centering\arraybackslash}m{4.95em}}{\small FA} & \multicolumn{1}{>{\centering\arraybackslash}m{4.95em}}{\small TA} & \multicolumn{1}{>{\centering\arraybackslash}m{4.95em}}{\small $\mathrm{RMIA}_{\mathrm{AUC}}$} & \multicolumn{1}{>{\columncolor{gapcol}[\tabcolsep][0pt]\centering\arraybackslash}m{4.95em}}{\small $\mathrm{Gap}_{\mathrm{Avg}}$} & \multicolumn{1}{>{\centering\arraybackslash}m{4.95em}}{\small $\mathrm{Gap}^{\mathrm{F\text{-}U}}_{\mathrm{Avg}}$} & \multicolumn{1}{>{\centering\arraybackslash}m{4.95em}}{\small $\mathrm{SMIA}_{\mathrm{AUC}}$} & \multicolumn{1}{>{\centering\arraybackslash}m{4.95em}}{\small $\mathrm{Gap}^{\mathrm{SMIA}}_{\mathrm{Avg}}$}\\
\addlinespace[0.1em]
\cmidrule(r){2-6}\cmidrule(l){7-9}

Held-out $0.1\%$ & \makebox[4.95em][c]{\makebox[2.85em][c]{$97.78$}\,\pmSmall{0.15}} & \makebox[4.95em][c]{\makebox[2.85em][c]{$54.41$}\,\pmSmall{2.79}} & \makebox[4.95em][c]{\makebox[2.85em][c]{$50.44$}\,\pmSmall{0.57}} & \makebox[4.95em][c]{\makebox[2.85em][c]{$\second{54.85}$}\,\pmSmall{1.31}} & \makebox[4.95em][c]{\makebox[2.85em][c]{$6.18$}\,\pmSmall{0.91}} & \makebox[4.95em][c]{\makebox[2.85em][c]{$\second{7.75}$}\,\pmSmall{0.58}} & \makebox[4.95em][c]{\makebox[2.85em][c]{$46.97$}\,\pmSmall{0.65}} & \makebox[4.95em][c]{\makebox[2.85em][c]{$5.82$}\,\pmSmall{0.81}} \\
\bottomrule
\end{tabular}
\end{center}
\end{table}

\begin{algorithm}[h!]
\caption{\textsc{RefDist}$_{\mathrm{uni}}$: Uniform Held-out Reference Distribution}
\label{alg:refdist_uniform}
\begin{algorithmic}[1]
\State \textbf{Input:} forget batch $B_{\mathrm{f}}$, held-out set $\mathcal{D}_{\mathrm{h}}$, reference model $f_{\phi}$
\State Sample $B_{\mathrm{h}}$: draw $|B_{\mathrm{f}}|$ examples uniformly from $\mathcal{D}_{\mathrm{h}}$ (with replacement if needed)
\State Aggregate reference predictions: $q \leftarrow \frac{1}{|B_{\mathrm{h}}|}\sum_{(x,\cdot)\in B_{\mathrm{h}}} p_{\phi}(\cdot\mid x)$
\State \textbf{Output:} $q \in \Delta^K$
\end{algorithmic}
\end{algorithm}
\begin{algorithm}[h!]
\caption{\textsc{RefDist}$_{\mathrm{1NN}}$: Feature-Matched Held-out Reference Distribution}
\label{alg:refdist_feature_1nn}
\begin{algorithmic}[1]
\State \textbf{Input:} forget batch $B_{\mathrm{f}}$, held-out set $\mathcal{D}_{\mathrm{h}}$, reference model $f_{\phi}$, feature map $g_{\phi}:\mathcal{X}\to\mathbb{R}^d$ induced by $f_{\phi}$ for nearest-neighbor retrieval
\State Compute normalized held-out features $\bar g_{\phi}(x)=g_{\phi}(x)/\|g_{\phi}(x)\|_2$ for all $(x,\cdot)\in\mathcal{D}_{\mathrm{h}}$
\State Initialize $B_{\mathrm{h}}\leftarrow\emptyset$
\For{each $(x_i,\cdot)\in B_{\mathrm{f}}$}
    \State Compute normalized forget feature $\bar g_{\phi}(x_i)=g_{\phi}(x_i)/\|g_{\phi}(x_i)\|_2$
    \State Find nearest held-out example: $j_i^\star \leftarrow \arg\max_{j:(x_j,\cdot)\in\mathcal{D}_{\mathrm{h}}} \bar g_{\phi}(x_i)^\top \bar g_{\phi}(x_j)$
    \State Add $(x_{j_i^\star},\cdot)$ to $B_{\mathrm{h}}$ \Comment{keeping multiplicities}
\EndFor
\State Aggregate reference predictions: $q \leftarrow \frac{1}{|B_{\mathrm{h}}|}\sum_{(x,\cdot)\in B_{\mathrm{h}}} p_{\phi}(\cdot\mid x)$
\State \textbf{Output:} $q \in \Delta^K$
\end{algorithmic}
\end{algorithm}
\clearpage

\subsection{Additional ResNet-18 Results on CIFAR-100 and Tiny-ImageNet}
\label[appsec]{appdx:add_experiments_extra}

This section provides additional results for ResNet-18 on CIFAR-100 and Tiny-ImageNet using the same experimental protocol described in \Cref{experimental_setup} and \Cref{appdx:exp_details}.
These experiments complement the main-text results by evaluating \method{} on additional ResNet-18 settings with larger label spaces and more challenging image data. 
\Cref{tab:resnet-cifar100,tab:resnet-imagenet} report the complete results across all forget fractions, and \Cref{fig:util_vs_forget_resc100_restiny} shows the corresponding forgetting--utility trade-offs under 10\% random forgetting.

The additional ResNet-18 results are in line with the main findings. 
Across CIFAR-100 and Tiny-ImageNet, \method{} is consistently among the strongest methods and achieves the best aggregate match to \textsc{Retrain} in most settings. 
The remaining cases follow the same pattern as in the main experiments: the best individual method can vary by dataset and forget fraction, but \method{} remains close to the strongest result while maintaining a stable balance between utility and membership indistinguishability. 
In particular, methods that preserve high test accuracy often retain a larger RMIA signal, whereas more aggressive forgetting methods can reduce this signal at the cost of utility. 
The trade-off curves in \Cref{fig:util_vs_forget_resc100_restiny} show the same behavior, with \method{} staying close to the \textsc{Retrain} RMIA reference while remaining in a competitive utility region. 
Overall, these additional benchmarks support that reference-guided distillation provides a robust unlearning signal beyond the primary settings. Note that we use $\mathrm{Gap}_{\mathrm{Avg}}$ with uniform weighting, which is a simple literature-aligned summary rather than a uniquely privileged weighting; as throughout the manuscript, all constituent metrics are also reported separately.

\input{inputs/plot_utility_forgetting_tradeoff_resnetcifar100_resnetimagenet}
\begin{table}[h!]
\begin{center}
\small
\caption{\textbf{Forgetting and utility results for ResNet-18 on CIFAR-100.}
Studied under 1\%, 10\%, and 50\% random forgetting. RA/FA/TA denote retain, forget, and test accuracy; $\mathrm{RMIA}_{\mathrm{AUC}}$ and $\mathrm{SMIA}_{\mathrm{AUC}}$ report membership inference performance, with 50\% corresponding to chance-level discrimination. Gap metrics measure deviation from \textsc{Retrain}, so lower is better. Results are mean~$\pm$~std over three seeds, with all values in percent. \textbf{Bold} and \second{underlined} denote best and second best, where ``best'' denotes smallest gap to \textsc{Retrain}.\looseness-1}
\label{tab:resnet-cifar100}
\vspace*{0.2cm}
\renewcommand{\arraystretch}{0.88}
\setlength{\tabcolsep}{1pt}
\begin{tabular}{@{} l *{4}{M} G *{3}{M} @{}}
\toprule
\addlinespace[0.3em]
& \multicolumn{8}{c}{\textbf{Forget 1\%}} \\
\addlinespace[0.3em]
\multicolumn{1}{c}{\small } & \multicolumn{1}{M}{\small RA} & \multicolumn{1}{M}{\small FA} & \multicolumn{1}{M}{\small TA} & \multicolumn{1}{M}{\small $\mathrm{RMIA}_{\mathrm{AUC}}$} & \multicolumn{1}{G}{\small $\mathrm{Gap}_{\mathrm{Avg}}$} & \multicolumn{1}{M}{\small $\mathrm{Gap}^{\mathrm{F\text{-}U}}_{\mathrm{Avg}}$} & \multicolumn{1}{M}{\small $\mathrm{SMIA}_{\mathrm{AUC}}$} & \multicolumn{1}{M}{\small $\mathrm{Gap}^{\mathrm{SMIA}}_{\mathrm{Avg}}$}\\
\addlinespace[0.2em]
\cmidrule(r){2-6}\cmidrule(l){7-9}

\textsc{Retrain} & \makebox[4.9em][c]{\makebox[2.75em][c]{$99.98$}\,\pmSmall{0.01}} & \makebox[4.9em][c]{\makebox[2.75em][c]{$75.18$}\,\pmSmall{1.22}} & \makebox[4.9em][c]{\makebox[2.75em][c]{$75.33$}\,\pmSmall{0.29}} & \makebox[4.9em][c]{\makebox[2.75em][c]{$49.88$}\,\pmSmall{1.56}} & \makebox[4.9em][c]{\makebox[2.75em][c]{$0.00$}\,\pmSmall{0.00}} & \makebox[4.9em][c]{\makebox[2.75em][c]{$0.00$}\,\pmSmall{0.00}} & \makebox[4.9em][c]{\makebox[2.75em][c]{$48.73$}\,\pmSmall{0.78}} & \makebox[4.9em][c]{\makebox[2.75em][c]{$0.00$}\,\pmSmall{0.00}} \\
\addlinespace[0.3em]
\textsc{Base} & \makebox[4.9em][c]{\makebox[2.75em][c]{$99.98$}\,\pmSmall{0.01}} & \makebox[4.9em][c]{\makebox[2.75em][c]{$100.00$}\,\pmSmall{0.00}} & \makebox[4.9em][c]{\makebox[2.75em][c]{$75.52$}\,\pmSmall{0.36}} & \makebox[4.9em][c]{\makebox[2.75em][c]{$76.09$}\,\pmSmall{1.32}} & \makebox[4.9em][c]{\makebox[2.75em][c]{$12.84$}\,\pmSmall{1.28}} & \makebox[4.9em][c]{\makebox[2.75em][c]{$13.47$}\,\pmSmall{0.68}} & \makebox[4.9em][c]{\makebox[2.75em][c]{$75.22$}\,\pmSmall{0.05}} & \makebox[4.9em][c]{\makebox[2.75em][c]{$12.97$}\,\pmSmall{0.38}} \\
\bottomrule
\end{tabular}
\end{center}
\end{table}

\begin{table*}[h!]
\begin{center}
\small
\caption{\textbf{Forgetting and utility results for ResNet-18 on Tiny-ImageNet.}
Studied under 1\%, 10\%, and 50\% random forgetting. RA/FA/TA denote retain, forget, and test accuracy; $\mathrm{RMIA}_{\mathrm{AUC}}$ and $\mathrm{SMIA}_{\mathrm{AUC}}$ report membership inference performance, with 50\% corresponding to chance-level discrimination. Gap metrics measure deviation from \textsc{Retrain}, so lower is better. Results are mean~$\pm$~std over three seeds, with all values in percent. \textbf{Bold} and \second{underlined} denote best and second best, where ``best'' denotes smallest gap to \textsc{Retrain}.\looseness-1}
\label{tab:resnet-imagenet}
\vspace*{0.2cm}
\renewcommand{\arraystretch}{0.88}
\setlength{\tabcolsep}{1pt}
\begin{tabular}{@{} l *{4}{M} G *{3}{M} @{}}

\toprule
\addlinespace[0.3em]
& \multicolumn{8}{c}{\textbf{Forget 1\%}} \\
\addlinespace[0.3em]
\multicolumn{1}{c}{\small } & \multicolumn{1}{M}{\small RA} & \multicolumn{1}{M}{\small FA} & \multicolumn{1}{M}{\small TA} & \multicolumn{1}{M}{\small $\mathrm{RMIA}_{\mathrm{AUC}}$} & \multicolumn{1}{G}{\small $\mathrm{Gap}_{\mathrm{Avg}}$} & \multicolumn{1}{M}{\small $\mathrm{Gap}^{\mathrm{F\text{-}U}}_{\mathrm{Avg}}$} & \multicolumn{1}{M}{\small $\mathrm{SMIA}_{\mathrm{AUC}}$} & \multicolumn{1}{M}{\small $\mathrm{Gap}^{\mathrm{SMIA}}_{\mathrm{Avg}}$}\\
\addlinespace[0.2em]
\cmidrule(r){2-6}\cmidrule(l){7-9}

\textsc{Retrain} & \makebox[4.9em][c]{\makebox[2.75em][c]{$99.98$}\,\pmSmall{0.00}} & \makebox[4.9em][c]{\makebox[2.75em][c]{$58.52$}\,\pmSmall{0.71}} & \makebox[4.9em][c]{\makebox[2.75em][c]{$59.54$}\,\pmSmall{0.21}} & \makebox[4.9em][c]{\makebox[2.75em][c]{$49.28$}\,\pmSmall{1.14}} & \makebox[4.9em][c]{\makebox[2.75em][c]{$0.00$}\,\pmSmall{0.00}} & \makebox[4.9em][c]{\makebox[2.75em][c]{$0.00$}\,\pmSmall{0.00}} & \makebox[4.9em][c]{\makebox[2.75em][c]{$49.84$}\,\pmSmall{0.85}} & \makebox[4.9em][c]{\makebox[2.75em][c]{$0.00$}\,\pmSmall{0.00}} \\
\addlinespace[0.3em]
\textsc{Base} & \makebox[4.9em][c]{\makebox[2.75em][c]{$99.98$}\,\pmSmall{0.00}} & \makebox[4.9em][c]{\makebox[2.75em][c]{$99.96$}\,\pmSmall{0.06}} & \makebox[4.9em][c]{\makebox[2.75em][c]{$59.64$}\,\pmSmall{0.44}} & \makebox[4.9em][c]{\makebox[2.75em][c]{$81.56$}\,\pmSmall{0.56}} & \makebox[4.9em][c]{\makebox[2.75em][c]{$18.50$}\,\pmSmall{0.61}} & \makebox[4.9em][c]{\makebox[2.75em][c]{$16.27$}\,\pmSmall{0.90}} & \makebox[4.9em][c]{\makebox[2.75em][c]{$84.68$}\,\pmSmall{0.16}} & \makebox[4.9em][c]{\makebox[2.75em][c]{$19.10$}\,\pmSmall{0.31}} \\
\bottomrule
\end{tabular}
\end{center}
\end{table*}

\clearpage

\subsection{Class-Forgetting Experiments}
\label[appsec]{appdx:class_forgetting}

To broaden the scope beyond random instance forgetting, we additionally evaluate full- and half-class forgetting for ResNet-18/CIFAR-10 and Swin-T/Tiny-ImageNet.

\paragraph{Experimental Setup.}
We consider two class-forgetting regimes: full-class forgetting, which removes all training examples from the target class or classes, and half-class forgetting, which removes 50\% of them. For ResNet-18 on CIFAR-10, we construct a separate single-class forgetting task for each of the ten classes. Since CIFAR-10 is class-balanced, the two regimes correspond to forgetting approximately 10\% and 5\% of the base-training data, respectively.
For Swin-T on Tiny-ImageNet, we define three semantically coherent groups of 20 classes, each treated as a superclass: arthropods and mollusks\footnote{WNIDs: \texttt{n01768244,n01770393,n01774384,n01774750,n01784675,n01944390,n01945685,n01950731,n01983481,n01984695,\\n02165456,n02190166,n02206856,n02226429,n02231487,n02233338,n02236044,n02268443,n02279972,n02281406}}, wearables and personal equipment\footnote{WNIDs: \texttt{n02669723,n02730930,n02769748,n02837789,n02883205,n02963159,n03404251,n03424325,n03617480,n03763968,\\n03770439,n03814639,n03980874,n04133789,n04251144,n04254777,n04259630,n04356056,n04371430,n04532106}}, and transportation and infrastructure\footnote{WNIDs: \texttt{n02814533,n02917067,n03100240,n03393912,n03444034,n03447447,n03599486,n03662601,n03670208,n03796401,\\n03891332,n03977966,n04146614,n04285008,n04311004,n04366367,n04465501,n04487081,n04501370,n04532670}}.
We construct both full- and half-class forgetting tasks for each group. This provides a tractable analogue of the CIFAR-10 setting, again corresponding to approximately 10\% and 5\% data removal.
Across both model--dataset settings, validation and held-out examples are reserved in a class-stratified manner before constructing the forget set. The resulting base-training set therefore remains balanced across classes, and forgetting is applied only to examples from the selected target class or class group within this pool.
The test set retains examples from all original classes. Consequently, particularly under full-class forgetting, the retrained model is expected to exhibit a decrease in overall test accuracy because it has not been trained on the removed target class or classes. In this setting, agreement with the test accuracy of \textsc{Retrain} is therefore more informative than maximizing test accuracy in absolute terms.
Regarding membership inference, for both RMIA and SMIA, the non-member evaluation set is restricted to unseen test examples from the same class as each forget example. This class-matched evaluation avoids confounding membership inference with differences in class composition between the structured forget set and the full test set. The four RMIA reference models are trained on the corresponding retain set $\mathcal D_r$.
Apart from this evaluation adjustment, we use the same implementations, optimization procedures, and hyperparameter-selection protocol as described in \cref{appdx:exp_details}. For \method, we replace the default class-conditioned held-out selector with the uniform held-out selector described in \cref{alg:refdist_uniform}.\looseness-1

\paragraph{Results.}
\Cref{tab:resnet-cifar10-class-forgetting,tab:resnet-cifar10-class-forgetting-forget0p5} report the full- and half-class forgetting results for ResNet-18/CIFAR-10, while \cref{tab:swin-imagenet-class-forgetting,tab:swin-imagenet-class-forgetting-forget0p5} present the corresponding results for Swin-T/Tiny-ImageNet.

Under full-class forgetting, \method is consistently among the methods closest to \textsc{Retrain} in aggregate across both model--dataset settings. While the strongest individual baselines differ between the two settings, \method maintains a favorable balance between utility and forgetting rather than optimizing either objective in isolation. On ResNet-18/CIFAR-10, its membership-inference behavior closely approaches the corresponding \textsc{Retrain} references; on Swin-T/Tiny-ImageNet, the remaining membership gaps are larger, but substantially reduced relative to the \textsc{Base} model while retaining comparatively strong utility.
Across both settings, \method retains a small nonzero forget accuracy, whereas \textsc{Retrain} and several baselines (e.g., \textsc{SalUn}) fully suppress the removed class or class group. This behavior may arise because the frozen-base teacher assigns some probability mass to the removed classes when constructing the reference distribution from uniformly sampled held-out examples. Consequently, the resulting target reflects the teacher's predictive behavior on non-member data rather than explicitly enforcing complete suppression of the forgotten classes.\looseness-1

Under half-class forgetting, \method achieves the strongest overall agreement with \textsc{Retrain} across both model--dataset settings and remains among the best methods on the utility--forgetting and SMIA-based aggregate gaps. Although it does not provide the closest match to \textsc{Retrain} on any individual metric, it achieves the most balanced trade-off between forgetting efficacy and model utility. Other baselines exhibit more one-sided behavior: some tend to drive forget accuracy below the retrained target at the cost of utility (e.g., \textsc{NegGrad+} and \textsc{LUR}), whereas others remain close to \textsc{Base} and provide limited evidence of removal (e.g., \textsc{Finetune} and $\ell_1$\textsc{-sparse}). This yields \method's clearest aggregate advantage under half-class forgetting.\looseness-1

\begin{table}[h!]
\begin{center}
\small
\caption{\textbf{Full-class forgetting results for ResNet-18 on CIFAR-10.} RA/FA/TA denote retain, forget, and test accuracy; $\mathrm{RMIA}_{\mathrm{AUC}}$ and $\mathrm{SMIA}_{\mathrm{AUC}}$ report membership inference performance, with 50\% corresponding to chance-level discrimination. Gap metrics measure deviation from \textsc{Retrain}, so lower is better. Results are averaged over all 10 classes within each seed and reported as mean~$\pm$~sample std across three seeds, with all values in percent. \textbf{Bold} and \second{underlined} denote best and second best, where ``best'' denotes smallest gap to \textsc{Retrain}.}
\label{tab:resnet-cifar10-class-forgetting}
\vspace*{0.2cm}
\renewcommand{\arraystretch}{0.88}
\setlength{\tabcolsep}{1pt}
\begin{tabular}{@{} l *{4}{M} G *{3}{M} @{}}
\toprule
\addlinespace[0.4em]
& \multicolumn{8}{c}{\textbf{Full-class forgetting: one target class}} \\
\addlinespace[0.4em]
\multicolumn{1}{c}{\small } & \multicolumn{1}{M}{\small RA} & \multicolumn{1}{M}{\small FA} & \multicolumn{1}{M}{\small TA} & \multicolumn{1}{M}{\small $\mathrm{RMIA}_{\mathrm{AUC}}$} & \multicolumn{1}{G}{\small $\mathrm{Gap}_{\mathrm{Avg}}$} & \multicolumn{1}{M}{\small $\mathrm{Gap}^{\mathrm{F\text{-}U}}_{\mathrm{Avg}}$} & \multicolumn{1}{M}{\small $\mathrm{SMIA}_{\mathrm{AUC}}$} & \multicolumn{1}{M}{\small $\mathrm{Gap}^{\mathrm{SMIA}}_{\mathrm{Avg}}$}\\
\addlinespace[0.2em]
\cmidrule(r){2-6}\cmidrule(l){7-9}

\textsc{Retrain} & \makebox[4.9em][c]{\makebox[2.75em][c]{$100.00$}\,\pmSmall{0.00}} & \makebox[4.9em][c]{\makebox[2.75em][c]{$0.00$}\,\pmSmall{0.00}} & \makebox[4.9em][c]{\makebox[2.75em][c]{$85.16$}\,\pmSmall{0.03}} & \makebox[4.9em][c]{\makebox[2.75em][c]{$50.50$}\,\pmSmall{0.15}} & \makebox[4.9em][c]{\makebox[2.75em][c]{$0.00$}\,\pmSmall{0.00}} & \makebox[4.9em][c]{\makebox[2.75em][c]{$0.00$}\,\pmSmall{0.00}} & \makebox[4.9em][c]{\makebox[2.75em][c]{$50.45$}\,\pmSmall{0.40}} & \makebox[4.9em][c]{\makebox[2.75em][c]{$0.00$}\,\pmSmall{0.00}} \\
\addlinespace[0.4em]
\textsc{Base} & \makebox[4.9em][c]{\makebox[2.75em][c]{$100.00$}\,\pmSmall{0.00}} & \makebox[4.9em][c]{\makebox[2.75em][c]{$100.00$}\,\pmSmall{0.00}} & \makebox[4.9em][c]{\makebox[2.75em][c]{$94.35$}\,\pmSmall{0.31}} & \makebox[4.9em][c]{\makebox[2.75em][c]{$52.39$}\,\pmSmall{0.39}} & \makebox[4.9em][c]{\makebox[2.75em][c]{$27.83$}\,\pmSmall{0.13}} & \makebox[4.9em][c]{\makebox[2.75em][c]{$5.65$}\,\pmSmall{0.25}} & \makebox[4.9em][c]{\makebox[2.75em][c]{$59.24$}\,\pmSmall{0.19}} & \makebox[4.9em][c]{\makebox[2.75em][c]{$29.50$}\,\pmSmall{0.15}} \\
\cmidrule(r){1-6}\cmidrule(l){7-9}
\textsc{NegGrad} & \makebox[4.9em][c]{\makebox[2.75em][c]{$55.30$}\,\pmSmall{2.76}} & \makebox[4.9em][c]{\makebox[2.75em][c]{$\best{0.00}$}\,\pmSmall{0.00}} & \makebox[4.9em][c]{\makebox[2.75em][c]{$47.52$}\,\pmSmall{2.30}} & \makebox[4.9em][c]{\makebox[2.75em][c]{$49.20$}\,\pmSmall{0.28}} & \makebox[4.9em][c]{\makebox[2.75em][c]{$20.96$}\,\pmSmall{1.22}} & \makebox[4.9em][c]{\makebox[2.75em][c]{$19.58$}\,\pmSmall{1.07}} & \makebox[4.9em][c]{\makebox[2.75em][c]{$48.86$}\,\pmSmall{0.09}} & \makebox[4.9em][c]{\makebox[2.75em][c]{$21.07$}\,\pmSmall{1.21}} \\
\addlinespace[0.4em]
\textsc{NegGrad+} & \makebox[4.9em][c]{\makebox[2.75em][c]{$\second{99.72}$}\,\pmSmall{0.03}} & \makebox[4.9em][c]{\makebox[2.75em][c]{$19.65$}\,\pmSmall{0.82}} & \makebox[4.9em][c]{\makebox[2.75em][c]{$85.83$}\,\pmSmall{0.26}} & \makebox[4.9em][c]{\makebox[2.75em][c]{$55.93$}\,\pmSmall{0.22}} & \makebox[4.9em][c]{\makebox[2.75em][c]{$6.53$}\,\pmSmall{0.33}} & \makebox[4.9em][c]{\makebox[2.75em][c]{$3.10$}\,\pmSmall{0.27}} & \makebox[4.9em][c]{\makebox[2.75em][c]{$55.82$}\,\pmSmall{0.13}} & \makebox[4.9em][c]{\makebox[2.75em][c]{$6.51$}\,\pmSmall{0.38}} \\
\addlinespace[0.4em]
\textsc{Finetune} & \makebox[4.9em][c]{\makebox[2.75em][c]{$86.37$}\,\pmSmall{0.55}} & \makebox[4.9em][c]{\makebox[2.75em][c]{$\second{0.00}$}\,\pmSmall{0.00}} & \makebox[4.9em][c]{\makebox[2.75em][c]{$74.59$}\,\pmSmall{0.38}} & \makebox[4.9em][c]{\makebox[2.75em][c]{$51.25$}\,\pmSmall{0.28}} & \makebox[4.9em][c]{\makebox[2.75em][c]{$6.33$}\,\pmSmall{0.16}} & \makebox[4.9em][c]{\makebox[2.75em][c]{$5.85$}\,\pmSmall{0.08}} & \makebox[4.9em][c]{\makebox[2.75em][c]{$51.48$}\,\pmSmall{0.04}} & \makebox[4.9em][c]{\makebox[2.75em][c]{$6.40$}\,\pmSmall{0.18}} \\
\addlinespace[0.4em]
$\ell_1$-\textsc{sparse} & \makebox[4.9em][c]{\makebox[2.75em][c]{$85.85$}\,\pmSmall{0.56}} & \makebox[4.9em][c]{\makebox[2.75em][c]{$0.00$}\,\pmSmall{0.00}} & \makebox[4.9em][c]{\makebox[2.75em][c]{$74.27$}\,\pmSmall{0.50}} & \makebox[4.9em][c]{\makebox[2.75em][c]{$\second{51.17}$}\,\pmSmall{0.20}} & \makebox[4.9em][c]{\makebox[2.75em][c]{$6.52$}\,\pmSmall{0.20}} & \makebox[4.9em][c]{\makebox[2.75em][c]{$5.98$}\,\pmSmall{0.12}} & \makebox[4.9em][c]{\makebox[2.75em][c]{$\second{51.44}$}\,\pmSmall{0.04}} & \makebox[4.9em][c]{\makebox[2.75em][c]{$6.59$}\,\pmSmall{0.22}} \\
\addlinespace[0.4em]
\textsc{LUR} & \makebox[4.9em][c]{\makebox[2.75em][c]{$98.52$}\,\pmSmall{0.32}} & \makebox[4.9em][c]{\makebox[2.75em][c]{$0.45$}\,\pmSmall{0.10}} & \makebox[4.9em][c]{\makebox[2.75em][c]{$82.42$}\,\pmSmall{0.45}} & \makebox[4.9em][c]{\makebox[2.75em][c]{$53.77$}\,\pmSmall{0.40}} & \makebox[4.9em][c]{\makebox[2.75em][c]{$2.03$}\,\pmSmall{0.16}} & \makebox[4.9em][c]{\makebox[2.75em][c]{$3.11$}\,\pmSmall{0.30}} & \makebox[4.9em][c]{\makebox[2.75em][c]{$53.76$}\,\pmSmall{0.34}} & \makebox[4.9em][c]{\makebox[2.75em][c]{$2.06$}\,\pmSmall{0.18}} \\
\addlinespace[0.4em]
\textsc{SSD} & \makebox[4.9em][c]{\makebox[2.75em][c]{$83.73$}\,\pmSmall{3.14}} & \makebox[4.9em][c]{\makebox[2.75em][c]{$15.59$}\,\pmSmall{7.72}} & \makebox[4.9em][c]{\makebox[2.75em][c]{$72.10$}\,\pmSmall{2.33}} & \makebox[4.9em][c]{\makebox[2.75em][c]{$52.87$}\,\pmSmall{0.48}} & \makebox[4.9em][c]{\makebox[2.75em][c]{$12.68$}\,\pmSmall{2.60}} & \makebox[4.9em][c]{\makebox[2.75em][c]{$9.44$}\,\pmSmall{0.91}} & \makebox[4.9em][c]{\makebox[2.75em][c]{$53.53$}\,\pmSmall{0.77}} & \makebox[4.9em][c]{\makebox[2.75em][c]{$12.83$}\,\pmSmall{2.87}} \\
\addlinespace[0.4em]
\textsc{SalUn} & \makebox[4.9em][c]{\makebox[2.75em][c]{$99.70$}\,\pmSmall{0.01}} & \makebox[4.9em][c]{\makebox[2.75em][c]{$0.00$}\,\pmSmall{0.00}} & \makebox[4.9em][c]{\makebox[2.75em][c]{$84.00$}\,\pmSmall{0.24}} & \makebox[4.9em][c]{\makebox[2.75em][c]{$51.70$}\,\pmSmall{0.30}} & \makebox[4.9em][c]{\makebox[2.75em][c]{$\best{0.79}$}\,\pmSmall{0.14}} & \makebox[4.9em][c]{\makebox[2.75em][c]{$\second{1.43}$}\,\pmSmall{0.28}} & \makebox[4.9em][c]{\makebox[2.75em][c]{$55.00$}\,\pmSmall{0.57}} & \makebox[4.9em][c]{\makebox[2.75em][c]{$\best{1.51}$}\,\pmSmall{0.05}} \\
\addlinespace[0.4em]
\textsc{Amun} & \makebox[4.9em][c]{\makebox[2.75em][c]{$99.54$}\,\pmSmall{0.03}} & \makebox[4.9em][c]{\makebox[2.75em][c]{$11.95$}\,\pmSmall{0.56}} & \makebox[4.9em][c]{\makebox[2.75em][c]{$\best{84.95}$}\,\pmSmall{0.08}} & \makebox[4.9em][c]{\makebox[2.75em][c]{$52.68$}\,\pmSmall{0.31}} & \makebox[4.9em][c]{\makebox[2.75em][c]{$3.85$}\,\pmSmall{0.21}} & \makebox[4.9em][c]{\makebox[2.75em][c]{$1.50$}\,\pmSmall{0.28}} & \makebox[4.9em][c]{\makebox[2.75em][c]{$52.27$}\,\pmSmall{0.20}} & \makebox[4.9em][c]{\makebox[2.75em][c]{$3.81$}\,\pmSmall{0.19}} \\
\addlinespace[0.4em]
\textsc{BadTeach} & \makebox[4.9em][c]{\makebox[2.75em][c]{$\best{99.94}$}\,\pmSmall{0.01}} & \makebox[4.9em][c]{\makebox[2.75em][c]{$9.66$}\,\pmSmall{0.61}} & \makebox[4.9em][c]{\makebox[2.75em][c]{$\second{85.38}$}\,\pmSmall{0.22}} & \makebox[4.9em][c]{\makebox[2.75em][c]{$51.93$}\,\pmSmall{0.26}} & \makebox[4.9em][c]{\makebox[2.75em][c]{$3.18$}\,\pmSmall{0.13}} & \makebox[4.9em][c]{\makebox[2.75em][c]{$1.49$}\,\pmSmall{0.13}} & \makebox[4.9em][c]{\makebox[2.75em][c]{$55.27$}\,\pmSmall{0.36}} & \makebox[4.9em][c]{\makebox[2.75em][c]{$3.93$}\,\pmSmall{0.14}} \\
\addlinespace[0.4em]
\textbf{\textsc{\method}} & \makebox[4.9em][c]{\makebox[2.75em][c]{$99.27$}\,\pmSmall{0.25}} & \makebox[4.9em][c]{\makebox[2.75em][c]{$3.42$}\,\pmSmall{0.60}} & \makebox[4.9em][c]{\makebox[2.75em][c]{$83.63$}\,\pmSmall{0.23}} & \makebox[4.9em][c]{\makebox[2.75em][c]{$\best{50.00}$}\,\pmSmall{0.28}} & \makebox[4.9em][c]{\makebox[2.75em][c]{$\second{1.73}$}\,\pmSmall{0.25}} & \makebox[4.9em][c]{\makebox[2.75em][c]{$\best{1.39}$}\,\pmSmall{0.17}} & \makebox[4.9em][c]{\makebox[2.75em][c]{$\best{51.30}$}\,\pmSmall{0.06}} & \makebox[4.9em][c]{\makebox[2.75em][c]{$\second{1.90}$}\,\pmSmall{0.24}} \\
\addlinespace[0.4em]
\bottomrule
\end{tabular}
\end{center}
\end{table}

\begin{table}[h!]
\begin{center}
\small
\caption{\textbf{Half-class forgetting results for ResNet-18 on CIFAR-10.} RA/FA/TA denote retain, forget, and test accuracy; $\mathrm{RMIA}_{\mathrm{AUC}}$ and $\mathrm{SMIA}_{\mathrm{AUC}}$ report membership inference performance, with 50\% corresponding to chance-level discrimination. Gap metrics measure deviation from \textsc{Retrain}, so lower is better. Results are averaged over all 10 classes within each seed and reported as mean~$\pm$~sample std across three seeds, with all values in percent. \textbf{Bold} and \second{underlined} denote best and second best, where ``best'' denotes smallest gap to \textsc{Retrain}.}
\label{tab:resnet-cifar10-class-forgetting-forget0p5}
\vspace*{0.2cm}
\renewcommand{\arraystretch}{0.88}
\setlength{\tabcolsep}{1pt}
\begin{tabular}{@{} l *{4}{M} G *{3}{M} @{}}
\toprule
\addlinespace[0.4em]
& \multicolumn{8}{c}{\textbf{Half-class forgetting: one target class}} \\
\addlinespace[0.4em]
\multicolumn{1}{c}{\small } & \multicolumn{1}{M}{\small RA} & \multicolumn{1}{M}{\small FA} & \multicolumn{1}{M}{\small TA} & \multicolumn{1}{M}{\small $\mathrm{RMIA}_{\mathrm{AUC}}$} & \multicolumn{1}{G}{\small $\mathrm{Gap}_{\mathrm{Avg}}$} & \multicolumn{1}{M}{\small $\mathrm{Gap}^{\mathrm{F\text{-}U}}_{\mathrm{Avg}}$} & \multicolumn{1}{M}{\small $\mathrm{SMIA}_{\mathrm{AUC}}$} & \multicolumn{1}{M}{\small $\mathrm{Gap}^{\mathrm{SMIA}}_{\mathrm{Avg}}$}\\
\addlinespace[0.2em]
\cmidrule(r){2-6}\cmidrule(l){7-9}

\textsc{Retrain} & \makebox[4.9em][c]{\makebox[2.75em][c]{$100.00$}\,\pmSmall{0.00}} & \makebox[4.9em][c]{\makebox[2.75em][c]{$90.62$}\,\pmSmall{0.15}} & \makebox[4.9em][c]{\makebox[2.75em][c]{$93.93$}\,\pmSmall{0.05}} & \makebox[4.9em][c]{\makebox[2.75em][c]{$50.29$}\,\pmSmall{0.17}} & \makebox[4.9em][c]{\makebox[2.75em][c]{$0.00$}\,\pmSmall{0.00}} & \makebox[4.9em][c]{\makebox[2.75em][c]{$0.00$}\,\pmSmall{0.00}} & \makebox[4.9em][c]{\makebox[2.75em][c]{$50.27$}\,\pmSmall{0.12}} & \makebox[4.9em][c]{\makebox[2.75em][c]{$0.00$}\,\pmSmall{0.00}} \\
\addlinespace[0.4em]
\textsc{Base} & \makebox[4.9em][c]{\makebox[2.75em][c]{$100.00$}\,\pmSmall{0.00}} & \makebox[4.9em][c]{\makebox[2.75em][c]{$100.00$}\,\pmSmall{0.00}} & \makebox[4.9em][c]{\makebox[2.75em][c]{$94.35$}\,\pmSmall{0.31}} & \makebox[4.9em][c]{\makebox[2.75em][c]{$55.79$}\,\pmSmall{0.43}} & \makebox[4.9em][c]{\makebox[2.75em][c]{$3.84$}\,\pmSmall{0.04}} & \makebox[4.9em][c]{\makebox[2.75em][c]{$2.99$}\,\pmSmall{0.16}} & \makebox[4.9em][c]{\makebox[2.75em][c]{$59.24$}\,\pmSmall{0.15}} & \makebox[4.9em][c]{\makebox[2.75em][c]{$4.71$}\,\pmSmall{0.12}} \\
\cmidrule(r){1-6}\cmidrule(l){7-9}
\textsc{NegGrad} & \makebox[4.9em][c]{\makebox[2.75em][c]{$76.10$}\,\pmSmall{1.63}} & \makebox[4.9em][c]{\makebox[2.75em][c]{$0.22$}\,\pmSmall{0.07}} & \makebox[4.9em][c]{\makebox[2.75em][c]{$67.60$}\,\pmSmall{1.50}} & \makebox[4.9em][c]{\makebox[2.75em][c]{$\second{48.55}$}\,\pmSmall{0.14}} & \makebox[4.9em][c]{\makebox[2.75em][c]{$35.68$}\,\pmSmall{0.86}} & \makebox[4.9em][c]{\makebox[2.75em][c]{$14.20$}\,\pmSmall{0.85}} & \makebox[4.9em][c]{\makebox[2.75em][c]{$\second{48.71}$}\,\pmSmall{0.12}} & \makebox[4.9em][c]{\makebox[2.75em][c]{$35.60$}\,\pmSmall{0.87}} \\
\addlinespace[0.4em]
\textsc{NegGrad+} & \makebox[4.9em][c]{\makebox[2.75em][c]{$98.44$}\,\pmSmall{0.03}} & \makebox[4.9em][c]{\makebox[2.75em][c]{$71.68$}\,\pmSmall{0.67}} & \makebox[4.9em][c]{\makebox[2.75em][c]{$90.27$}\,\pmSmall{0.20}} & \makebox[4.9em][c]{\makebox[2.75em][c]{$59.87$}\,\pmSmall{0.18}} & \makebox[4.9em][c]{\makebox[2.75em][c]{$8.44$}\,\pmSmall{0.29}} & \makebox[4.9em][c]{\makebox[2.75em][c]{$6.62$}\,\pmSmall{0.18}} & \makebox[4.9em][c]{\makebox[2.75em][c]{$56.24$}\,\pmSmall{0.14}} & \makebox[4.9em][c]{\makebox[2.75em][c]{$7.53$}\,\pmSmall{0.26}} \\
\addlinespace[0.4em]
\textsc{Finetune} & \makebox[4.9em][c]{\makebox[2.75em][c]{$\second{99.99}$}\,\pmSmall{0.00}} & \makebox[4.9em][c]{\makebox[2.75em][c]{$99.99$}\,\pmSmall{0.00}} & \makebox[4.9em][c]{\makebox[2.75em][c]{$\second{94.18}$}\,\pmSmall{0.26}} & \makebox[4.9em][c]{\makebox[2.75em][c]{$56.26$}\,\pmSmall{0.17}} & \makebox[4.9em][c]{\makebox[2.75em][c]{$3.94$}\,\pmSmall{0.02}} & \makebox[4.9em][c]{\makebox[2.75em][c]{$3.19$}\,\pmSmall{0.12}} & \makebox[4.9em][c]{\makebox[2.75em][c]{$58.76$}\,\pmSmall{0.18}} & \makebox[4.9em][c]{\makebox[2.75em][c]{$4.57$}\,\pmSmall{0.09}} \\
\addlinespace[0.4em]
$\ell_1$-\textsc{sparse} & \makebox[4.9em][c]{\makebox[2.75em][c]{$99.99$}\,\pmSmall{0.00}} & \makebox[4.9em][c]{\makebox[2.75em][c]{$99.99$}\,\pmSmall{0.00}} & \makebox[4.9em][c]{\makebox[2.75em][c]{$\best{94.17}$}\,\pmSmall{0.27}} & \makebox[4.9em][c]{\makebox[2.75em][c]{$56.26$}\,\pmSmall{0.19}} & \makebox[4.9em][c]{\makebox[2.75em][c]{$3.94$}\,\pmSmall{0.03}} & \makebox[4.9em][c]{\makebox[2.75em][c]{$3.19$}\,\pmSmall{0.13}} & \makebox[4.9em][c]{\makebox[2.75em][c]{$58.76$}\,\pmSmall{0.18}} & \makebox[4.9em][c]{\makebox[2.75em][c]{$4.57$}\,\pmSmall{0.09}} \\
\addlinespace[0.4em]
\textsc{LUR} & \makebox[4.9em][c]{\makebox[2.75em][c]{$99.40$}\,\pmSmall{0.02}} & \makebox[4.9em][c]{\makebox[2.75em][c]{$\second{89.29}$}\,\pmSmall{0.20}} & \makebox[4.9em][c]{\makebox[2.75em][c]{$91.92$}\,\pmSmall{0.22}} & \makebox[4.9em][c]{\makebox[2.75em][c]{$63.32$}\,\pmSmall{0.20}} & \makebox[4.9em][c]{\makebox[2.75em][c]{$4.96$}\,\pmSmall{0.17}} & \makebox[4.9em][c]{\makebox[2.75em][c]{$7.51$}\,\pmSmall{0.28}} & \makebox[4.9em][c]{\makebox[2.75em][c]{$57.08$}\,\pmSmall{0.13}} & \makebox[4.9em][c]{\makebox[2.75em][c]{$3.41$}\,\pmSmall{0.12}} \\
\addlinespace[0.4em]
\textsc{SSD} & \makebox[4.9em][c]{\makebox[2.75em][c]{$\best{100.00}$}\,\pmSmall{0.00}} & \makebox[4.9em][c]{\makebox[2.75em][c]{$100.00$}\,\pmSmall{0.00}} & \makebox[4.9em][c]{\makebox[2.75em][c]{$94.35$}\,\pmSmall{0.31}} & \makebox[4.9em][c]{\makebox[2.75em][c]{$55.79$}\,\pmSmall{0.43}} & \makebox[4.9em][c]{\makebox[2.75em][c]{$3.84$}\,\pmSmall{0.04}} & \makebox[4.9em][c]{\makebox[2.75em][c]{$2.99$}\,\pmSmall{0.16}} & \makebox[4.9em][c]{\makebox[2.75em][c]{$59.24$}\,\pmSmall{0.15}} & \makebox[4.9em][c]{\makebox[2.75em][c]{$4.71$}\,\pmSmall{0.12}} \\
\addlinespace[0.4em]
\textsc{SalUn} & \makebox[4.9em][c]{\makebox[2.75em][c]{$99.09$}\,\pmSmall{0.10}} & \makebox[4.9em][c]{\makebox[2.75em][c]{$86.08$}\,\pmSmall{1.43}} & \makebox[4.9em][c]{\makebox[2.75em][c]{$91.59$}\,\pmSmall{0.35}} & \makebox[4.9em][c]{\makebox[2.75em][c]{$60.35$}\,\pmSmall{0.11}} & \makebox[4.9em][c]{\makebox[2.75em][c]{$4.85$}\,\pmSmall{0.43}} & \makebox[4.9em][c]{\makebox[2.75em][c]{$6.19$}\,\pmSmall{0.32}} & \makebox[4.9em][c]{\makebox[2.75em][c]{$57.52$}\,\pmSmall{0.16}} & \makebox[4.9em][c]{\makebox[2.75em][c]{$4.15$}\,\pmSmall{0.41}} \\
\addlinespace[0.4em]
\textsc{Amun} & \makebox[4.9em][c]{\makebox[2.75em][c]{$99.37$}\,\pmSmall{0.04}} & \makebox[4.9em][c]{\makebox[2.75em][c]{$\best{89.51}$}\,\pmSmall{2.04}} & \makebox[4.9em][c]{\makebox[2.75em][c]{$92.10$}\,\pmSmall{0.22}} & \makebox[4.9em][c]{\makebox[2.75em][c]{$54.56$}\,\pmSmall{0.08}} & \makebox[4.9em][c]{\makebox[2.75em][c]{$\second{2.58}$}\,\pmSmall{0.27}} & \makebox[4.9em][c]{\makebox[2.75em][c]{$3.07$}\,\pmSmall{0.20}} & \makebox[4.9em][c]{\makebox[2.75em][c]{$\best{51.40}$}\,\pmSmall{0.18}} & \makebox[4.9em][c]{\makebox[2.75em][c]{$\best{1.92}$}\,\pmSmall{0.25}} \\
\addlinespace[0.4em]
\textsc{BadTeach} & \makebox[4.9em][c]{\makebox[2.75em][c]{$99.83$}\,\pmSmall{0.22}} & \makebox[4.9em][c]{\makebox[2.75em][c]{$95.16$}\,\pmSmall{6.40}} & \makebox[4.9em][c]{\makebox[2.75em][c]{$93.47$}\,\pmSmall{0.60}} & \makebox[4.9em][c]{\makebox[2.75em][c]{$\best{51.12}$}\,\pmSmall{0.27}} & \makebox[4.9em][c]{\makebox[2.75em][c]{$3.12$}\,\pmSmall{0.47}} & \makebox[4.9em][c]{\makebox[2.75em][c]{$\best{1.70}$}\,\pmSmall{0.22}} & \makebox[4.9em][c]{\makebox[2.75em][c]{$52.04$}\,\pmSmall{0.40}} & \makebox[4.9em][c]{\makebox[2.75em][c]{$3.15$}\,\pmSmall{0.46}} \\
\addlinespace[0.4em]
\textbf{\textsc{\method}} & \makebox[4.9em][c]{\makebox[2.75em][c]{$99.39$}\,\pmSmall{0.07}} & \makebox[4.9em][c]{\makebox[2.75em][c]{$94.34$}\,\pmSmall{0.74}} & \makebox[4.9em][c]{\makebox[2.75em][c]{$92.11$}\,\pmSmall{0.35}} & \makebox[4.9em][c]{\makebox[2.75em][c]{$53.58$}\,\pmSmall{0.19}} & \makebox[4.9em][c]{\makebox[2.75em][c]{$\best{2.47}$}\,\pmSmall{0.14}} & \makebox[4.9em][c]{\makebox[2.75em][c]{$\second{2.63}$}\,\pmSmall{0.45}} & \makebox[4.9em][c]{\makebox[2.75em][c]{$55.19$}\,\pmSmall{0.16}} & \makebox[4.9em][c]{\makebox[2.75em][c]{$\second{2.84}$}\,\pmSmall{0.11}} \\
\addlinespace[0.4em]
\bottomrule
\end{tabular}
\end{center}
\end{table}

\begin{table}[h!]
\begin{center}
\small
\caption{\textbf{Full-class forgetting results for Swin-T on Tiny-ImageNet.} RA/FA/TA denote retain, forget, and test accuracy; $\mathrm{RMIA}_{\mathrm{AUC}}$ and $\mathrm{SMIA}_{\mathrm{AUC}}$ report membership inference performance, with 50\% corresponding to chance-level discrimination. Gap metrics measure deviation from \textsc{Retrain}, so lower is better. Results are averaged over the three 20-class forgetting groups within each seed and reported as mean~$\pm$~sample std across three seeds, with all values in percent. \textbf{Bold} and \second{underlined} denote best and second best, where ``best'' denotes smallest gap to \textsc{Retrain}.}
\label{tab:swin-imagenet-class-forgetting}
\vspace*{0.2cm}
\renewcommand{\arraystretch}{0.88}
\setlength{\tabcolsep}{1pt}
\begin{tabular}{@{} l *{4}{M} G *{3}{M} @{}}
\toprule
\addlinespace[0.4em]
& \multicolumn{8}{c}{\textbf{Full-class forgetting: one 20-class group}} \\
\addlinespace[0.4em]
\multicolumn{1}{c}{\small } & \multicolumn{1}{M}{\small RA} & \multicolumn{1}{M}{\small FA} & \multicolumn{1}{M}{\small TA} & \multicolumn{1}{M}{\small $\mathrm{RMIA}_{\mathrm{AUC}}$} & \multicolumn{1}{G}{\small $\mathrm{Gap}_{\mathrm{Avg}}$} & \multicolumn{1}{M}{\small $\mathrm{Gap}^{\mathrm{F\text{-}U}}_{\mathrm{Avg}}$} & \multicolumn{1}{M}{\small $\mathrm{SMIA}_{\mathrm{AUC}}$} & \multicolumn{1}{M}{\small $\mathrm{Gap}^{\mathrm{SMIA}}_{\mathrm{Avg}}$}\\
\addlinespace[0.2em]
\cmidrule(r){2-6}\cmidrule(l){7-9}

\textsc{Retrain} & \makebox[4.9em][c]{\makebox[2.75em][c]{$99.98$}\,\pmSmall{0.00}} & \makebox[4.9em][c]{\makebox[2.75em][c]{$0.00$}\,\pmSmall{0.00}} & \makebox[4.9em][c]{\makebox[2.75em][c]{$54.82$}\,\pmSmall{0.09}} & \makebox[4.9em][c]{\makebox[2.75em][c]{$49.83$}\,\pmSmall{0.48}} & \makebox[4.9em][c]{\makebox[2.75em][c]{$0.00$}\,\pmSmall{0.00}} & \makebox[4.9em][c]{\makebox[2.75em][c]{$0.00$}\,\pmSmall{0.00}} & \makebox[4.9em][c]{\makebox[2.75em][c]{$49.71$}\,\pmSmall{0.14}} & \makebox[4.9em][c]{\makebox[2.75em][c]{$0.00$}\,\pmSmall{0.00}} \\
\addlinespace[0.4em]
\textsc{Base} & \makebox[4.9em][c]{\makebox[2.75em][c]{$99.98$}\,\pmSmall{0.00}} & \makebox[4.9em][c]{\makebox[2.75em][c]{$99.99$}\,\pmSmall{0.00}} & \makebox[4.9em][c]{\makebox[2.75em][c]{$61.01$}\,\pmSmall{0.10}} & \makebox[4.9em][c]{\makebox[2.75em][c]{$86.94$}\,\pmSmall{0.38}} & \makebox[4.9em][c]{\makebox[2.75em][c]{$35.82$}\,\pmSmall{0.07}} & \makebox[4.9em][c]{\makebox[2.75em][c]{$21.65$}\,\pmSmall{0.14}} & \makebox[4.9em][c]{\makebox[2.75em][c]{$94.05$}\,\pmSmall{0.12}} & \makebox[4.9em][c]{\makebox[2.75em][c]{$37.63$}\,\pmSmall{0.10}} \\
\cmidrule(r){1-6}\cmidrule(l){7-9}
\textsc{NegGrad} & \makebox[4.9em][c]{\makebox[2.75em][c]{$\best{99.98}$}\,\pmSmall{0.00}} & \makebox[4.9em][c]{\makebox[2.75em][c]{$99.99$}\,\pmSmall{0.00}} & \makebox[4.9em][c]{\makebox[2.75em][c]{$61.00$}\,\pmSmall{0.10}} & \makebox[4.9em][c]{\makebox[2.75em][c]{$86.97$}\,\pmSmall{0.38}} & \makebox[4.9em][c]{\makebox[2.75em][c]{$35.83$}\,\pmSmall{0.07}} & \makebox[4.9em][c]{\makebox[2.75em][c]{$21.67$}\,\pmSmall{0.14}} & \makebox[4.9em][c]{\makebox[2.75em][c]{$94.04$}\,\pmSmall{0.12}} & \makebox[4.9em][c]{\makebox[2.75em][c]{$37.63$}\,\pmSmall{0.10}} \\
\addlinespace[0.4em]
\textsc{NegGrad+} & \makebox[4.9em][c]{\makebox[2.75em][c]{$38.59$}\,\pmSmall{1.91}} & \makebox[4.9em][c]{\makebox[2.75em][c]{$\best{0.00}$}\,\pmSmall{0.00}} & \makebox[4.9em][c]{\makebox[2.75em][c]{$27.23$}\,\pmSmall{1.20}} & \makebox[4.9em][c]{\makebox[2.75em][c]{$\best{50.09}$}\,\pmSmall{0.37}} & \makebox[4.9em][c]{\makebox[2.75em][c]{$22.52$}\,\pmSmall{0.93}} & \makebox[4.9em][c]{\makebox[2.75em][c]{$14.35$}\,\pmSmall{0.90}} & \makebox[4.9em][c]{\makebox[2.75em][c]{$\best{49.16}$}\,\pmSmall{0.11}} & \makebox[4.9em][c]{\makebox[2.75em][c]{$22.40$}\,\pmSmall{0.83}} \\
\addlinespace[0.4em]
\textsc{Finetune} & \makebox[4.9em][c]{\makebox[2.75em][c]{$52.81$}\,\pmSmall{0.29}} & \makebox[4.9em][c]{\makebox[2.75em][c]{$0.00$}\,\pmSmall{0.01}} & \makebox[4.9em][c]{\makebox[2.75em][c]{$33.35$}\,\pmSmall{0.21}} & \makebox[4.9em][c]{\makebox[2.75em][c]{$62.89$}\,\pmSmall{0.09}} & \makebox[4.9em][c]{\makebox[2.75em][c]{$20.43$}\,\pmSmall{0.11}} & \makebox[4.9em][c]{\makebox[2.75em][c]{$17.26$}\,\pmSmall{0.27}} & \makebox[4.9em][c]{\makebox[2.75em][c]{$62.74$}\,\pmSmall{0.09}} & \makebox[4.9em][c]{\makebox[2.75em][c]{$20.42$}\,\pmSmall{0.06}} \\
\addlinespace[0.4em]
$\ell_1$-\textsc{sparse} & \makebox[4.9em][c]{\makebox[2.75em][c]{$73.44$}\,\pmSmall{0.30}} & \makebox[4.9em][c]{\makebox[2.75em][c]{$0.35$}\,\pmSmall{0.04}} & \makebox[4.9em][c]{\makebox[2.75em][c]{$39.55$}\,\pmSmall{0.31}} & \makebox[4.9em][c]{\makebox[2.75em][c]{$70.77$}\,\pmSmall{0.35}} & \makebox[4.9em][c]{\makebox[2.75em][c]{$15.78$}\,\pmSmall{0.21}} & \makebox[4.9em][c]{\makebox[2.75em][c]{$18.10$}\,\pmSmall{0.26}} & \makebox[4.9em][c]{\makebox[2.75em][c]{$70.58$}\,\pmSmall{0.39}} & \makebox[4.9em][c]{\makebox[2.75em][c]{$15.76$}\,\pmSmall{0.17}} \\
\addlinespace[0.4em]
\textsc{LUR} & \makebox[4.9em][c]{\makebox[2.75em][c]{$66.48$}\,\pmSmall{0.23}} & \makebox[4.9em][c]{\makebox[2.75em][c]{$\second{0.00}$}\,\pmSmall{0.00}} & \makebox[4.9em][c]{\makebox[2.75em][c]{$47.02$}\,\pmSmall{0.14}} & \makebox[4.9em][c]{\makebox[2.75em][c]{$\second{50.09}$}\,\pmSmall{0.37}} & \makebox[4.9em][c]{\makebox[2.75em][c]{$10.61$}\,\pmSmall{0.25}} & \makebox[4.9em][c]{\makebox[2.75em][c]{$\best{4.46}$}\,\pmSmall{0.37}} & \makebox[4.9em][c]{\makebox[2.75em][c]{$\second{47.85}$}\,\pmSmall{0.08}} & \makebox[4.9em][c]{\makebox[2.75em][c]{$10.79$}\,\pmSmall{0.14}} \\
\addlinespace[0.4em]
\textsc{SSD} & \makebox[4.9em][c]{\makebox[2.75em][c]{$99.06$}\,\pmSmall{1.60}} & \makebox[4.9em][c]{\makebox[2.75em][c]{$0.00$}\,\pmSmall{0.00}} & \makebox[4.9em][c]{\makebox[2.75em][c]{$\second{52.16}$}\,\pmSmall{4.09}} & \makebox[4.9em][c]{\makebox[2.75em][c]{$78.67$}\,\pmSmall{6.21}} & \makebox[4.9em][c]{\makebox[2.75em][c]{$8.11$}\,\pmSmall{0.11}} & \makebox[4.9em][c]{\makebox[2.75em][c]{$15.75$}\,\pmSmall{0.89}} & \makebox[4.9em][c]{\makebox[2.75em][c]{$80.91$}\,\pmSmall{6.33}} & \makebox[4.9em][c]{\makebox[2.75em][c]{$8.70$}\,\pmSmall{0.20}} \\
\addlinespace[0.4em]
\textsc{SalUn} & \makebox[4.9em][c]{\makebox[2.75em][c]{$92.36$}\,\pmSmall{0.43}} & \makebox[4.9em][c]{\makebox[2.75em][c]{$0.00$}\,\pmSmall{0.00}} & \makebox[4.9em][c]{\makebox[2.75em][c]{$46.54$}\,\pmSmall{0.31}} & \makebox[4.9em][c]{\makebox[2.75em][c]{$65.98$}\,\pmSmall{0.82}} & \makebox[4.9em][c]{\makebox[2.75em][c]{$8.01$}\,\pmSmall{0.25}} & \makebox[4.9em][c]{\makebox[2.75em][c]{$12.21$}\,\pmSmall{0.31}} & \makebox[4.9em][c]{\makebox[2.75em][c]{$67.39$}\,\pmSmall{0.79}} & \makebox[4.9em][c]{\makebox[2.75em][c]{$8.40$}\,\pmSmall{0.31}} \\
\addlinespace[0.4em]
\textsc{Amun} & \makebox[4.9em][c]{\makebox[2.75em][c]{$61.95$}\,\pmSmall{14.96}} & \makebox[4.9em][c]{\makebox[2.75em][c]{$29.23$}\,\pmSmall{13.55}} & \makebox[4.9em][c]{\makebox[2.75em][c]{$42.53$}\,\pmSmall{3.72}} & \makebox[4.9em][c]{\makebox[2.75em][c]{$55.79$}\,\pmSmall{5.39}} & \makebox[4.9em][c]{\makebox[2.75em][c]{$21.38$}\,\pmSmall{0.23}} & \makebox[4.9em][c]{\makebox[2.75em][c]{$9.12$}\,\pmSmall{1.04}} & \makebox[4.9em][c]{\makebox[2.75em][c]{$55.75$}\,\pmSmall{5.41}} & \makebox[4.9em][c]{\makebox[2.75em][c]{$21.40$}\,\pmSmall{0.16}} \\
\addlinespace[0.4em]
\textsc{BadTeach} & \makebox[4.9em][c]{\makebox[2.75em][c]{$\second{99.90}$}\,\pmSmall{0.01}} & \makebox[4.9em][c]{\makebox[2.75em][c]{$1.34$}\,\pmSmall{0.19}} & \makebox[4.9em][c]{\makebox[2.75em][c]{$\best{54.14}$}\,\pmSmall{0.10}} & \makebox[4.9em][c]{\makebox[2.75em][c]{$72.68$}\,\pmSmall{0.73}} & \makebox[4.9em][c]{\makebox[2.75em][c]{$\second{6.24}$}\,\pmSmall{0.16}} & \makebox[4.9em][c]{\makebox[2.75em][c]{$11.76$}\,\pmSmall{0.31}} & \makebox[4.9em][c]{\makebox[2.75em][c]{$74.76$}\,\pmSmall{0.51}} & \makebox[4.9em][c]{\makebox[2.75em][c]{$\second{6.79}$}\,\pmSmall{0.13}} \\
\addlinespace[0.4em]
\textbf{\textsc{\method}} & \makebox[4.9em][c]{\makebox[2.75em][c]{$97.53$}\,\pmSmall{1.74}} & \makebox[4.9em][c]{\makebox[2.75em][c]{$1.39$}\,\pmSmall{0.93}} & \makebox[4.9em][c]{\makebox[2.75em][c]{$51.35$}\,\pmSmall{1.04}} & \makebox[4.9em][c]{\makebox[2.75em][c]{$61.38$}\,\pmSmall{2.24}} & \makebox[4.9em][c]{\makebox[2.75em][c]{$\best{4.72}$}\,\pmSmall{0.26}} & \makebox[4.9em][c]{\makebox[2.75em][c]{$\second{7.51}$}\,\pmSmall{0.69}} & \makebox[4.9em][c]{\makebox[2.75em][c]{$64.38$}\,\pmSmall{2.69}} & \makebox[4.9em][c]{\makebox[2.75em][c]{$\best{5.50}$}\,\pmSmall{0.32}} \\
\addlinespace[0.4em]
\bottomrule
\end{tabular}
\end{center}
\end{table}

\begin{table}[h!]
\begin{center}
\small
\caption{\textbf{Half-class forgetting results for Swin-T on Tiny-ImageNet.} RA/FA/TA denote retain, forget, and test accuracy; $\mathrm{RMIA}_{\mathrm{AUC}}$ and $\mathrm{SMIA}_{\mathrm{AUC}}$ report membership inference performance, with 50\% corresponding to chance-level discrimination. Gap metrics measure deviation from \textsc{Retrain}, so lower is better. Results are averaged over the three 20-class forgetting groups within each seed and reported as mean~$\pm$~sample std across three seeds, with all values in percent. \textbf{Bold} and \second{underlined} denote best and second best, where ``best'' denotes smallest gap to \textsc{Retrain}.}
\label{tab:swin-imagenet-class-forgetting-forget0p5}
\vspace*{0.2cm}
\renewcommand{\arraystretch}{0.88}
\setlength{\tabcolsep}{1pt}
\begin{tabular}{@{} l *{4}{M} G *{3}{M} @{}}
\toprule
\addlinespace[0.4em]
& \multicolumn{8}{c}{\textbf{Half-class forgetting: one 20-class group}} \\
\addlinespace[0.4em]
\multicolumn{1}{c}{\small } & \multicolumn{1}{M}{\small RA} & \multicolumn{1}{M}{\small FA} & \multicolumn{1}{M}{\small TA} & \multicolumn{1}{M}{\small $\mathrm{RMIA}_{\mathrm{AUC}}$} & \multicolumn{1}{G}{\small $\mathrm{Gap}_{\mathrm{Avg}}$} & \multicolumn{1}{M}{\small $\mathrm{Gap}^{\mathrm{F\text{-}U}}_{\mathrm{Avg}}$} & \multicolumn{1}{M}{\small $\mathrm{SMIA}_{\mathrm{AUC}}$} & \multicolumn{1}{M}{\small $\mathrm{Gap}^{\mathrm{SMIA}}_{\mathrm{Avg}}$}\\
\addlinespace[0.2em]
\cmidrule(r){2-6}\cmidrule(l){7-9}

\textsc{Retrain} & \makebox[4.9em][c]{\makebox[2.75em][c]{$99.98$}\,\pmSmall{0.00}} & \makebox[4.9em][c]{\makebox[2.75em][c]{$52.38$}\,\pmSmall{0.49}} & \makebox[4.9em][c]{\makebox[2.75em][c]{$60.14$}\,\pmSmall{0.35}} & \makebox[4.9em][c]{\makebox[2.75em][c]{$49.75$}\,\pmSmall{0.26}} & \makebox[4.9em][c]{\makebox[2.75em][c]{$0.00$}\,\pmSmall{0.00}} & \makebox[4.9em][c]{\makebox[2.75em][c]{$0.00$}\,\pmSmall{0.00}} & \makebox[4.9em][c]{\makebox[2.75em][c]{$49.69$}\,\pmSmall{0.36}} & \makebox[4.9em][c]{\makebox[2.75em][c]{$0.00$}\,\pmSmall{0.00}} \\
\addlinespace[0.4em]
\textsc{Base} & \makebox[4.9em][c]{\makebox[2.75em][c]{$99.98$}\,\pmSmall{0.00}} & \makebox[4.9em][c]{\makebox[2.75em][c]{$99.98$}\,\pmSmall{0.00}} & \makebox[4.9em][c]{\makebox[2.75em][c]{$61.01$}\,\pmSmall{0.10}} & \makebox[4.9em][c]{\makebox[2.75em][c]{$80.19$}\,\pmSmall{0.63}} & \makebox[4.9em][c]{\makebox[2.75em][c]{$19.73$}\,\pmSmall{0.31}} & \makebox[4.9em][c]{\makebox[2.75em][c]{$15.65$}\,\pmSmall{0.39}} & \makebox[4.9em][c]{\makebox[2.75em][c]{$94.05$}\,\pmSmall{0.14}} & \makebox[4.9em][c]{\makebox[2.75em][c]{$23.21$}\,\pmSmall{0.22}} \\
\cmidrule(r){1-6}\cmidrule(l){7-9}
\textsc{NegGrad} & \makebox[4.9em][c]{\makebox[2.75em][c]{$\best{99.98}$}\,\pmSmall{0.00}} & \makebox[4.9em][c]{\makebox[2.75em][c]{$99.98$}\,\pmSmall{0.00}} & \makebox[4.9em][c]{\makebox[2.75em][c]{$\best{61.01}$}\,\pmSmall{0.10}} & \makebox[4.9em][c]{\makebox[2.75em][c]{$80.22$}\,\pmSmall{0.63}} & \makebox[4.9em][c]{\makebox[2.75em][c]{$19.73$}\,\pmSmall{0.31}} & \makebox[4.9em][c]{\makebox[2.75em][c]{$15.67$}\,\pmSmall{0.39}} & \makebox[4.9em][c]{\makebox[2.75em][c]{$94.05$}\,\pmSmall{0.14}} & \makebox[4.9em][c]{\makebox[2.75em][c]{$23.21$}\,\pmSmall{0.22}} \\
\addlinespace[0.4em]
\textsc{NegGrad+} & \makebox[4.9em][c]{\makebox[2.75em][c]{$75.01$}\,\pmSmall{1.15}} & \makebox[4.9em][c]{\makebox[2.75em][c]{$\second{50.79}$}\,\pmSmall{5.55}} & \makebox[4.9em][c]{\makebox[2.75em][c]{$44.96$}\,\pmSmall{0.58}} & \makebox[4.9em][c]{\makebox[2.75em][c]{$66.68$}\,\pmSmall{2.42}} & \makebox[4.9em][c]{\makebox[2.75em][c]{$15.26$}\,\pmSmall{0.47}} & \makebox[4.9em][c]{\makebox[2.75em][c]{$16.05$}\,\pmSmall{1.12}} & \makebox[4.9em][c]{\makebox[2.75em][c]{$\second{60.60}$}\,\pmSmall{1.72}} & \makebox[4.9em][c]{\makebox[2.75em][c]{$13.76$}\,\pmSmall{0.61}} \\
\addlinespace[0.4em]
\textsc{Finetune} & \makebox[4.9em][c]{\makebox[2.75em][c]{$99.88$}\,\pmSmall{0.02}} & \makebox[4.9em][c]{\makebox[2.75em][c]{$99.89$}\,\pmSmall{0.02}} & \makebox[4.9em][c]{\makebox[2.75em][c]{$56.97$}\,\pmSmall{0.35}} & \makebox[4.9em][c]{\makebox[2.75em][c]{$78.97$}\,\pmSmall{0.26}} & \makebox[4.9em][c]{\makebox[2.75em][c]{$20.00$}\,\pmSmall{0.19}} & \makebox[4.9em][c]{\makebox[2.75em][c]{$16.19$}\,\pmSmall{0.34}} & \makebox[4.9em][c]{\makebox[2.75em][c]{$87.93$}\,\pmSmall{0.68}} & \makebox[4.9em][c]{\makebox[2.75em][c]{$22.26$}\,\pmSmall{0.34}} \\
\addlinespace[0.4em]
$\ell_1$-\textsc{sparse} & \makebox[4.9em][c]{\makebox[2.75em][c]{$\second{99.88}$}\,\pmSmall{0.03}} & \makebox[4.9em][c]{\makebox[2.75em][c]{$99.88$}\,\pmSmall{0.03}} & \makebox[4.9em][c]{\makebox[2.75em][c]{$56.89$}\,\pmSmall{0.26}} & \makebox[4.9em][c]{\makebox[2.75em][c]{$78.98$}\,\pmSmall{0.21}} & \makebox[4.9em][c]{\makebox[2.75em][c]{$20.02$}\,\pmSmall{0.18}} & \makebox[4.9em][c]{\makebox[2.75em][c]{$16.24$}\,\pmSmall{0.30}} & \makebox[4.9em][c]{\makebox[2.75em][c]{$87.82$}\,\pmSmall{0.70}} & \makebox[4.9em][c]{\makebox[2.75em][c]{$22.25$}\,\pmSmall{0.33}} \\
\addlinespace[0.4em]
\textsc{LUR} & \makebox[4.9em][c]{\makebox[2.75em][c]{$59.71$}\,\pmSmall{0.31}} & \makebox[4.9em][c]{\makebox[2.75em][c]{$37.02$}\,\pmSmall{0.88}} & \makebox[4.9em][c]{\makebox[2.75em][c]{$48.02$}\,\pmSmall{0.22}} & \makebox[4.9em][c]{\makebox[2.75em][c]{$\best{53.14}$}\,\pmSmall{0.31}} & \makebox[4.9em][c]{\makebox[2.75em][c]{$17.78$}\,\pmSmall{0.33}} & \makebox[4.9em][c]{\makebox[2.75em][c]{$7.75$}\,\pmSmall{0.32}} & \makebox[4.9em][c]{\makebox[2.75em][c]{$\best{51.76}$}\,\pmSmall{0.09}} & \makebox[4.9em][c]{\makebox[2.75em][c]{$17.45$}\,\pmSmall{0.32}} \\
\addlinespace[0.4em]
\textsc{SSD} & \makebox[4.9em][c]{\makebox[2.75em][c]{$94.71$}\,\pmSmall{0.01}} & \makebox[4.9em][c]{\makebox[2.75em][c]{$0.00$}\,\pmSmall{0.00}} & \makebox[4.9em][c]{\makebox[2.75em][c]{$54.61$}\,\pmSmall{0.45}} & \makebox[4.9em][c]{\makebox[2.75em][c]{$57.57$}\,\pmSmall{0.20}} & \makebox[4.9em][c]{\makebox[2.75em][c]{$17.75$}\,\pmSmall{0.19}} & \makebox[4.9em][c]{\makebox[2.75em][c]{$\second{6.67}$}\,\pmSmall{0.37}} & \makebox[4.9em][c]{\makebox[2.75em][c]{$85.08$}\,\pmSmall{1.95}} & \makebox[4.9em][c]{\makebox[2.75em][c]{$24.64$}\,\pmSmall{0.41}} \\
\addlinespace[0.4em]
\textsc{SalUn} & \makebox[4.9em][c]{\makebox[2.75em][c]{$88.89$}\,\pmSmall{0.48}} & \makebox[4.9em][c]{\makebox[2.75em][c]{$\best{52.33}$}\,\pmSmall{1.49}} & \makebox[4.9em][c]{\makebox[2.75em][c]{$48.54$}\,\pmSmall{0.48}} & \makebox[4.9em][c]{\makebox[2.75em][c]{$59.41$}\,\pmSmall{0.42}} & \makebox[4.9em][c]{\makebox[2.75em][c]{$\second{8.54}$}\,\pmSmall{0.52}} & \makebox[4.9em][c]{\makebox[2.75em][c]{$10.63$}\,\pmSmall{0.51}} & \makebox[4.9em][c]{\makebox[2.75em][c]{$65.33$}\,\pmSmall{0.30}} & \makebox[4.9em][c]{\makebox[2.75em][c]{$\second{10.04}$}\,\pmSmall{0.37}} \\
\addlinespace[0.4em]
\textsc{Amun} & \makebox[4.9em][c]{\makebox[2.75em][c]{$76.71$}\,\pmSmall{1.34}} & \makebox[4.9em][c]{\makebox[2.75em][c]{$59.96$}\,\pmSmall{1.51}} & \makebox[4.9em][c]{\makebox[2.75em][c]{$46.08$}\,\pmSmall{0.32}} & \makebox[4.9em][c]{\makebox[2.75em][c]{$69.44$}\,\pmSmall{0.27}} & \makebox[4.9em][c]{\makebox[2.75em][c]{$16.15$}\,\pmSmall{0.37}} & \makebox[4.9em][c]{\makebox[2.75em][c]{$16.87$}\,\pmSmall{0.36}} & \makebox[4.9em][c]{\makebox[2.75em][c]{$62.56$}\,\pmSmall{0.48}} & \makebox[4.9em][c]{\makebox[2.75em][c]{$14.45$}\,\pmSmall{0.28}} \\
\addlinespace[0.4em]
\textsc{BadTeach} & \makebox[4.9em][c]{\makebox[2.75em][c]{$99.28$}\,\pmSmall{0.06}} & \makebox[4.9em][c]{\makebox[2.75em][c]{$83.18$}\,\pmSmall{1.28}} & \makebox[4.9em][c]{\makebox[2.75em][c]{$\second{57.79}$}\,\pmSmall{0.23}} & \makebox[4.9em][c]{\makebox[2.75em][c]{$66.96$}\,\pmSmall{0.64}} & \makebox[4.9em][c]{\makebox[2.75em][c]{$12.76$}\,\pmSmall{0.11}} & \makebox[4.9em][c]{\makebox[2.75em][c]{$9.78$}\,\pmSmall{0.45}} & \makebox[4.9em][c]{\makebox[2.75em][c]{$82.23$}\,\pmSmall{0.22}} & \makebox[4.9em][c]{\makebox[2.75em][c]{$16.60$}\,\pmSmall{0.14}} \\
\addlinespace[0.4em]
\textbf{\textsc{\method}} & \makebox[4.9em][c]{\makebox[2.75em][c]{$97.86$}\,\pmSmall{0.09}} & \makebox[4.9em][c]{\makebox[2.75em][c]{$56.90$}\,\pmSmall{2.73}} & \makebox[4.9em][c]{\makebox[2.75em][c]{$53.21$}\,\pmSmall{0.11}} & \makebox[4.9em][c]{\makebox[2.75em][c]{$\second{55.01}$}\,\pmSmall{0.41}} & \makebox[4.9em][c]{\makebox[2.75em][c]{$\best{5.67}$}\,\pmSmall{1.26}} & \makebox[4.9em][c]{\makebox[2.75em][c]{$\best{6.09}$}\,\pmSmall{0.28}} & \makebox[4.9em][c]{\makebox[2.75em][c]{$63.55$}\,\pmSmall{0.54}} & \makebox[4.9em][c]{\makebox[2.75em][c]{$\best{7.83}$}\,\pmSmall{1.23}} \\
\addlinespace[0.4em]
\bottomrule
\end{tabular}
\end{center}
\end{table}

\clearpage

\paragraph{Teacher Ablation.}
Following the analysis of the teacher model in \cref{sec:results}, we also compare the default frozen-base teacher $f_{\theta_0}$ with the \textsc{Retrain}-Teacher $f_{\theta_r}$ in the class-forgetting setting in \Cref{tab:ablation_teacher_resnet_class}.

Under full-class forgetting, the \textsc{Retrain}-Teacher---which is notably unavailable in approximate unlearning---provides a stronger reference for reproducing full-class removal, improving the overall $\mathrm{Gap}_{\mathrm{Avg}}$ in both model--dataset settings and reducing forget accuracy to zero or near zero. This advantage is particularly pronounced for ResNet-18/CIFAR-10, where this factor substantially improves both $\mathrm{Gap}_{\mathrm{Avg}}$ and $\mathrm{Gap}^{\mathrm{SMIA}}_{\mathrm{Avg}}$. This difference is expected: while the frozen-base teacher has learned the removed class or class group, the \textsc{Retrain}-Teacher has never been trained on examples from it and is therefore expected to assign less probability mass to the removed classes when constructing the reference distribution from sampled held-out examples. In this regime, the difference between the teachers is systematic at the class level rather than confined to individual forget examples and can therefore persist after aggregation. Nevertheless, the frozen-base teacher remains competitive on the aggregate metrics and still yields a strong approximation to \textsc{Retrain}.

Under half-class forgetting, the two teachers again perform comparably. Their results are nearly identical for ResNet-18/CIFAR-10, while the differences for Swin-T/Tiny-ImageNet are small relative to the reported variability, with neither teacher consistently improving all aggregate metrics. This is closer to the pattern observed under random instance forgetting: because the retrained model still observes the remaining target-class examples, both teachers have been trained on every class and are therefore likely to produce more similar predictions on held-out data than under full-class removal. This pattern is qualitatively consistent with the attenuation through aggregation observed in \cref{tab:teacher_target_similarity}, although that analysis is restricted to random forgetting. These results suggest that the frozen base model remains a sufficiently strong teacher in the tested half-class forgetting settings.

\begin{table}[h]
\begin{center}
\small
\caption{\textbf{Teacher ablation under class forgetting.} Comparison of the frozen-base and \textsc{Retrain} teachers under full- and half-class forgetting for ResNet-18 on CIFAR-10 and Swin-T on Tiny-ImageNet.}
\label{tab:ablation_teacher_resnet_class}
\vspace*{0.2cm}
\renewcommand{\arraystretch}{0.88}
\setlength{\tabcolsep}{1pt}
\vspace{0.2cm}
\begin{tabular}{@{} l *{4}{>{\centering\arraybackslash}m{4.95em}} >{\columncolor{gapcol}[\tabcolsep][\tabcolsep]\centering\arraybackslash}m{4.95em} *{3}{>{\centering\arraybackslash}m{4.95em}} @{}}

\toprule
\addlinespace[0.4em]
& \multicolumn{8}{c}{\textbf{Full-class forgetting}} \\
\addlinespace[0.4em]
\multicolumn{1}{c}{\small } & \multicolumn{1}{>{\centering\arraybackslash}m{4.95em}}{\small RA} & \multicolumn{1}{>{\centering\arraybackslash}m{4.95em}}{\small FA} & \multicolumn{1}{>{\centering\arraybackslash}m{4.95em}}{\small TA} & \multicolumn{1}{>{\centering\arraybackslash}m{4.95em}}{\small $\mathrm{RMIA}_{\mathrm{AUC}}$} & \multicolumn{1}{>{\columncolor{gapcol}[\tabcolsep][0pt]\centering\arraybackslash}m{4.95em}}{\small $\mathrm{Gap}_{\mathrm{Avg}}$} & \multicolumn{1}{>{\centering\arraybackslash}m{4.95em}}{\small $\mathrm{Gap}^{\mathrm{F\text{-}U}}_{\mathrm{Avg}}$} & \multicolumn{1}{>{\centering\arraybackslash}m{4.95em}}{\small $\mathrm{SMIA}_{\mathrm{AUC}}$} & \multicolumn{1}{>{\centering\arraybackslash}m{4.95em}}{\small $\mathrm{Gap}^{\mathrm{SMIA}}_{\mathrm{Avg}}$}\\
\addlinespace[0.2em]
\midrule
\multicolumn{8}{l}{\textbf{ResNet-18 on CIFAR-10}} \\
\midrule
\addlinespace[0.2em]
\textsc{Retrain}-Teacher & \makebox[4.95em][c]{\makebox[2.85em][c]{$99.47$}\,\pmSmall{0.04}} & \makebox[4.95em][c]{\makebox[2.85em][c]{$0.00$}\,\pmSmall{0.00}} & \makebox[4.95em][c]{\makebox[2.85em][c]{$83.72$}\,\pmSmall{0.15}} & \makebox[4.95em][c]{\makebox[2.85em][c]{$50.82$}\,\pmSmall{0.29}} & \makebox[4.95em][c]{\makebox[2.85em][c]{$0.79$}\,\pmSmall{0.11}} & \makebox[4.95em][c]{\makebox[2.85em][c]{$1.32$}\,\pmSmall{0.23}} & \makebox[4.95em][c]{\makebox[2.85em][c]{$53.21$}\,\pmSmall{0.24}} & \makebox[4.95em][c]{\makebox[2.85em][c]{$1.27$}\,\pmSmall{0.10}} \\
\addlinespace[0.2em]
Frozen Base & \makebox[4.95em][c]{\makebox[2.85em][c]{$99.27$}\,\pmSmall{0.25}} & \makebox[4.95em][c]{\makebox[2.85em][c]{$3.42$}\,\pmSmall{0.60}} & \makebox[4.95em][c]{\makebox[2.85em][c]{$83.63$}\,\pmSmall{0.23}} & \makebox[4.95em][c]{\makebox[2.85em][c]{$50.00$}\,\pmSmall{0.28}} & \makebox[4.95em][c]{\makebox[2.85em][c]{$1.73$}\,\pmSmall{0.25}} & \makebox[4.95em][c]{\makebox[2.85em][c]{$1.39$}\,\pmSmall{0.17}} & \makebox[4.95em][c]{\makebox[2.85em][c]{$51.30$}\,\pmSmall{0.06}} & \makebox[4.95em][c]{\makebox[2.85em][c]{$1.90$}\,\pmSmall{0.24}} \\
\addlinespace[0.2em]
\midrule
\multicolumn{8}{l}{\textbf{Swin-T on Tiny-ImageNet}} \\
\midrule
\addlinespace[0.2em]
\textsc{Retrain}-Teacher & \makebox[4.95em][c]{\makebox[2.85em][c]{$97.80$}\,\pmSmall{0.19}} & \makebox[4.95em][c]{\makebox[2.85em][c]{$0.02$}\,\pmSmall{0.01}} & \makebox[4.95em][c]{\makebox[2.85em][c]{$52.50$}\,\pmSmall{0.02}} & \makebox[4.95em][c]{\makebox[2.85em][c]{$63.00$}\,\pmSmall{0.91}} & \makebox[4.95em][c]{\makebox[2.85em][c]{$4.42$}\,\pmSmall{0.08}} & \makebox[4.95em][c]{\makebox[2.85em][c]{$7.74$}\,\pmSmall{0.25}} & \makebox[4.95em][c]{\makebox[2.85em][c]{$70.73$}\,\pmSmall{0.25}} & \makebox[4.95em][c]{\makebox[2.85em][c]{$6.39$}\,\pmSmall{0.07}} \\
\addlinespace[0.2em]
Frozen Base & \makebox[4.95em][c]{\makebox[2.85em][c]{$97.53$}\,\pmSmall{1.74}} & \makebox[4.95em][c]{\makebox[2.85em][c]{$1.39$}\,\pmSmall{0.93}} & \makebox[4.95em][c]{\makebox[2.85em][c]{$51.35$}\,\pmSmall{1.04}} & \makebox[4.95em][c]{\makebox[2.85em][c]{$61.38$}\,\pmSmall{2.24}} & \makebox[4.95em][c]{\makebox[2.85em][c]{$4.72$}\,\pmSmall{0.26}} & \makebox[4.95em][c]{\makebox[2.85em][c]{$7.51$}\,\pmSmall{0.69}} & \makebox[4.95em][c]{\makebox[2.85em][c]{$64.38$}\,\pmSmall{2.69}} & \makebox[4.95em][c]{\makebox[2.85em][c]{$5.50$}\,\pmSmall{0.32}} \\
\midrule
\addlinespace[0.4em]
& \multicolumn{8}{c}{\textbf{Half-class forgetting}} \\
\addlinespace[0.4em]
\multicolumn{1}{c}{\small } & \multicolumn{1}{>{\centering\arraybackslash}m{4.95em}}{\small RA} & \multicolumn{1}{>{\centering\arraybackslash}m{4.95em}}{\small FA} & \multicolumn{1}{>{\centering\arraybackslash}m{4.95em}}{\small TA} & \multicolumn{1}{>{\centering\arraybackslash}m{4.95em}}{\small $\mathrm{RMIA}_{\mathrm{AUC}}$} & \multicolumn{1}{>{\columncolor{gapcol}[\tabcolsep][0pt]\centering\arraybackslash}m{4.95em}}{\small $\mathrm{Gap}_{\mathrm{Avg}}$} & \multicolumn{1}{>{\centering\arraybackslash}m{4.95em}}{\small $\mathrm{Gap}^{\mathrm{F\text{-}U}}_{\mathrm{Avg}}$} & \multicolumn{1}{>{\centering\arraybackslash}m{4.95em}}{\small $\mathrm{SMIA}_{\mathrm{AUC}}$} & \multicolumn{1}{>{\centering\arraybackslash}m{4.95em}}{\small $\mathrm{Gap}^{\mathrm{SMIA}}_{\mathrm{Avg}}$}\\
\addlinespace[0.2em]
\midrule
\multicolumn{8}{l}{\textbf{ResNet-18 on CIFAR-10}} \\
\midrule
\addlinespace[0.2em]
\textsc{Retrain}-Teacher & \makebox[4.95em][c]{\makebox[2.85em][c]{$99.33$}\,\pmSmall{0.10}} & \makebox[4.95em][c]{\makebox[2.85em][c]{$92.85$}\,\pmSmall{0.76}} & \makebox[4.95em][c]{\makebox[2.85em][c]{$91.99$}\,\pmSmall{0.40}} & \makebox[4.95em][c]{\makebox[2.85em][c]{$53.59$}\,\pmSmall{0.34}} & \makebox[4.95em][c]{\makebox[2.85em][c]{$2.40$}\,\pmSmall{0.34}} & \makebox[4.95em][c]{\makebox[2.85em][c]{$2.68$}\,\pmSmall{0.51}} & \makebox[4.95em][c]{\makebox[2.85em][c]{$55.23$}\,\pmSmall{0.22}} & \makebox[4.95em][c]{\makebox[2.85em][c]{$2.78$}\,\pmSmall{0.25}} \\
\addlinespace[0.2em]
Frozen Base & \makebox[4.95em][c]{\makebox[2.85em][c]{$99.39$}\,\pmSmall{0.07}} & \makebox[4.95em][c]{\makebox[2.85em][c]{$94.34$}\,\pmSmall{0.74}} & \makebox[4.95em][c]{\makebox[2.85em][c]{$92.11$}\,\pmSmall{0.35}} & \makebox[4.95em][c]{\makebox[2.85em][c]{$53.58$}\,\pmSmall{0.19}} & \makebox[4.95em][c]{\makebox[2.85em][c]{$2.47$}\,\pmSmall{0.14}} & \makebox[4.95em][c]{\makebox[2.85em][c]{$2.63$}\,\pmSmall{0.45}} & \makebox[4.95em][c]{\makebox[2.85em][c]{$55.19$}\,\pmSmall{0.16}} & \makebox[4.95em][c]{\makebox[2.85em][c]{$2.84$}\,\pmSmall{0.11}} \\
\addlinespace[0.2em]
\midrule
\multicolumn{8}{l}{\textbf{Swin-T on Tiny-ImageNet}} \\
\midrule
\addlinespace[0.2em]
\textsc{Retrain}-Teacher & \makebox[4.95em][c]{\makebox[2.85em][c]{$98.24$}\,\pmSmall{0.24}} & \makebox[4.95em][c]{\makebox[2.85em][c]{$57.62$}\,\pmSmall{2.26}} & \makebox[4.95em][c]{\makebox[2.85em][c]{$53.61$}\,\pmSmall{0.45}} & \makebox[4.95em][c]{\makebox[2.85em][c]{$55.88$}\,\pmSmall{0.74}} & \makebox[4.95em][c]{\makebox[2.85em][c]{$5.32$}\,\pmSmall{0.79}} & \makebox[4.95em][c]{\makebox[2.85em][c]{$6.32$}\,\pmSmall{0.63}} & \makebox[4.95em][c]{\makebox[2.85em][c]{$63.59$}\,\pmSmall{0.98}} & \makebox[4.95em][c]{\makebox[2.85em][c]{$7.27$}\,\pmSmall{0.57}} \\
\addlinespace[0.2em]
Frozen Base & \makebox[4.95em][c]{\makebox[2.85em][c]{$97.86$}\,\pmSmall{0.09}} & \makebox[4.95em][c]{\makebox[2.85em][c]{$56.90$}\,\pmSmall{2.73}} & \makebox[4.95em][c]{\makebox[2.85em][c]{$53.21$}\,\pmSmall{0.11}} & \makebox[4.95em][c]{\makebox[2.85em][c]{$55.01$}\,\pmSmall{0.41}} & \makebox[4.95em][c]{\makebox[2.85em][c]{$5.67$}\,\pmSmall{1.26}} & \makebox[4.95em][c]{\makebox[2.85em][c]{$6.09$}\,\pmSmall{0.28}} & \makebox[4.95em][c]{\makebox[2.85em][c]{$63.55$}\,\pmSmall{0.54}} & \makebox[4.95em][c]{\makebox[2.85em][c]{$7.83$}\,\pmSmall{1.23}} \\
\bottomrule
\end{tabular}

\end{center}
\end{table}

\end{document}